\documentclass[10pt]{article} 
\usepackage[preprint]{tmlr}

\usepackage{microtype}
\usepackage{graphicx}
\usepackage{subfig}
\usepackage{booktabs} 

\usepackage{hyperref}

\usepackage{amsmath}
\usepackage{amssymb}
\usepackage{mathtools}
\usepackage{amsthm}
\usepackage{amsbsy}

\usepackage{tabularray}
\usepackage{colortbl}

\theoremstyle{plain}
\newtheorem{theorem}{Theorem}[section]

\theoremstyle{definition}
\newtheorem{definition}[theorem]{Definition}

\theoremstyle{remark}

\newcommand{\approptoinn}[2]{\mathrel{\vcenter{
  \offinterlineskip\halign{\hfil$##$\cr
    #1\propto\cr\noalign{\kern2pt}#1\sim\cr\noalign{\kern-2pt}}}}}

\newcommand{\appropto}{\mathpalette\approptoinn\relax}

\newcommand{\indep}{\perp \!\!\! \perp}

\newcommand{\vx}{\mathbf{x}}

\DeclareMathOperator*{\argmin}{arg\,min}
\newcommand{\xhdr}[1]{\vspace{1mm} \noindent{\bf #1}.}
\newcommand{\qt}[1]{\textit{#1}}

\usepackage{hyperref}
\usepackage{url}

\title{Rethinking Distance Metrics for Counterfactual Explainability}

\author{
\name Joshua Nathaniel Williams \\ \email jnwillia@cs.cmu.edu \\ 
\addr Department of Computer Science \\
Carnegie Mellon University \\
Pittsburgh, PA 15213, USA  
\AND
\name Anurag Katakkar 
\\ \addr
Department of Computer Science \\
Carnegie Mellon University \\
Pittsburgh, PA 15213, USA 
\AND
\name Hoda Heidari \\ 
\addr Department of Machine Learning \\
Carnegie Mellon University \\
Pittsburgh, PA 15213, USA
\AND
\name J. Zico Kolter \\
\addr Department of Machine Learning \\
Carnegie Mellon University \\
Pittsburgh, PA 15213, USA 
}

\def\month{7}  
\def\year{2023} 
\def\openreview{\url{https://openreview.net/forum?id=XXXX}} 

\usepackage[textsize=tiny]{todonotes}
\usepackage{comment}

\begin{document}

\maketitle

\begin{abstract}
Counterfactual explanations have been a popular method of post-hoc explainability for a variety of settings in Machine Learning. Such methods focus on explaining classifiers by generating new data points that are similar to a given reference, while receiving a more desirable prediction. In this work, we investigate a framing for counterfactual generation methods that considers counterfactuals not as independent draws from a region around the reference, but as jointly sampled with the reference from the underlying data distribution. Through this framing, we derive a distance metric, tailored for counterfactual similarity that can be applied to a broad range of settings. Through both quantitative and qualitative analyses of counterfactual generation methods, we show that this framing allows us to express more nuanced dependencies among the covariates.
\end{abstract}

\section{Introduction}

The ubiquity of modern Machine Learning (ML) applications in high-stakes contexts, such as parole decisions, lending, or healthcare, has long necessitated mechanisms for explaining outcomes to those impacted by their predictions---including the direct \emph{subjects} of their predictions. Many local explanation techniques have been proposed, and of these, \emph{counterfactual explanations}~\citep{wachter2017counterfactual} have been particularly popular. These techniques focus on deriving explanations by investigating ``what-if'' scenarios: ``What if my salary was higher? Would my loan application have been approved?'' 
Such explanations have the potential to provide a form of \emph{recourse}~\citep{ustun2019actionable} if they are \emph{plausible}, wherein the explanation is not self-contradictory and points to a viable real-world profile of attributes; and \emph{actionable}, wherein explanations recommend modifications that one could act on (e.g., not recommending that a person reduces their age, or get a doctorate, when they only have high-school education) \citep{mahajan2019preserving}.
However, as pointed out in \citep{barocas2020hidden}, counterfactual explanations have distinct challenges, including: 1) emphasizing the features that are easiest to change may conceal the fact that decisions still rely on immutable characteristics; 2) explanations may react to underlying information that is invisible to the model; 3) `The Framing Trap' as described in \citep{selbst2019fairness}, pointing to the failure of the model to capture the entire social system from which the data is generated. 

In this work, we investigate the relationship between a known data point, its counterfactuals, and the underlying data distribution. We show in Section \ref{sec:background}, that the implicit decisions made on this relationship have strong implications for the resultant counterfactuals. While there exists a significant body of work that studies how to generate counterfactuals that respect the underlying data distribution (For example, \citep{karimi2020algorithmic} show that even under imperfect knowledge of an underlying causal model, we can craft approaches that encourage meaningful forms of recourse and \citep{pawelczyk2020learning} show that the latent space of a variational autoencoder holds a depth of knowledge that allows us to find counterfactuals), we show that our framing of the relationship between counterfactual and reference is enough to encourage semantically meaningful counterfactuals, even under comparatively weak assumptions on the structure of the underlying data. Our contributions are summarized as follows: 

\textit{(1)} We posit a simple change to the Probablistic Graphical Model (PGM) that underlies common methods of generating counterfactual explanations (Section~\ref{sec:background}) and argue that this approach results in explanations that are representative of the underlying data distribution.
\textit{(2)} We show how to enforce, within the explanations, several ideas of plausibility and actionability that have been discussed in prior literature (Section~\ref{sec:feasible_cf}).
\textit{(3)} We use our new assumption on the relationship between the counterfactual and reference to derive a specialized counterfactual distance function (Section~\ref{sec:fashion_mnist}).
\textit{(4)} Finally, in Section~\ref{sec:eval}, we compare the efficacy of our approach across several datasets and metrics, to show that we are able to generate counterfactuals that are more faithful to the underlying distribution of ground truth data.

\section{Background and Motivation}
\label{sec:background}

Depending on the underlying decision-making model, the difficulty of providing explanations varies. From simpler rule-based systems in which we understand decisions in the context of the rules; to decision trees in which explainability amounts to following along branches; to deep networks, where, while we can trace the model's activations, doing so is largely meaningless in providing a human understanding of a decision. Due to the complexity inherent to such decision-making systems, there is a large body of prior work that focuses on finding local explanations by relying on `feature highlighting' techniques. In a broad sense, these methods explain a given input by selecting relevant features that heavily influence the model's output. Throughout this section we first provide a general overview of several `feature highlighting' methods, and then turn our focus to understanding the counterfactual explanation setting and situating our work within this space.

\subsection{Overview of Feature Highlighting Methods for Explainability}

\label{app:feature_highlight}

Feature highlighting methods encompass a broad set of explanation techniques that show users a set of features that are `important' for the underlying decision-maker. Many of the approaches describe explainability through the lens of a specific motivating question.
Proxy models \citep{ribeiro2016should}, for example, focus on answering the question, "What if we learn an interpretable model that makes the same decisions as a complex model?". As many interpretable models, such as logistic regressions, allow us to explicitly see how a feature influences our output, we expect that understanding how the interpretable model behaves with respect to each feature will serve as an explanation for the complex model by proxy. 
Such methods can be contrasted with Saliency Maps \citep{selvaraju2017grad, sundararajan2017axiomatic, smilkov2017smoothgrad, li2023towards} that provide a score for each of the input features (commonly using information about the gradients at some point in the network) and present to the user how each feature relates to the output. Additional work \citep{adebayo2018sanity,tomsett2020sanity, amorim2023evaluating} has also provided sanity checks for such methods in order to guide researchers in deciding when and which method best meets the needs of their task at hand. 

Ideas presented by gradient-based saliency maps have given rise to saliency maps that incorporate our ideas of causality \citep{baron2023explainable}. \citet{zhao2021causal} had the insight that a commonly used visualization of black-box models, Partial Dependence Plots (PDP) \citep{greenwell2017pdp}, is effectively equivalent to Pearl's Backdoor Criterion \citep{peters2017elements}. Thus, PDPs not only provide information on the relationship between the target output and a feature, but also their \emph{causal} relationship. In a similarly vein, further work \citet{schwab2019cxplain} considers the case of ``Granger Causality'' in which a signal, $X$ is said to \emph{cause}, $Y$, if there exist no features outside of $X$ that provide additional predictive performance. The change in predictive performance with/without each feature and can then be scored to see how much each feature can be said to cause the target variable. 

Other ideas of feature highlighting have also been popularized. Pulling ideas from Game Theory \citep{roth1988introduction}, explainability methods that rely on shapley values \citep{giudici2021shapley, wang2021shapley, sundararajan2020many,chen2023algorithms} treat each feature as one of a set of players working together toward the goal of minimizing the loss for the learning task. The unique division of contributions for the entire group, using Lloyd Shapley's approach determines payments in proportion to that player's (feature's) marginal contribution. 

Additional work has also sought to highlight input features indirectly. One popular subset, concept bottlenecks \citep{koh2020concept, wong2021explainable, huang2024concept}, focus on generating a set of user-defined and understood concepts (eg. color, shape, size, etc ) and training a model to learn these concepts. These concepts are intended to provide no less information than that in the input, so that they then act as its proxy. An interpretable model is then used on these concepts in order to make a decision, eg. "This bird was classified as a robin, because of its round body, the color of its belly was red, and its length was 25cm." Notably, it has been highlighted \citep{margeloiu2021concept, furby2023towards} that learned concepts may not be based on semantically meaningful representations in the input space, prompting further research in this space. 
Of the many avenues for growth within the space of explanations that highlight specific input features, this work focuses on the particularly popular method, counterfactual explainability. 

\subsection{Counterfactual Explanations}
\label{vs_optim}

 \begin{figure*}
    \centering
    \subfloat[Standard PGM\label{fig:pgm1}]{%
      \includegraphics[height=2.7cm]{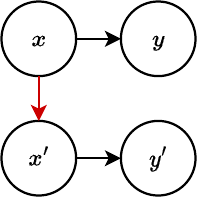}
    }
    \hspace{0.5cm}
    \subfloat[$L = I, \gamma = 1$\label{fig:pgm1_distb}]{%
      \includegraphics[height=2.9cm]{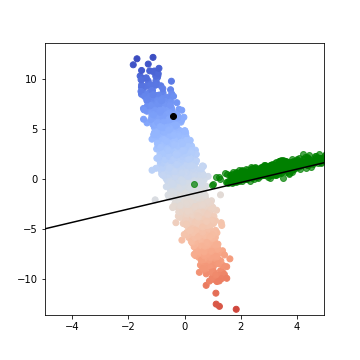}
    }
    \hspace{0.5cm}
    \subfloat[$L = 0, \gamma = 1$\label{fig:pgm1_distc}]{%
      \includegraphics[height=2.9cm]{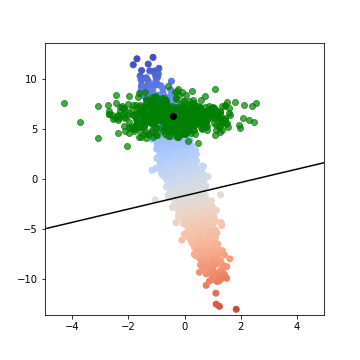}
    }
    \hspace{0.5cm}
    \subfloat[$L = I, \gamma = 0.05$\label{fig:pgm1_distd}]{%
      \includegraphics[height=2.9cm]{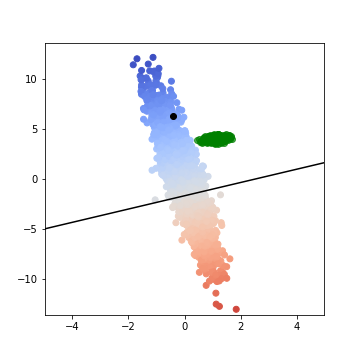}
    }
    
    \subfloat[Proposed PGM\label{fig:pgm2}]{%
      \includegraphics[height=2.7cm]{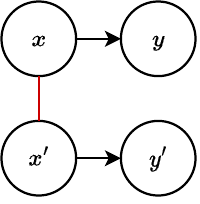}
    }
    \hspace{0.5cm}
    \subfloat[$L = I, \alpha = 0$\label{fig:pgm2_distb}]{%
      \includegraphics[height=2.9cm]{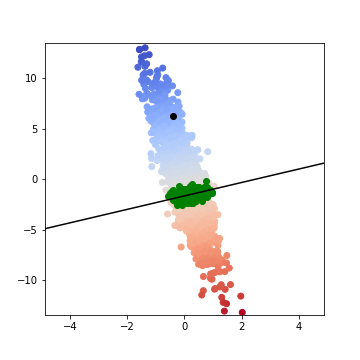}
    }
    \hspace{0.5cm}
    \subfloat[$L = 0, \alpha = 0$\label{fig:pgm2_distc}]{%
      \includegraphics[height=2.9cm]{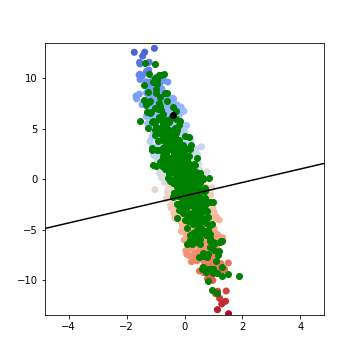}
    }
    \hspace{0.5cm}
    \subfloat[$L = I, \alpha = 0.995$\label{fig:pgm2_distd}]{%
      \includegraphics[height=2.9cm]{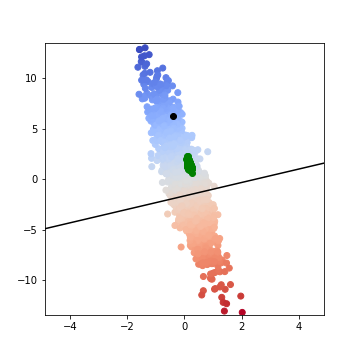}
    }
\caption{Comparison of approaches to counterfactual generation; counterfactuals with the proposed prior never leave the data distribution.
~
(Black Dot) Reference, $\vx$. (Green) Counterfactual Distribution. (Black Line) Desired predicted output, $y' = A \vx' + b$. 
~
In all figures, $L$ is the precision of residuals, $\gamma$ is the weight on $l_{2}$ distance, and $\alpha$ controls similarity/distance in our approach. 
~
}
\label{fig:pgm1_all}
\end{figure*}

Counterfactual Explanations \citep{kang2020counterfactual, mahajan2019preserving, wachter2017counterfactual}\nocite{mothilal2020explaining,hendricks2018generating,black2021consistent} seek to provide a user with a set of points from the input space that are similar to the initial feature vector or \emph{reference}, but receive a different prediction by the decision-making model.

\begin{definition}[Counterfactual Explanations]
For some input space, $\mathcal{X}$, consider a decision-making model $f: \mathcal{X} \rightarrow \mathcal{Y}$, a reference point, $\vx \in \mathcal{X}$, and a desired predicted label $y'$. Let $\epsilon, \delta \in \mathbb{R}^+$ be two given parameters. The set of counterfactual explanations, with parameters $\epsilon, \delta$ for the predicted label, $f( \vx ) \neq y'$, is defined as follows:
\begin{align}
\begin{split}
     \mathrm{cf}( \vx, y'; \epsilon, \delta ) := \{ \vx' &\sim \mathcal{X} : \\
     \mathrm{dist}_{1}(  y', f( &\vx' ) ) \leq \delta, \hspace{3mm} 
     \mathrm{dist}_{2}( \vx, \vx' ) \leq \epsilon \},
     \label{eq:cf_form}
\end{split}
\end{align}
where $\mathrm{dist}_{i}: \mathbb{R}^{n} \times \mathbb{R}^{n} \rightarrow \mathbb{R}^{+}$ are distance functions. 
\label{def:cf}
\end{definition}

Prior work often expresses the distance between the desired predicted outcome, $y'$, and the label of the counterfactual, $f( \vx' )$, as some convex loss function (eg. squared error), and the distance between the reference and counterfactual as some applicable distance metric, such as the $l_1$ norm scaled by the Median Absolute Deviation (MAD), Edit Distances \citep{guo2023counterfactual} or the Euclidean/Mahalanobis distance \citep{mothilal2020explaining,kang2020counterfactual,mahajan2019preserving,wachter2017counterfactual}:
\begin{align*}
    \mathrm{dist}_{1}( y', f( \vx' ) ) &:= || y' - f( \vx' ) ||_{2}^{2}, \hspace{5mm} \\
    \mathrm{dist}_{2}( \vx, \vx' ) &:= || \vx - \vx' ||_{2}^{2} 
\end{align*}
This immediately gives rise to the most common method of solving Eq \eqref{eq:cf_form}; minimize the sum of $\mathrm{dist}_{1}$ and $\mathrm{dist}_{2}$, which as pointed out in \citep{freiesleben2020counterfactual}, is akin to an adversarial attack on the classifier,
\begin{equation}
    \vx' = \arg \min_{\tilde{\vx}} || y' - f( \tilde{\vx} ) ||_{2}^{2} + \gamma || \vx - \tilde{\vx} ||_{2}^{2},
    \label{eq:opt}
\end{equation}
where $\gamma \in \mathbb{R}^{+}$ scales the influence on distance.
 While this form underlies a large portion of work, authors often apply additional regularization or pre/post-processing to create desirable properties. For example, \citet{mothilal2020explaining} introduce a diversity regularizer to encourage subsequent explanations to be distinct from one another. \citet{kang2020counterfactual} solve Eq.~\eqref{eq:opt} via coordinate descent in order to minimize hamming distance, and \citep{slack2021counterfactual, mahajan2019preserving} introduce regularizers that encourage low-cost solutions in terms of fairness/causality respectively. A detailed review of other such methods can be found in \citep{verma2020counterfactual}.
 
 \subsection{Motivation for the Proposed Method}
 \label{sec:motivation}
 
 Prior work \citep{laugel2019dangers, barocas2020hidden} has expressed concern about generating counterfactuals via a variation of Eq.~\eqref{eq:opt}, due to their inability to guarantee actionability for explainees. In order to present a perspective on why such methods lead to these issues and to motivate our approach, consider the simple case of a Linear Regression Model, $y = A \vx + b + \epsilon$. Our labels, $y$, are samples from, $\mathcal{N}( A \vx + b, L^{-1} )$, where $\vx \sim \mathcal{N}( \mu, \Lambda^{-1} )$, $\mu \in \mathbb{R}^{n}$, and $\Lambda^{-1}, L$ are PSD matrices in $\mathbb{R}^{n \times n}$ and $\mathbb{R}^{|y| \times |y|}$ respectively. By re-framing Eq.~\eqref{eq:opt} as an equivalent quadratic,  
   \begin{align}
   \begin{split}
        \vx'  = \argmin_{\tilde{\vx}} &( y' - A \tilde{\vx} - b )^{T} L ( y' - A \tilde{\vx} - b ) \\
        &+ ( \vx - \tilde{\vx} ) ( \gamma I )  ( \vx - \tilde{\vx} ).
        \label{eq:quad_opt}
    \end{split}
    \end{align}
The earlier objective becomes the negative log probability of some known Gaussian distribution (derivation and the parameters provided in Appendix \ref{app:derivation_pgm1}). Counterfactual generation methods, in the linear setting, can be framed as an instance of sampling from this Gaussian distribution. Moreover the solution to Eq.~\eqref{eq:opt} is the mode of that entailed distribution.

Underlying this process is the generative model expressed in Fig.~ \ref{fig:pgm1}. This graph is a representation of the counterfactual posterior for reference, $\vx$, and desired predicted outcome, $y'$,
\begin{equation}
    p( \vx' | \vx, y' ) \propto p( y' | \vx' ) p( \vx' | \vx ) p( \vx ).
    \label{eq:opt_posterior}
\end{equation}
As the reference,  $\vx$, is fixed a priori, $p( \vx )$ can be pushed into the proportionality constant and our prior over the explanations effectively becomes $p( \vx' | x ) = \mathcal{N}( \vx, \gamma I )$. Such a prior, assumes that counterfactual explanations do not come from the true data distribution. Instead, this states that such explanations \emph{only} exist in relation to the reference point. 

Note that this simple generative model depicts, not the \emph{data} generation process, but the assumptions inherent within the \emph{counterfactual} generation process. Advancements from prior work that focus on the data generation process, are parallel to our investigation of the assumptions on the counterfactual generation process. We emphasize that by not associating the explanation generation process with the underlying data distribution, it gives rise to the potential for the generative model \ref{fig:pgm1} to produce explanations outside of the data distribution\footnote{One may suspect for Gaussian data, the lack of representation of the underlying distribution can be corrected by applying a Gaussian regularizer. Appendix \ref{app:regularization} investigates this setting, and we show that not only does such regularization not address the issues above, the graphical model that underlies such a case, goes against our understanding of the definition of a counterfactual explanation.}. 

We show several visualizations of this effect in Fig.~\ref{fig:pgm1_all}. Fig.~\ref{fig:pgm1_distb} shows that under common conditions (euclidean distance and variance of residuals is $1$), the distribution of counterfactuals can sit entirely in a regions of the space that have near-zero probability wrt. the distribution of data. Fig.~\ref{fig:pgm1_distc} shows that under the case where we place no emphasis on accuracy for the desired counterfactual, the distribution of counterfactuals centers around the reference, yet it still has tails that lie in these near-zero probability regions.

\section{Ensuring Representative Counterfactuals}
\label{sec:method_intro}

In this section, we introduce the proposed framework for generating counterfactual explanations. For ease of exposition, we continue to focus on the case of explaining Linear Regression Models, $f( \vx ) = A \vx + b $, before expanding to more complex settings, including neural networks, in Appendix \ref{app:complex}. Although linear models often do not need explanations, such models exactly express the distribution of counterfactual explanations and serve as a clear comparison to Eq. \ref{eq:opt}.

Given an input, $\vx \sim \mathcal{N}( \mu, \Lambda^{-1} )$, to a decision-making model, $f$, with output $y = f( \vx )$, counterfactual explanations methods seek to explain why the model labeled $\vx$ with label $y$, by choosing points, $\vx'$, from the set of all possible counterfactuals (Def. \ref{def:cf}).
This set of explanations is expressed via three components: A prior on the relationship between the reference and the counterfactual, the likelihood of the desired $y'$ given $x'$, and a prior on the data distribution.

The key idea of our approach is that while counterfactuals are often considered to be wholly dependent on the reference, as shown by the directed edge in Fig.~\ref{fig:pgm1}, we should treat $\vx$ and $\vx'$ as dependent on one another. Just as we consider a reference, $\vx$, as existing somewhere within the input space, counterfactual explanations exist a priori within this space. Their codependency is expressed in the generative model (Fig.~\ref{fig:pgm2}) via an \emph{undirected} edge between $\vx$ and $\vx'$. 

While a subtle distinction, the choice of joint distribution over $\vx$ and $\vx'$ has a significant impact on the selected counterfactuals. In this work we express the distribution over reference and counterfactual with the form,
\begin{equation}
    \label{eq:xx'_joint}
    p( \vx, \vx' ) = \mathcal{N}\Bigg( \begin{bmatrix} \mu \\ \mu \end{bmatrix}, \begin{bmatrix} \Lambda^{-1} & W \\ W^{T} & \Lambda^{-1} \end{bmatrix} \Bigg)\\
\end{equation}
The relationship between $x$ and $x'$ are entirely defined by a correlation matrix, $W$, and the marginals are defined as the observed data distribution, $\mathcal{N}( \mu, \Lambda^{-1} )$. While $W$ can be any positive semi-definite matrix, in order to express the correlation between counterfactual and reference, we suggest defining,
$
    \nonumber
    W = \alpha \Lambda^{-1}\text{, where } \alpha \in ( 0, 1 ).
$
Should $\alpha = 1$, we have the degenerate case in which $\vx$ and $\vx'$ are perfectly correlated. This places no emphasis on having $f( \vx' ) = y'$. On the other hand, $\alpha = 0$ implies that $\vx$ and $\vx'$ are independent draws from the same distribution, which in turn emphasizes choosing $\vx'$ such that $f( \vx' ) = y'$. Scaling $\alpha$ from $1$ to $0$ scales the similarity between reference and counterfactual.

As in the previous section, the posterior of our recommended graphical model remains Gaussian. Moreover, we can express its distribution, for a linear regression, analytically (full derivation and parameters provided in Appendix \ref{app:derivation_pgm2}).
~
Under this framing, we generate similar distributions to those shown in the top half of Figure \ref{fig:pgm1_all}. The joint prior recommended here restricts the distributions of counterfactual explanations to stay within the data distribution. The most striking example of which, Figure \ref{fig:pgm2_distc} well illustrates the implications of this new prior, and the semantic questions that we pose. If we ask an algorithm to generate a counterfactual which neither emphasizes the desired predicted label, $y'$, nor the similarity to the reference, $\vx$, the \citet{wachter2017counterfactual} framing from Eq.~\eqref{eq:opt}, returns any value, $\vx' \in \mathcal{R}^{n}$, however, in this same circumstance, the form introduced here is constructed to exactly match the data distribution. Without emphasis on $y'$ nor $\vx$, counterfactuals are simply samples from the data distribution.

\section{Domain Knowledge in the Prior}

As stated in prior work \citep{karimi2021algorithmic, laugel2019dangers}, the challenge of generating counterfactual explanations hinges on finding changes to the input that are plausible (ie. the explanation could potentially exist), actionable (ie. the explanation recommends changes that are possible for one to make), and give the explainee direction to change themselves. In this section, we show how the counterfactual prior, $p( \vx, \vx' )$, and the resultant posterior, can express several forms of actionability.
While, one can use any off-the-shelf method of sampling from a non-Gaussian posterior, throughout the remainder of this work, we focus on the Gaussian case in order to ensure an easy to visualization. 

\subsection{Accounting for Actionability Constraints}
\label{sec:feasible_cf}

As described in \citet{karimi2021algorithmic}, the features of a actionable counterfactual explanations can be subdivided into three distinct categories:
(a) \textbf{Mutable:} features for which a counterfactual explanation may change freely (Eg. bank account balance);
(b) \textbf{Immutable:} Non-Actionable features for which under no circumstances we change from the reference input (eg. race);
(c) \textbf{Mutable but Non-Actionable:} features that can change only as a result of other features changing (eg. credit score).
Such explanations can be achieved by manipulation of the prior on the distance between the reference and counterfactual, the prior on counterfactual distribution, and the posterior, $p( \vx' | \vx, y' )$.

\xhdr{Mutable} Mutable features may be freely changed and require no additional transformations.

\xhdr{Immutable} Recall that we express the correlation between reference $\vx$ and counterfactual $\vx'$ as, $W = \alpha \Lambda^{-1}$, where $\alpha \in (0,1)$ and $\Lambda^{-1}$ is positive semi-definite. If $\alpha = 1$, the reference and the counterfactual are perfectly correlated and $\vx = \vx'$. As such, we can express immutable features through the covariance, $cov( \vx, \vx' ) = W$. We set features as immutable through the following adjustment to $W$:
\begin{align*}
    W &= \sigma \sigma^T \odot ( \alpha - 1 ) \Lambda^{-1} + \Lambda^{-1}\\
    \sigma_{i} &= \begin{cases} 
        0 \hspace{1cm} x_{i} \in \mathrm{immutable} \\ 
        1 \hspace{1cm} o.w. ,
        \end{cases}
\end{align*}
In other words, we enforce immutability by requiring a perfect correlation between immutable features of $\vx'$ and $\vx$. 

\xhdr{Mutable, Non-Actionable} For such cases in which an explainee may be unable to directly influence an outcome (eg. one cannot directly affect credit score; scores change as a result of other actions), a counterfactual treats the non-actionable features as being collinear with respect to their causal ancestors, regardless of the evaluated posterior. We express these features, through a prior that encodes causal dependencies between features. First, find the distribution of counterfactual explanations $p( \vx' | \vx, y' )$. Then consider a counterfactual as a tuple of causal ancestors and descendants, $\vx' = \begin{pmatrix} c',  e'  \end{pmatrix}^{T}$ in which $e'$ are mutable, non-actionable features and $c'$ are all others. We express mutable, non-actionable features by first marginalizing over $e'$,
\begin{equation}
    \nonumber
    p( c' | x, y' ) = \int_{e'} p\big( \begin{pmatrix} c',  e'  \end{pmatrix}^{T} | y', x \big)\ d e' = \mathcal{N}( \mu', \Lambda^{-1}_{c'} ).
\end{equation}
We then find the weights of the linear model $e' = A c' + b$, and express the mutable, non-actionable features as having come from the conditional distribution, $p( e' | c' ) = \mathcal{N}( e' | A c' + b, \Lambda^{-1}_{z'} )$, where $\Lambda^{-1}_{z'}$ is covariance of the residuals. The updated counterfactual distribution takes the form,
\begin{equation}
    \nonumber
    p( \vx' | \vx, y' ) = p( e' | c' ) p( c' | \vx, y' ).
\end{equation}
For a more thorough evaluation of the causal perspective here, and for a description of how one encodes causal relationships in this framework, see Appendix \ref{app:causal_feasibility}.

\section{Revisiting Counterfactual Optimization}
\label{sec:fashion_mnist}

Up to this point, we have primarily focused on sampling explanations from a known probability distribution, however, it may be helpful to understand our approach in terms of optimizing an objective. Recall the posterior of the counterfactual distribution from Eq.~\eqref{eq:opt_posterior},
\begin{equation*}
       p( \vx' | \vx, y' ) \propto p( y' | \vx' ) p( \vx' | \vx ) p( \vx ).
\end{equation*}
By minimizing the negative log-likelihood of this posterior for our chosen prior, we can express the task of generating counterfactual explanations as optimizing the following objective (Appendix~\ref{app:derivation_our_objective}),
{
\begin{align}
\begin{split}
    \vx' = \arg \min_{\tilde{\vx}} \hspace{3pt} \tilde{\vx}^{T} \Lambda \tilde{\vx} &- 2 \tilde{\vx}^{T} \Lambda \big( ( 1 - \alpha ) \mu + \alpha \vx \big) \\ 
    &+  \gamma || y' - f_{\theta}( \tilde{\vx} ) ||.
    \label{eq:opt_ours}
\end{split}
\end{align}
}
The previously considered norm-ball on the distances used by prior work becomes the mahalanobis distance of samples $\vx'$ from a set of observations with mean, $( 1 - \alpha ) \mu - \alpha \vx$, and covariance, $\Lambda^{-1}$. In other words, we are drawing a line from the mean of the data distribution to the reference, $\vx$ and returning points that have the desired class by sampling $\vx'$ from around a point on this line.

\section{Evaluation}
\label{sec:eval}

In this Section, we evaluate our approach through both a quantitative and qualitative lens. We first compare our proposed approach with several counterfactual generation techniques across a variety of evaluation metrics and datasets. We then investigate its efficacy for more complex image data. We show that the proposed framing encourages explanations to lie further from the decision boundary, so as to produce counterfactuals that are more representative of the ground truth data. We further perform a qualitative evaluation on whether users find explanations across methods satisfying through an Amazon Mechanical Turk Survey.

\subsection{Quantitative Evaluations}
\begin{table*}[tb!]
    \centering
    \scshape
    \resizebox{\textwidth}{!}{
        \setlength{\tabcolsep}{0.35em} 
        \begin{tabular}{c c c c}
        \multicolumn{2}{c}{\scalebox{1.6}{Adult}} & \multicolumn{2}{c}{\scalebox{1.6}{Rice}} \\
        \hline
        \begin{tabular}{c} 
    Method \\
    \hline
    Wachter \\
    Wachter (Ours) \\
    Dice \\
    Dice (Ours) \\
    FACE \\
    FACE (Ours)\\
    Growing Spheres \\
    Growing Spheres (Ours) \\
    CCHVAE
\end{tabular} & \setlength{\tabcolsep}{1em} 
\begin{tabular}{c c c c c c}
    $l_{2}$ & $l_{\infty}$ & \texttt{yNN} & \texttt{Redun.} & \texttt{Div.} & \texttt{t(s)} \\
    \hline
    \rowcolor[gray]{1.}
    0.009 & 0.081 & 0.058 & 4.910 & - & 0.758 \\
    \rowcolor[gray]{.8}
    0.018 & 0.069 & 0.115 & 3.872 & - & 0.280 \\
    \rowcolor[gray]{1.}
    0.028 & 0.088 & 0.137 & 3.703 & 0.101 & 0.007 \\
    \rowcolor[gray]{.8}
    0.039 & 0.123 & 0.191 & 3.870 & 0.121 & 0.650 \\
    \rowcolor[gray]{1.}
    2.201 & 0.891 & 0.612 & 3.789 & - & 22.175 \\
    \rowcolor[gray]{.8}
    2.444 & 0.842 & 0.877 & 4.358 & - & 15.439 \\
    \rowcolor[gray]{1.}
    1.057 & 0.684 & 0.137 & 3.920 & - & 0.002 \\
    \rowcolor[gray]{.8}
    1.553 & 0.694 & 0.159 & 4.400 & - & 0.036 \\
    \rowcolor[gray]{1.}
    0.206 & 0.279 & 1.000 & 9.111 & - & 0.002 \\
\end{tabular}
 &\begin{tabular}{c} 
    Method \\
    \hline
    Wachter \\
    Wachter (Ours) \\
    Dice \\
    Dice (Ours) \\
    FACE \\
    FACE (Ours)\\
    Growing Spheres \\
    Growing Spheres (Ours) \\
    CCHVAE
\end{tabular} &   \\
        \hline \\
        ~\\
        \multicolumn{2}{c}{\scalebox{1.6}{Home Equity Line of Credit}} & \multicolumn{2}{c}{\scalebox{1.6}{Give Me Some Credit}} \\
        \hline
        \begin{tabular}{c} 
    Method \\
    \hline
    Wachter \\
    Wachter (Ours) \\
    Dice \\
    Dice (Ours) \\
    FACE \\
    FACE (Ours)\\
    Growing Spheres \\
    Growing Spheres (Ours) \\
    CCHVAE
\end{tabular} & \setlength{\tabcolsep}{1em} 
\begin{tabular}{c c c c c c} 
    $l_{2}$ & $l_{\infty}$ & \texttt{yNN} & \texttt{Redun.} & \texttt{Div.} & \texttt{t(s)} \\
    \hline
    \rowcolor[gray]{1.}
    0.069 & 0.053 & 0.112 & 1.788 & - & 0.232 \\
    \rowcolor[gray]{.8}
    0.078 & 0.079 & 0.186 & 6.948 & - & 0.048 \\
    \rowcolor[gray]{1.}
    0.059 & 0.082 & 0.161 & 9.869 & 0.104 & 0.477 \\
    \rowcolor[gray]{.8}
    0.088 & 0.112 & 0.340 & 11.465& 0.0854 & 0.506 \\
    \rowcolor[gray]{1.}
    1.054 & 0.701 & 0.729 & 13.863 & - & 2.873 \\
    \rowcolor[gray]{.8}
    1.172 & 0.735 & 0.992 & 16.535 & - & 2.109 \\
    \rowcolor[gray]{1.}
    0.074 & 0.112 & 0.147 & 14.986 & - & 0.003 \\
    \rowcolor[gray]{.8}
    0.086 & 0.120 & 0.152 & 15.243 & - & 0.061 \\
    \rowcolor[gray]{1.}
    1.524 & 0.635 & 0.997 &  & - & 0.489 \\
\end{tabular}
 &\begin{tabular}{c} 
    Method \\
    \hline
    Wachter \\
    Wachter (Ours) \\
    Dice \\
    Dice (Ours) \\
    FACE \\
    FACE (Ours)\\
    Growing Spheres \\
    Growing Spheres (Ours) \\
    CCHVAE
\end{tabular} & \setlength{\tabcolsep}{1em} 
\begin{tabular}{c c c c c c} 
    $l_{2}$ & $l_{\infty}$ & \texttt{yNN} & \texttt{Redun.} & \texttt{Div.} & \texttt{t(s)} \\
    \hline
    \rowcolor[gray]{1.}
    0.006 & 0.018 & 0.289 & 7.558 & - & 0.005 \\
    \rowcolor[gray]{.8}
    0.010 & 0.053 & 0.341 & 7.214 & - & 0.589 \\
    \rowcolor[gray]{1.}
    0.072 & 0.155 & 0.737 & 7.437 & 0.170 & 0.872 \\
    \rowcolor[gray]{.8}
    0.092 & 0.148 & 0.772 & 7.879 & 0.127 & 0.533 \\
    \rowcolor[gray]{1.}
    0.625 & 0.530 & 0.993 & 8.154 & - & 1.792 \\
    \rowcolor[gray]{.8}
    0.668 & 0.542 & 1.000 & 8.399 & - & 3.697 \\
    \rowcolor[gray]{1.}
    0.006 & 0.044 & 0.258 & 7.055 & - & 0.003 \\
    \rowcolor[gray]{.8}
    0.013 & 0.071 & 0.403 & 7.036 & - & 0.148 \\
    \rowcolor[gray]{1.}
    0.491 & 0.467 & 1.000 & 9.401 & - & 0.001 \\
\end{tabular}  \\
        
        \end{tabular}
    }

    \caption{Benchmarking table comparing our proposed counterfactual distance with an l2 distance metric across 4 datasets, showing that while our approach increases runtime, it generates counterfactuals significantly closer to the underlying data distribution as measured by the number of nearest neighbors who share the desired label (YNN), without a significant decrease in performance across any other metric.}
    \label{table:benchmark}
\end{table*}

We use the CARLA \citep{pawelczyk2021carla} counterfactual benchmarking tool in order to compare our proposal with several existing counterfactual generation methods: 
\begin{itemize}
    \item \texttt{Wachter} \citep{wachter2017counterfactual}, which optimizes Eq. \eqref{eq:opt}.
    \item \texttt{DiCE} \citep{mothilal2020explaining}, which adds a diversity regularizer to Eq. \eqref{eq:opt} to generate a large, diverse set of counterfactuals at once. For this evaluation, we generate $3$ counterfactuals per reference point.
    \item \texttt{FACE} \citep{poyiadzi2020face}, which chooses counterfactuals by traversing a nearest-neighbor graph over the observed data, until reaching an instance that has the desired label.
    \item \texttt{Growing Spheres} \citep{laugel2017inverse}, which iteratively samples an expanding set of points around a given reference until a sample lies across the decision boundary.
    \item \texttt{CCHVAE} \citep{pawelczyk2020learning}, which uses a variational autoencoder (VAE) to estimate the generative process for a given instance, and returns counterfactuals by sampling within the $l_{p}$ sphere around a reference in the latent space.
\end{itemize}

In order to compare against our approach, we replace the distance metric in \texttt{Wachter}, \texttt{DiCE}, \texttt{Growing Spheres} and \texttt{FACE} with ours in Eq. \eqref{eq:opt_ours}. We designate this choice of the distance metric with the identifier \texttt{(Ours)} in Table \ref{table:benchmark}. Additionally, as our approach is dependent on the underlying data distribution, we include a comparison against \texttt{CCHVAE} in order to evaluate the effectiveness of a method that traverses a learned latent space, rather than staying within the feature sapce. Each method's parameters were chosen independently via a grid search that sought to find the parameters that minimize the $l_{2}$ distance to the reference, while ensuring that the method generates counterfactuals of the desired class with at least $99\%$ success rate.

We generate $3000$ counterfactuals for every method across each dataset and evaluate different methods over five metrics (See \citep{pawelczyk2021carla} for more information on the specifics of how these metrics are calculated.)

\begin{itemize}
    \item $\boldsymbol{l_{2}}$, the average $l_{2}$ distance between the generated counterfactuals and the reference.
    
    \item $\boldsymbol{l_{\infty}}$, the average $l_{\infty}$ distance between the generated counterfactuals and the reference.

    \item \textsc{\bf yNN}, the number of nearest neighbors with the desired label. Based on a desideratum formulated by \citep{laugel2019dangers}, a desirable property of counterfactuals is that they lie close to observed data that has the desired label. This metric captures this property by finding the proportion of a counterfactual's $K$ nearest neighbors in the observed data that have the desired label (here, we set $K=5$).

    \item \textsc{\bf Redundancy}, the number of features for a given counterfactual that can be changed back to the reference value without changing the counterfactual class (i.e., the number of unnecessary changes wrt. the classifier's predicted output).

    \item \textsc{\bf Diversity}, the diversity of the generated counterfactuals based on the metric defined in \citep{mothilal2020explaining}.
    
    \item \textsc{\bf T(S)}, the average number of seconds required for a method to generate a single counterfactual.
\end{itemize}

\subsubsection{Results}

In nearly all cases, using the metric in Eq. \eqref{eq:opt_ours} encourages counterfactuals to sit more closely to the region of the feature space for which their neighbors have the desired predicted class (i.e. increases \textsc{\bf yNN}). We see this effect regardless of the method used. 

Moreover, we see that our method generally increases the euclidean distance to the reference. This is expected behavior as we are comparing against methods that explicitly optimize for this metric. Yet, despite our approach not improving over the alternatives for this metric, we find that our approach is not significantly worse in terms of $l_{2}$ distance. Using the objective in equation \eqref{eq:opt_ours} effectively gives up a small degree of $l_{2}$ similarity in order to encourage counterfactuals that are more clear examples of the desired class.

We also see except in the case of the \textit{Home Equity Line of Credit} dataset, generating counterfactuals according to \citet{wachter2017counterfactual}, we decrease the number of unnecessary features changed from the reference (i.e. \textsc{Redundancy}). However, when adding the diversity regularizer from \citet{mothilal2020explaining}, we lose this benefit. Upon further investigation for this specific case, we find that the distribution is highly anisotropic; there is a very large difference between the largest and smallest eigenvalues, 2 orders of magnitude larger than any other considered dataset. The principal axis as defined by the eigenvalues of the covariance matrix is also not particularly informative for the classifier. Thus in order to maintain faithfulness to the original distribution, the counterfactuals change along the minor axes. This encourages changes to a large number of features, only some of which are necessary for crossing the decision boundary.

Outside of the case of \citet{wachter2017counterfactual}, we find that applying additional regularizers encourages our method to change a larger number of redundant features than the alternative. For similar reasons to the \textit{Home Equity Line of Credit} dataset above, applying a diversity regularizer with our proposed approach encourages points to be distinct from one another. This puts a greater emphasis on the minor axes as defined by the eigenvalues of the covariance matrix and in turn encourages more redundant changes as the number of counterfactuals generated by \textsc{DiCE} increases. Similarly in the case of \textsc{FACE}, the nearest neighbor to a point as defined by the Mahalanobis Distance in equation \eqref{eq:opt_ours}, will define nearby points as those with small changes along the principal axes of the data. If the principal axis is uniformative for the classifier, the method will traverse along the minor axes. As in the previous cases, this more quickly builds up small changes to a counterfactual, increasing the number of redundant features changed from the reference.



\subsection{Qualitative Evaluation}
\begin{figure*}[t!]
    \centering
    \resizebox{\textwidth}{!}{ 
        \begin{tblr}{ c | c | c c c c c c c c c c }
             \scalebox{7.5}{Reference} & \hspace{1cm}\scalebox{7.5}{Method}\hspace{1cm} & \scalebox{7.5}{T-shirt} & \scalebox{7.5}{Trousers} & \scalebox{7.5}{Pullover} & \scalebox{7.5}{Dress} 
               & \scalebox{7.5}{Coat} & \scalebox{7.5}{Sandal} & \scalebox{7.5}{Shirt} & \scalebox{7.5}{Sneaker} 
               & \scalebox{7.5}{Bag} & \scalebox{7.5}{Ankle boot}
             ~\\
             \hline
             ~\\
             \SetCell[r=3]{m}{} \includegraphics[width=0.8\textwidth]{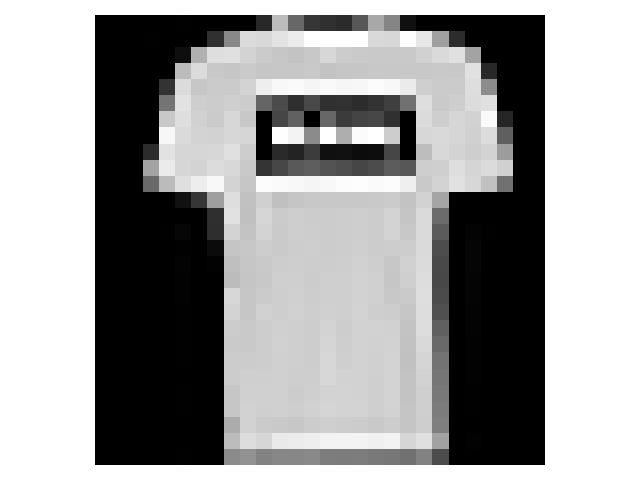} &
             ~
             \raisebox{1.5\height}{\scalebox{7.5}{Ours}} &
             \includegraphics[width=0.8\textwidth]{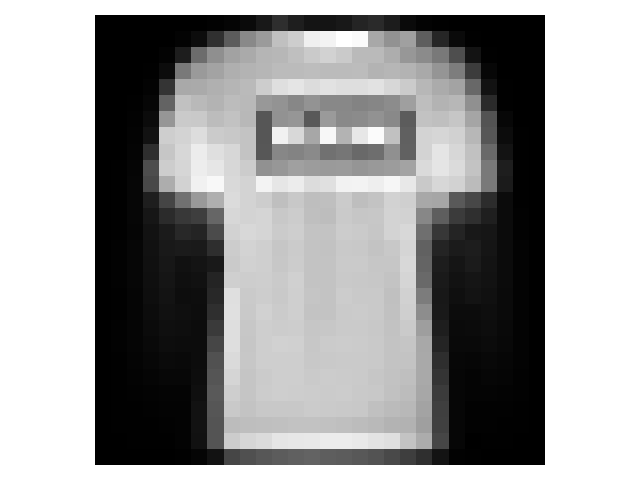} &
             \includegraphics[width=0.8\textwidth]{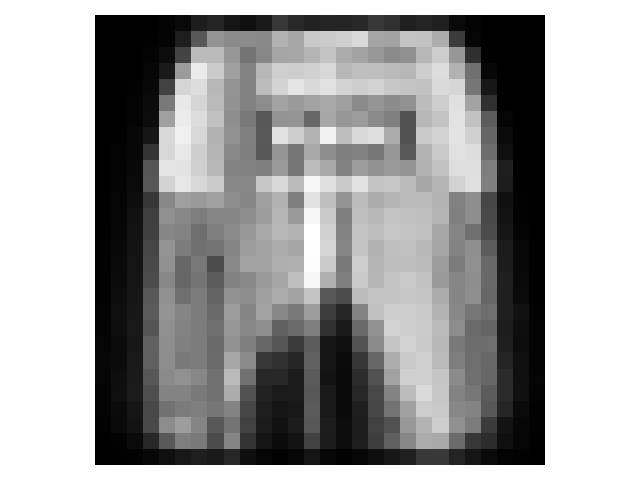} &
             \includegraphics[width=0.8\textwidth]{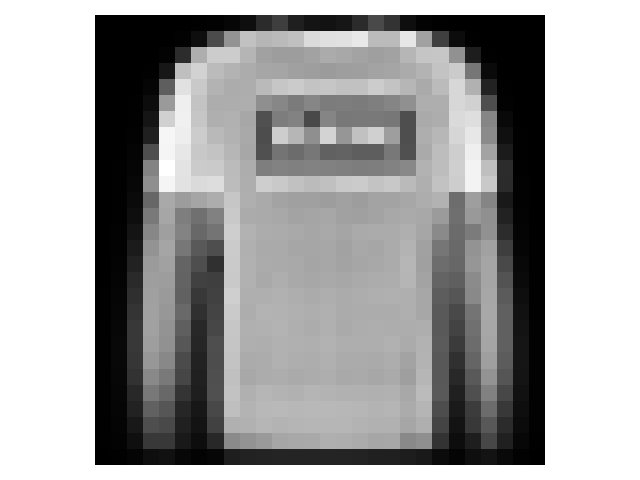} &
             \includegraphics[width=0.8\textwidth]{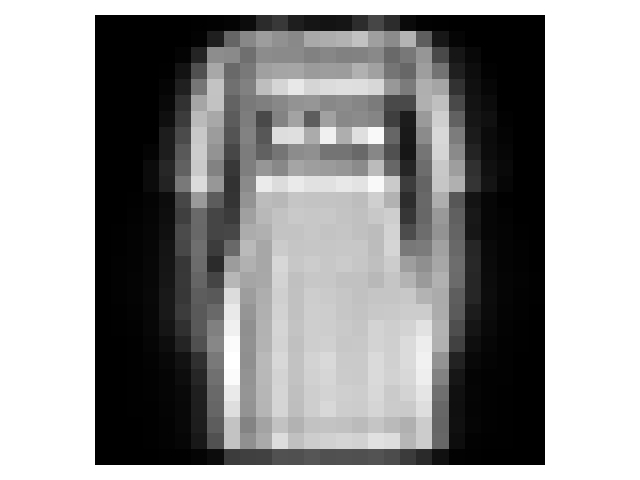} &
             \includegraphics[width=0.8\textwidth]{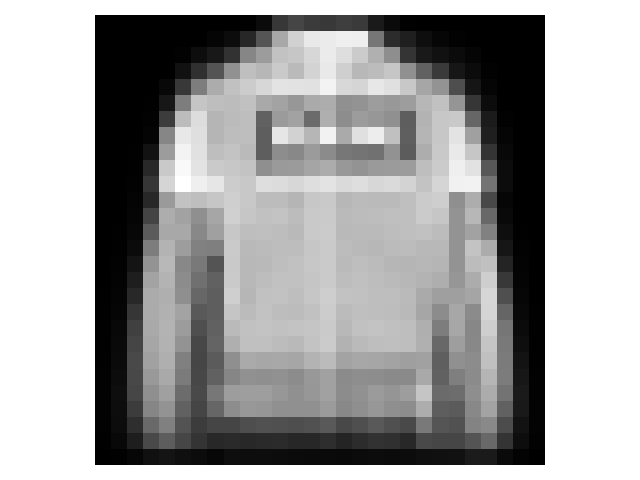} &
             \includegraphics[width=0.8\textwidth]{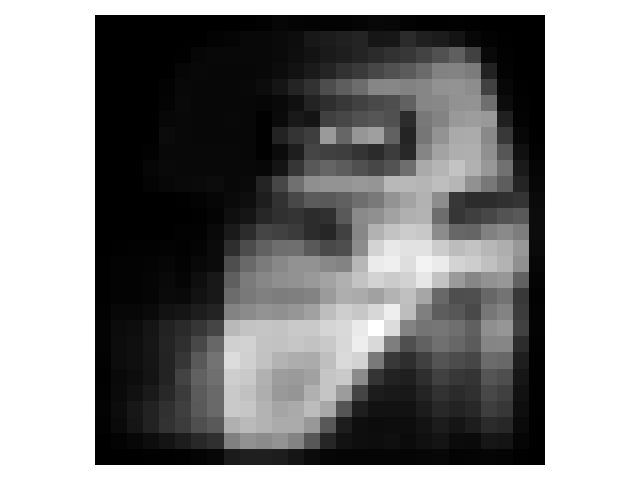} &
             \includegraphics[width=0.8\textwidth]{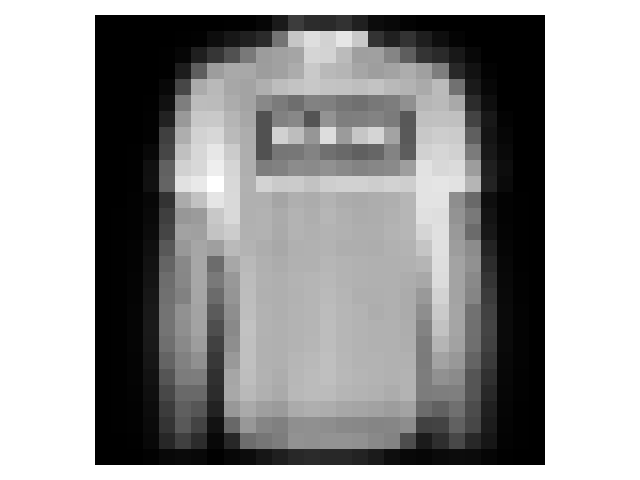} &
             \includegraphics[width=0.8\textwidth]{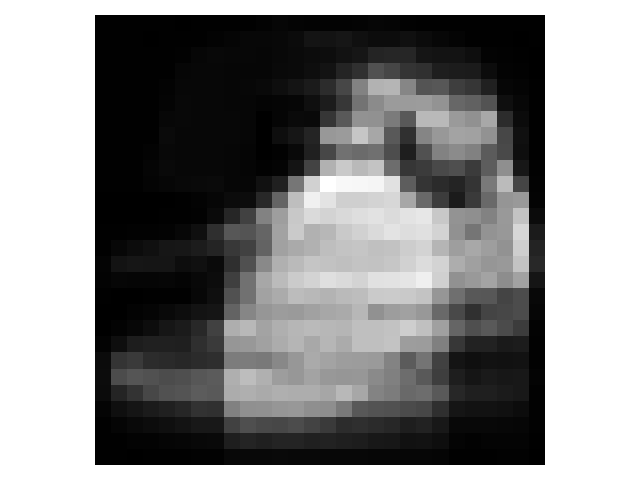} &
             \includegraphics[width=0.8\textwidth]{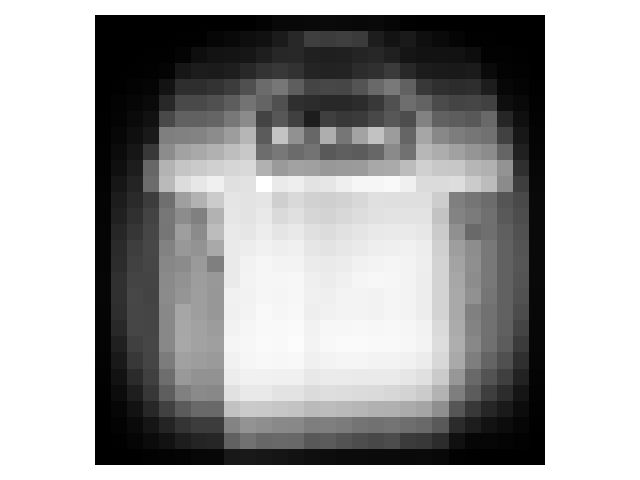} &
             \includegraphics[width=0.8\textwidth]{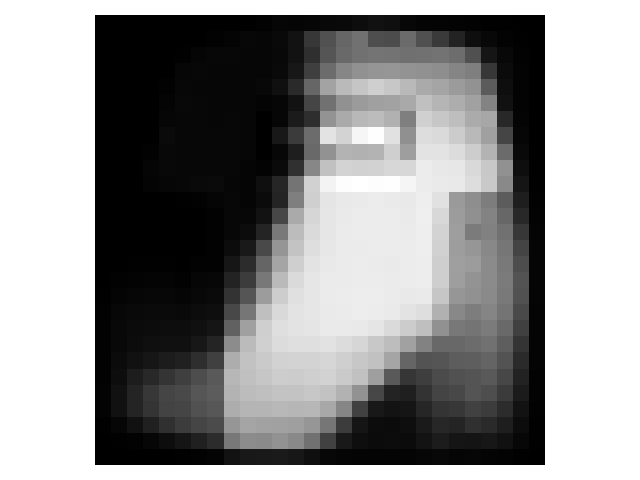}
             ~\\
             ~ &
             \raisebox{1.5\height}{\scalebox{7.5}{L2}} &
             \includegraphics[width=0.8\textwidth]{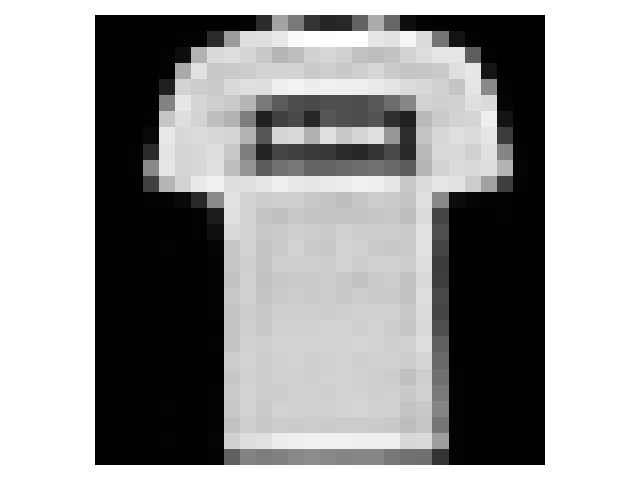} &
             \includegraphics[width=0.8\textwidth]{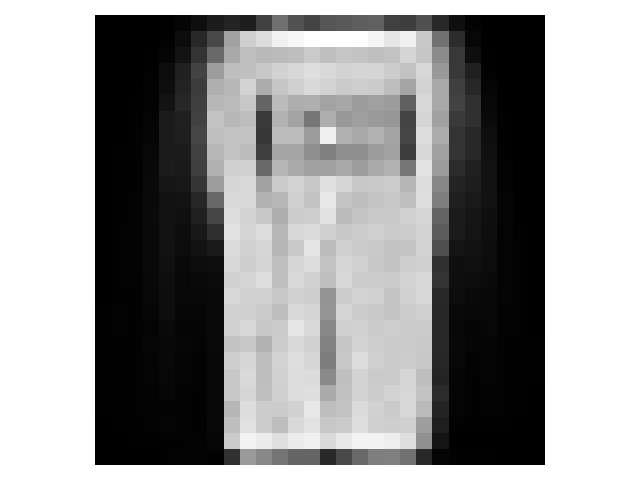} &
             \includegraphics[width=0.8\textwidth]{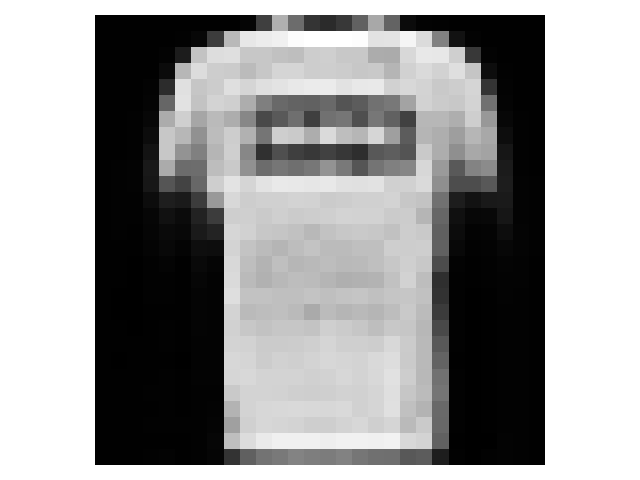} &
             \includegraphics[width=0.8\textwidth]{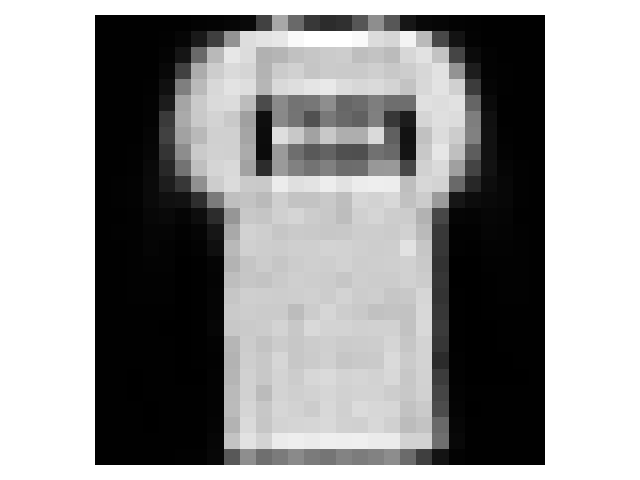} &
             \includegraphics[width=0.8\textwidth]{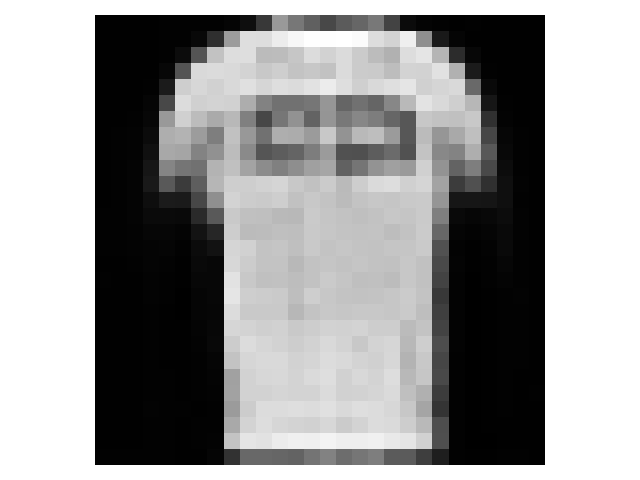} &
             \includegraphics[width=0.8\textwidth]{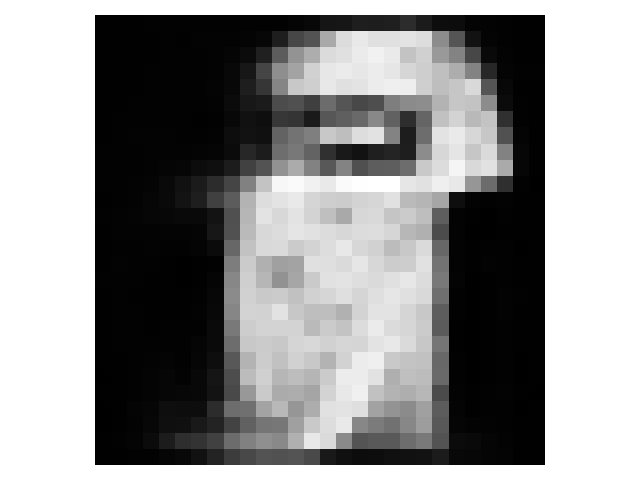} &
             \includegraphics[width=0.8\textwidth]{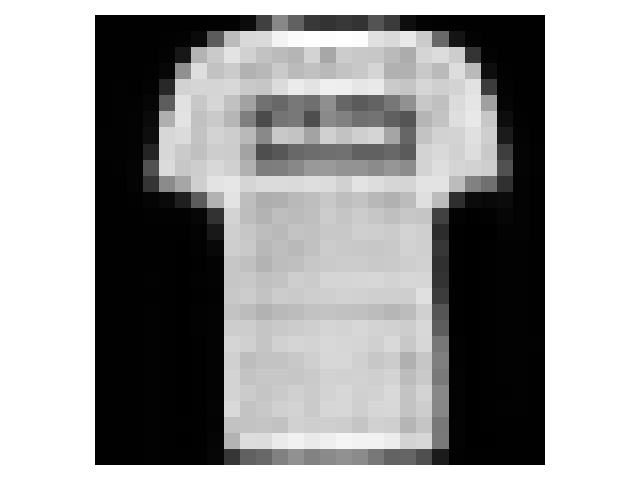} &
             \includegraphics[width=0.8\textwidth]{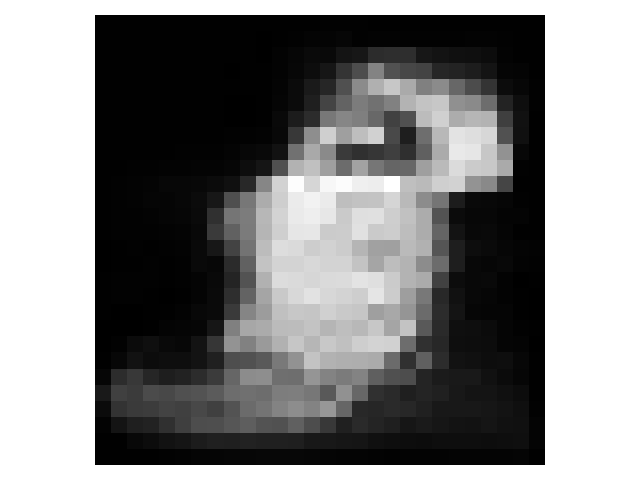} &
             \includegraphics[width=0.8\textwidth]{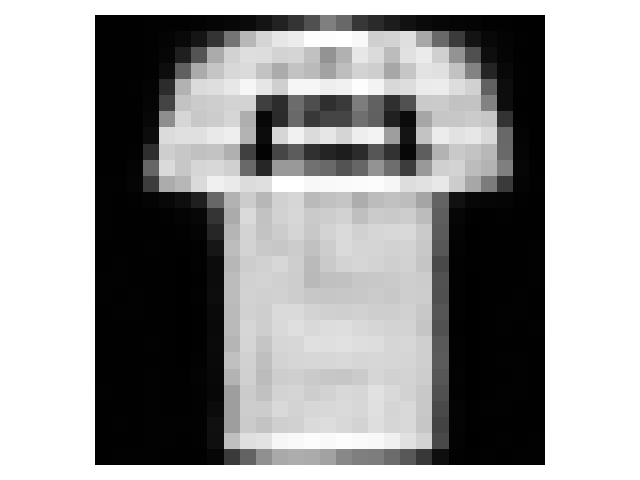} &
             \includegraphics[width=0.8\textwidth]{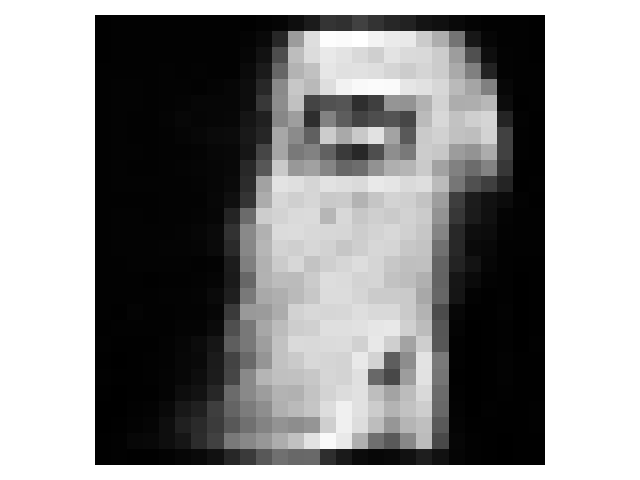}
             ~\\
             ~ & 
             \raisebox{1.5\height}{\scalebox{7.5}{VAE}} &
             \includegraphics[width=0.8\textwidth]{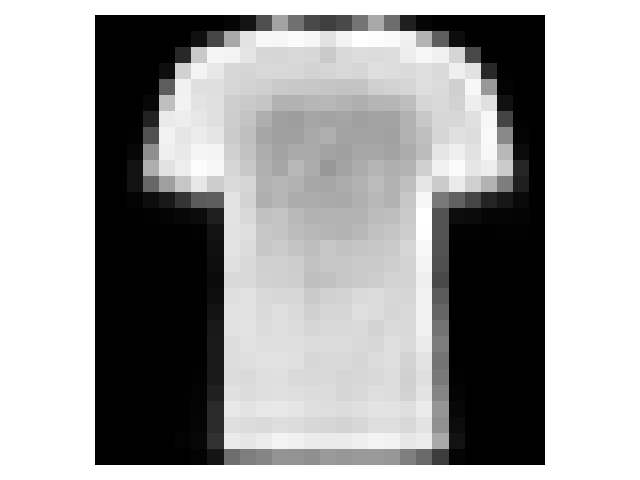} &
             \includegraphics[width=0.8\textwidth]{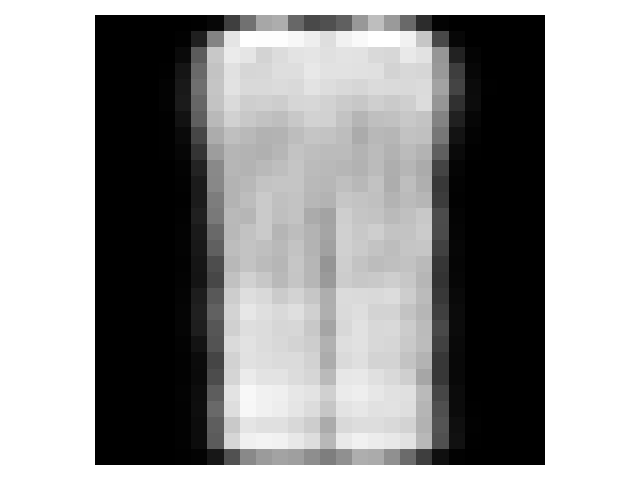} &
             \includegraphics[width=0.8\textwidth]{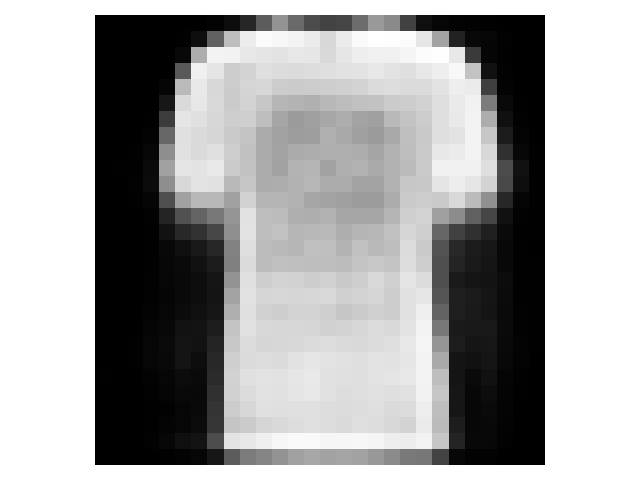} &
             \includegraphics[width=0.8\textwidth]{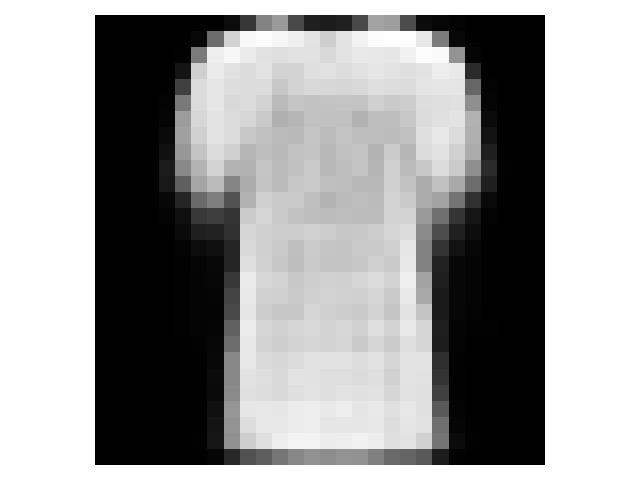} &
             \includegraphics[width=0.8\textwidth]{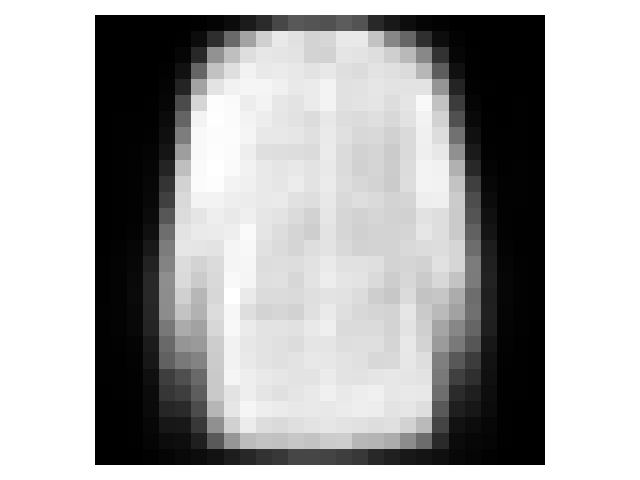} &
             \includegraphics[width=0.8\textwidth]{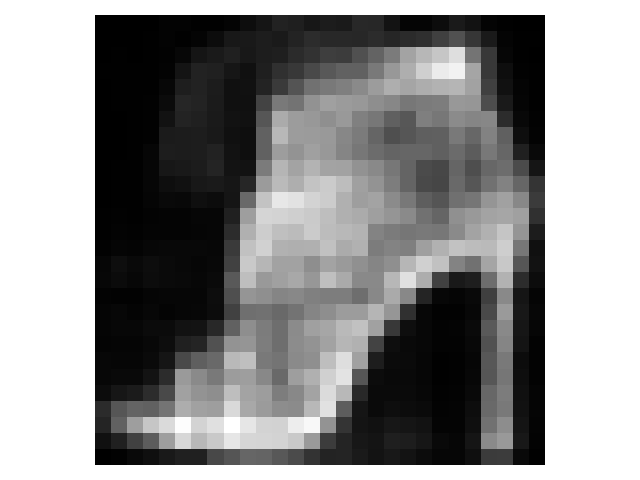} &
             \includegraphics[width=0.8\textwidth]{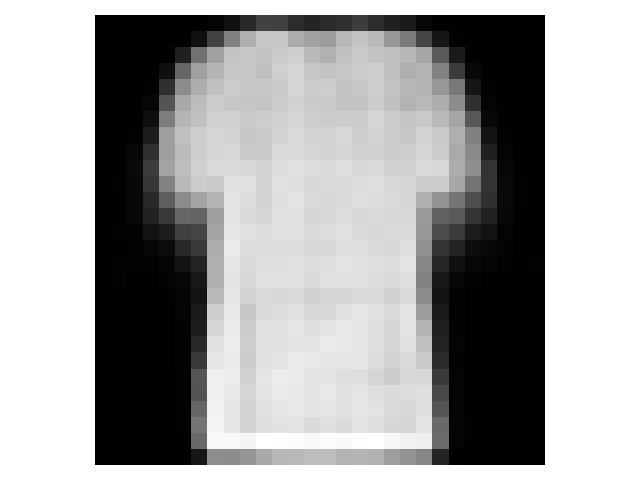} &
             \includegraphics[width=0.8\textwidth]{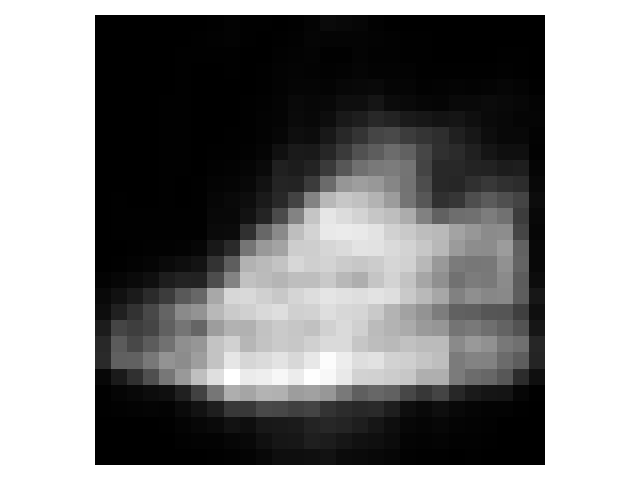} &
             \includegraphics[width=0.8\textwidth]{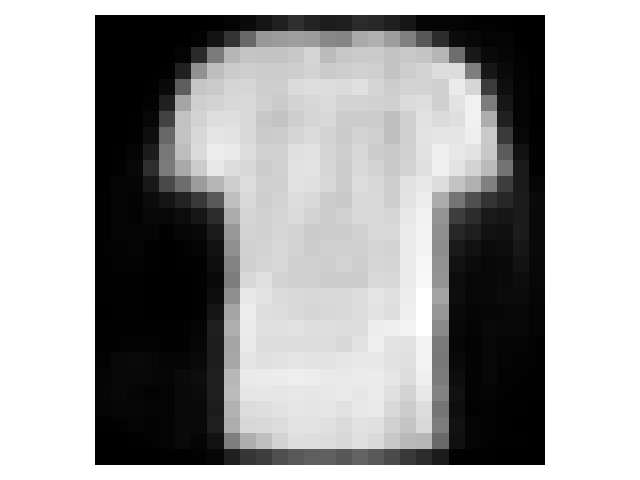} &
             \includegraphics[width=0.8\textwidth]{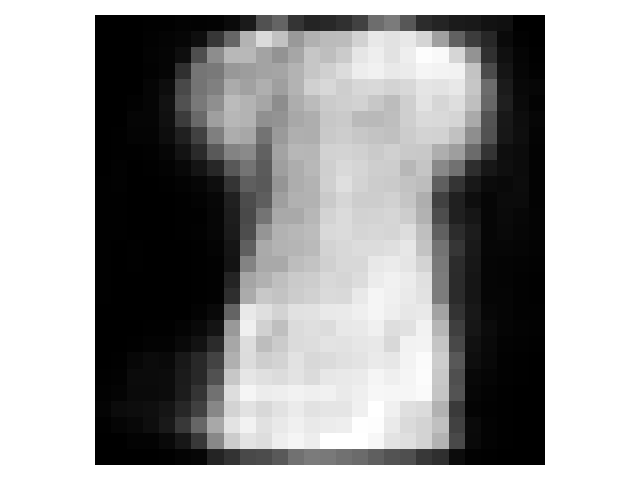}
             ~
             ~
             ~
             ~\\
             \hline
             ~\\
             \SetCell[r=3]{m}{} \includegraphics[width=0.8\textwidth]{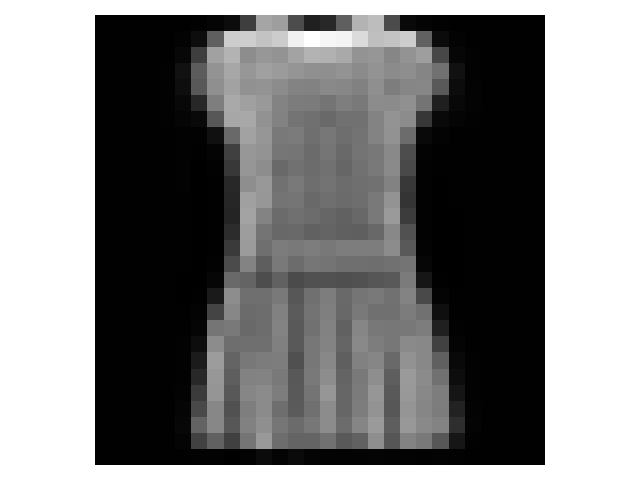} &
             ~         
             \raisebox{1.5\height}{\scalebox{7.5}{Ours}} &
             \includegraphics[width=0.8\textwidth]{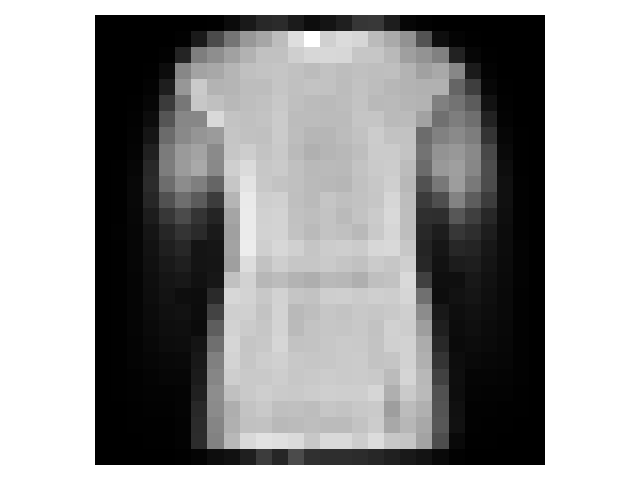} &
             \includegraphics[width=0.8\textwidth]{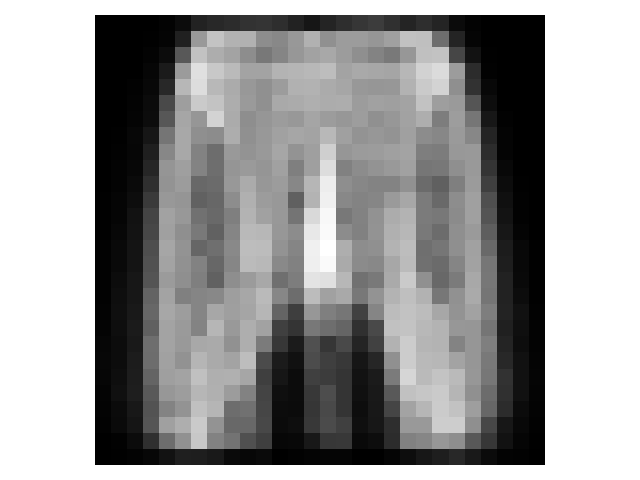} &
             \includegraphics[width=0.8\textwidth]{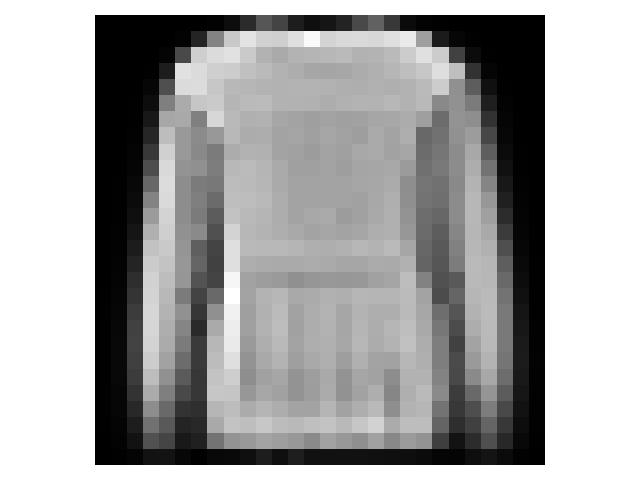} &
             \includegraphics[width=0.8\textwidth]{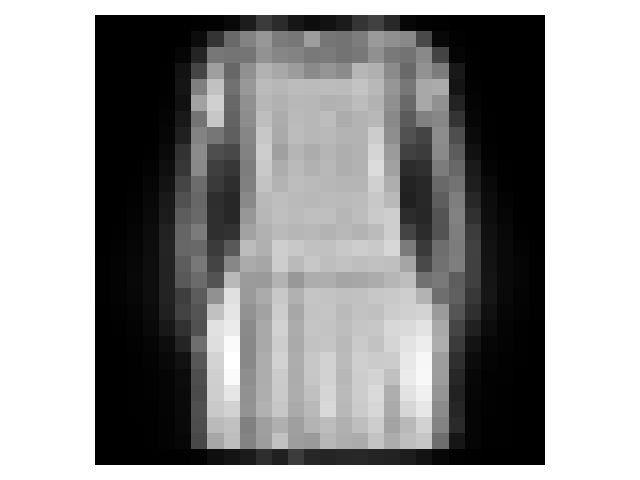} &
             \includegraphics[width=0.8\textwidth]{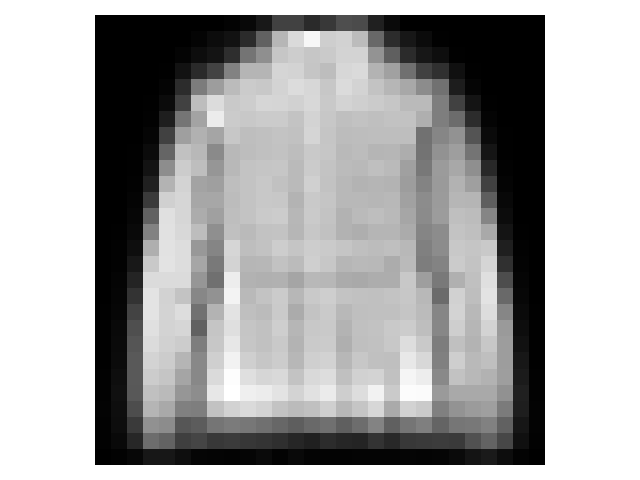} &
             \includegraphics[width=0.8\textwidth]{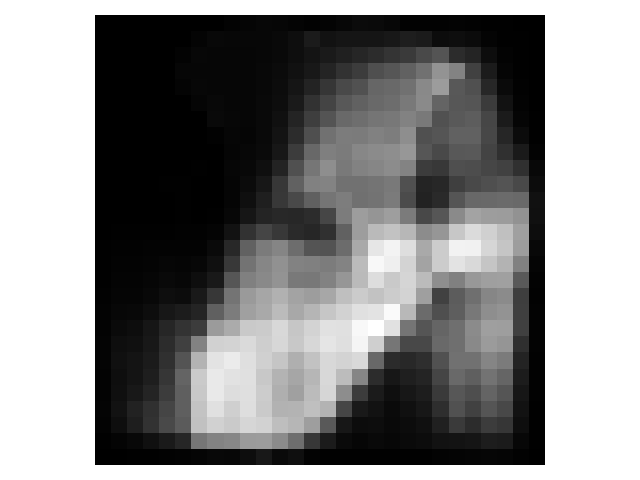} &
             \includegraphics[width=0.8\textwidth]{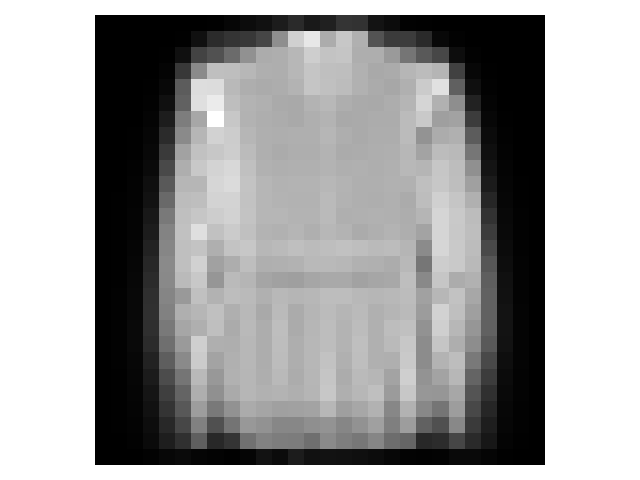} &
             \includegraphics[width=0.8\textwidth]{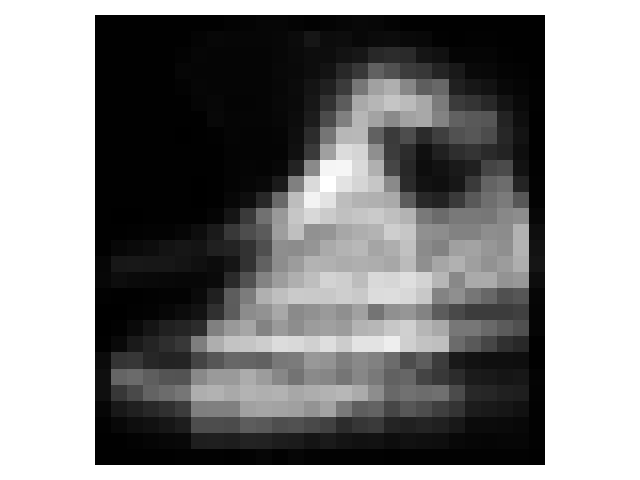} &
             \includegraphics[width=0.8\textwidth]{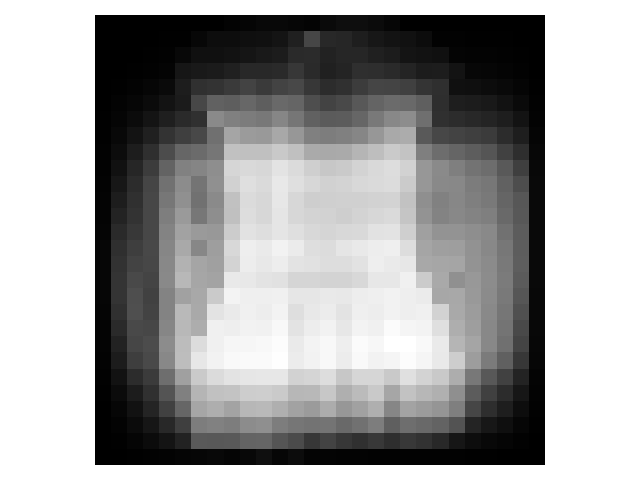} &
             \includegraphics[width=0.8\textwidth]{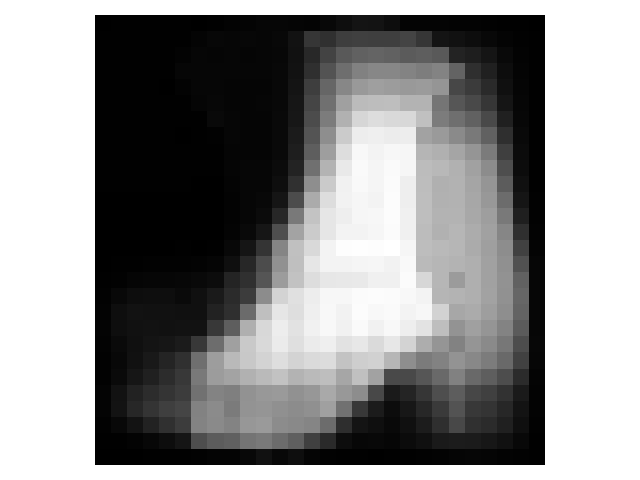}
             ~\\
             ~ &
             \raisebox{1.5\height}{\scalebox{7.5}{L2}} &
             \includegraphics[width=0.8\textwidth]{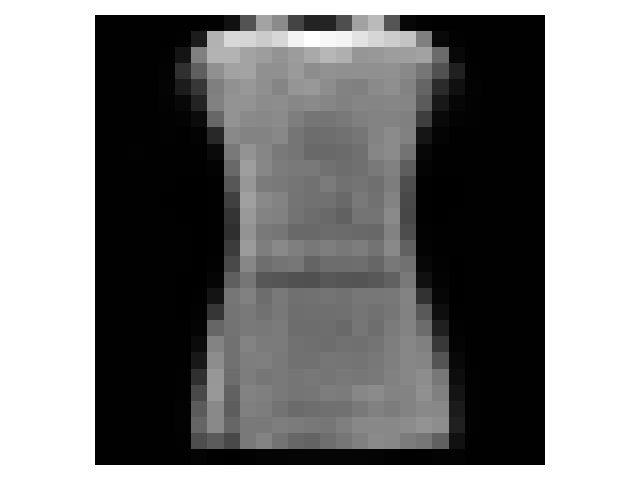} &
             \includegraphics[width=0.8\textwidth]{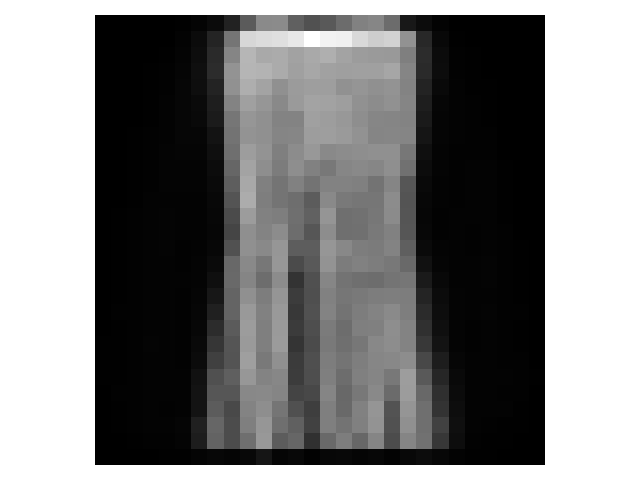} &
             \includegraphics[width=0.8\textwidth]{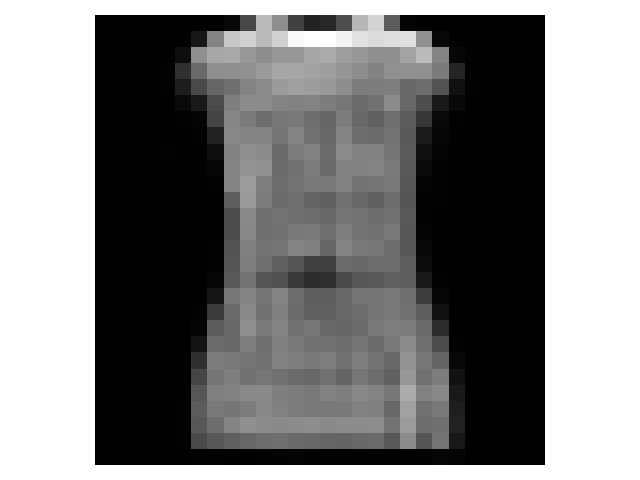} &
             \includegraphics[width=0.8\textwidth]{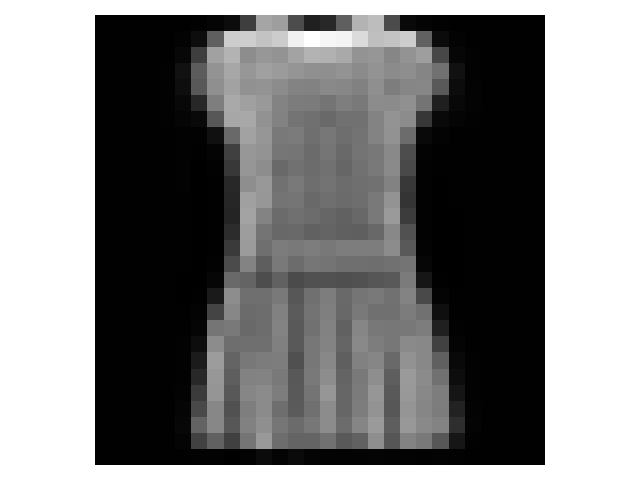} &
             \includegraphics[width=0.8\textwidth]{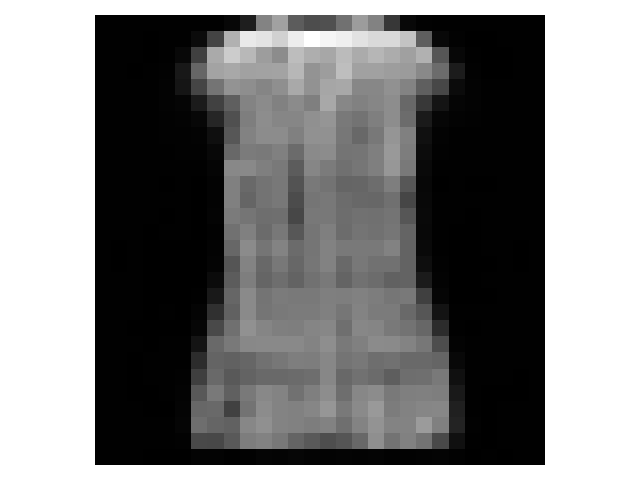} &
             \includegraphics[width=0.8\textwidth]{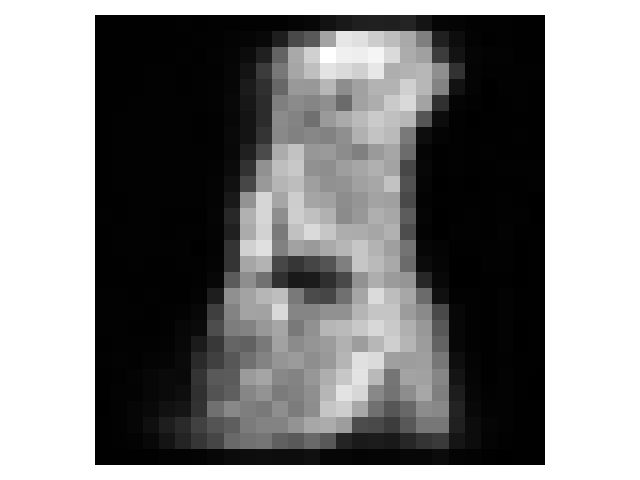} &
             \includegraphics[width=0.8\textwidth]{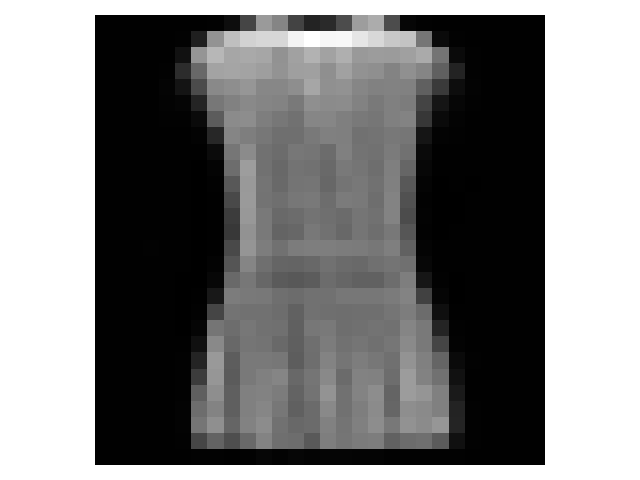} &
             \includegraphics[width=0.8\textwidth]{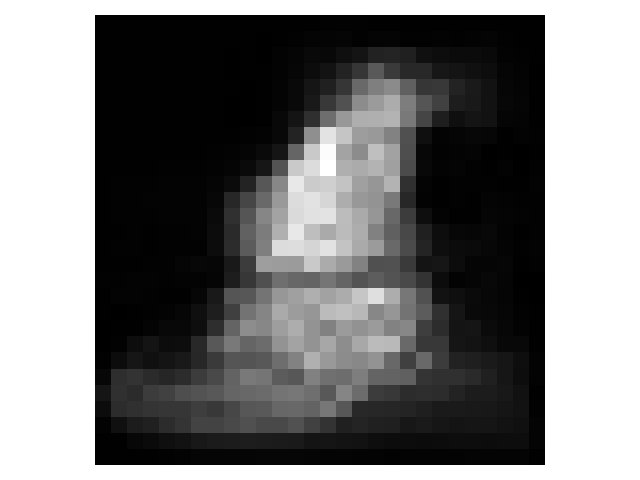} &
             \includegraphics[width=0.8\textwidth]{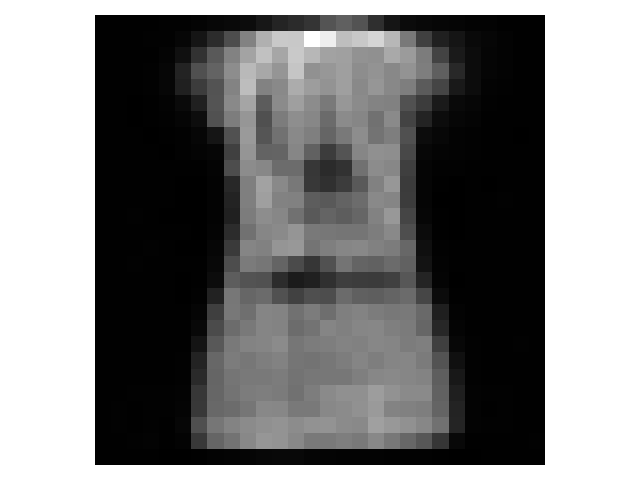} &
             \includegraphics[width=0.8\textwidth]{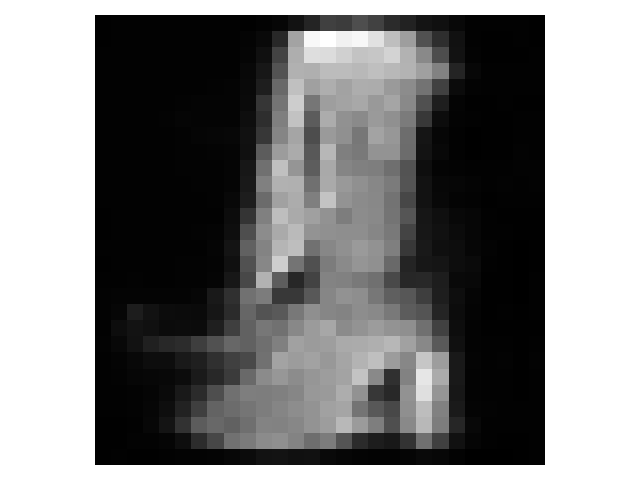}
             ~\\
             ~ &
             \raisebox{1.5\height}{\scalebox{7.5}{VAE}} &
             \includegraphics[width=0.8\textwidth]{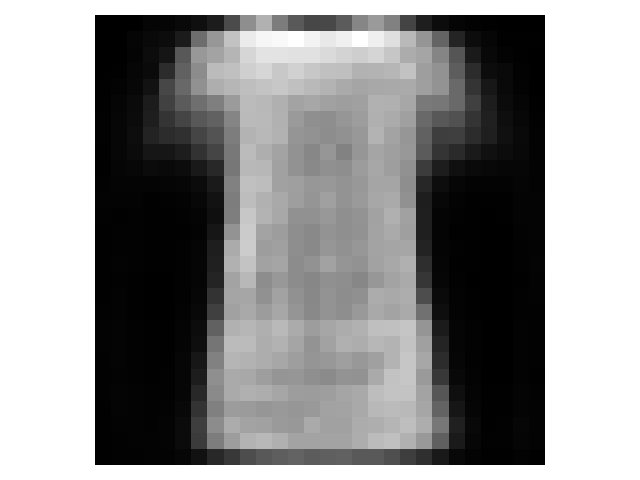} &
             \includegraphics[width=0.8\textwidth]{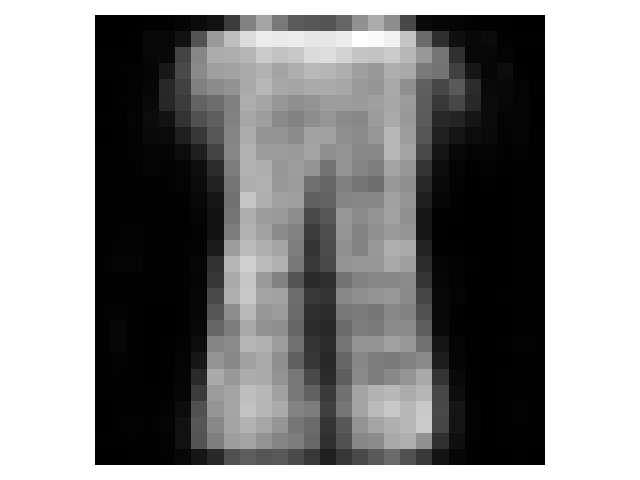} &
             \includegraphics[width=0.8\textwidth]{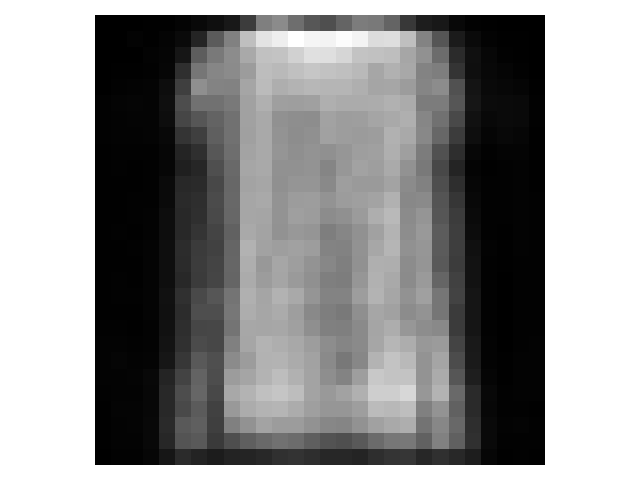} &
             \includegraphics[width=0.8\textwidth]{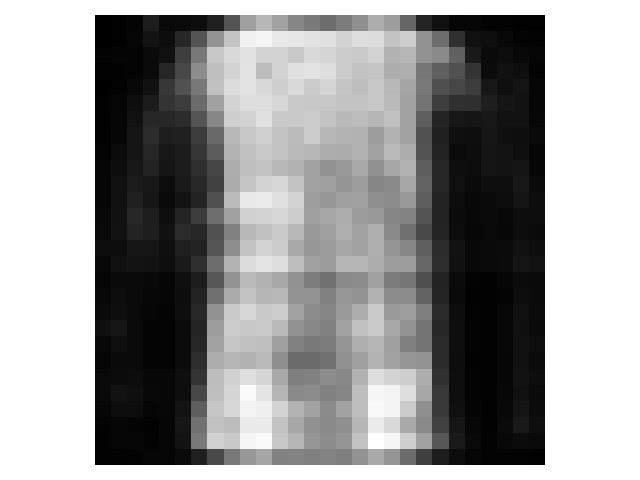} &
             \includegraphics[width=0.8\textwidth]{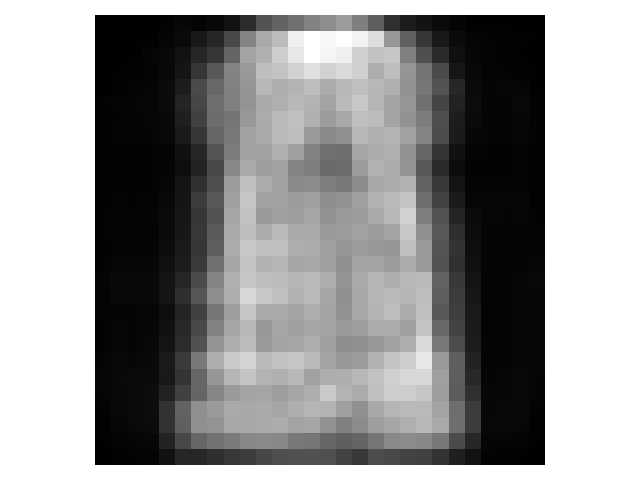} &
             \includegraphics[width=0.8\textwidth]{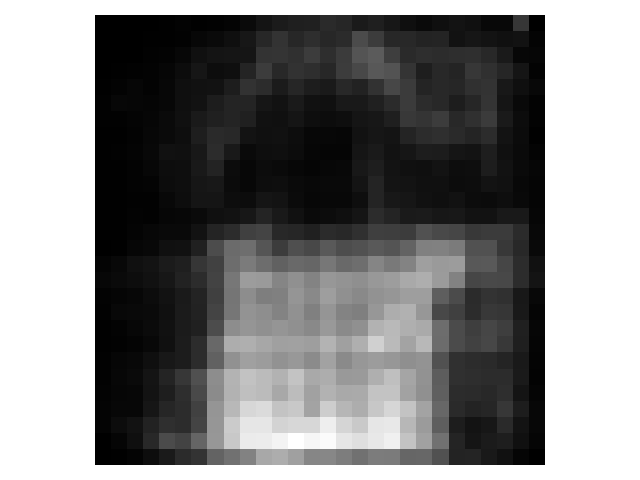} &
             \includegraphics[width=0.8\textwidth]{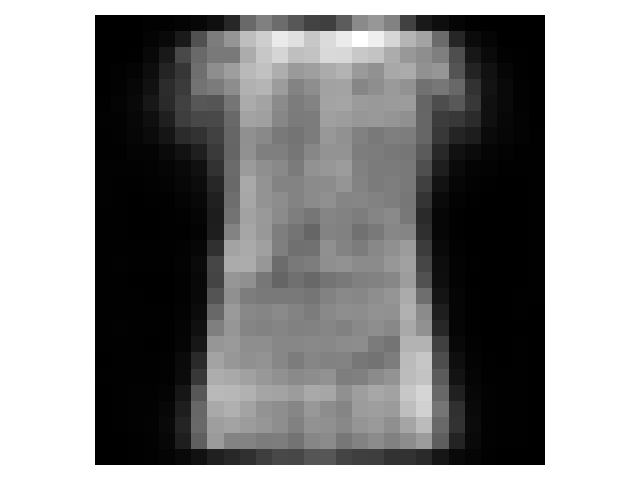} &
             \includegraphics[width=0.8\textwidth]{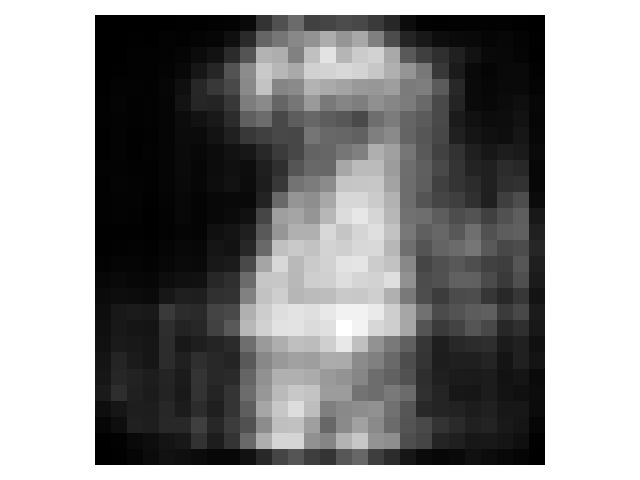} &
             \includegraphics[width=0.8\textwidth]{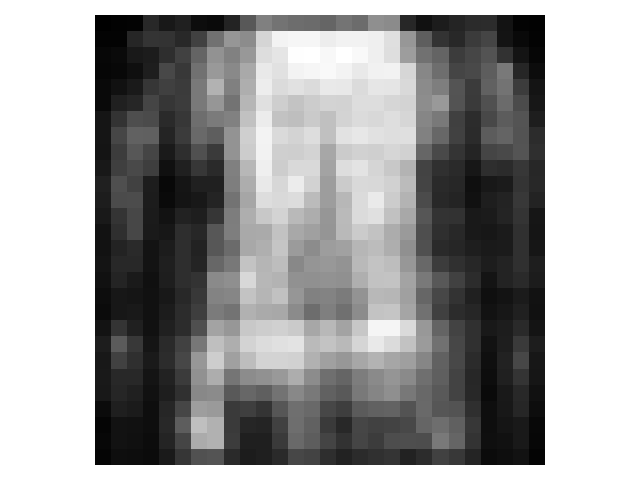} &
             \includegraphics[width=0.8\textwidth]{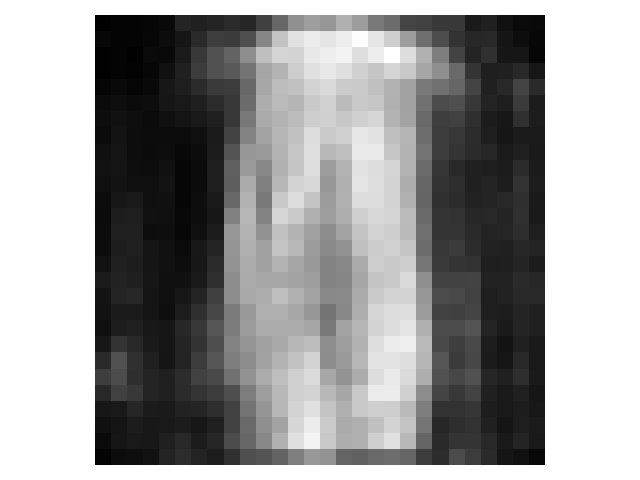}
             
        \end{tblr}
    }
    \caption{All images have 99\% certainty for the desired class based on the trained classifier. Our proposed approach produces counterfactual images that, while further from the reference images than those generated using the L2 distance, exhibit more semantically meaningful features associated with each class. Additionally, our approach avoids the class mixing observed when traversing the VAE's latent space.}
    \label{fig:fashion_mnist_all}
\end{figure*}

In Figure \ref{fig:fashion_mnist_all}, we show how this new objective changes counterfactuals as compared to the \citet{wachter2017counterfactual} objective in Eq.~\eqref{eq:opt} wrt. euclidean distance. We generate a variety of counterfactuals for the Fashion MNIST dataset \cite{xiao2017/online} and focus solely on the implications of the change in the underlying graphical model by comparing the distance metric used in Eq.~\eqref{eq:opt} to the metric used in Eq.~\eqref{eq:opt_ours}. In addition, we compare to counterfactuals generated by a variational autoencoder, by finding counterfactuals by traversing the learned latent space. While a great deal of work has built on Eq.~\eqref{eq:opt} via a variety of different approaches, these techniques and recommendations still apply under our recommended mahalanobis distance. We show how our prior changes the baseline for generating counterfactuals. 

\xhdr{Fashion MNIST Counterfactual Explanations}

In order to generate the images in Figure \ref{fig:fashion_mnist_all}, we train a simple neural network, $f_{\theta}: \mathcal{X} \rightarrow \{0,1\}^{10}$ to classify articles of clothing from Fashion MNIST. Our training pipeline is included in Appendix \ref{app:reproducibility}.

While not a dataset that one traditionally treats as Gaussian, we map Fashion MNIST into our setting by applying a logit transform, $\log( \frac{| x - \epsilon|}{ 1 - | x - \epsilon |} )$ to the grayscaled images and express the data distribution's mean and covariance as the mean and covariance of the dataset's logits. In order to ensure that the covariance matrix is non-singular, we apply a small degree of Gaussian noise to each of the pixel logits. 

Figure \ref{fig:fashion_mnist_all} shows that our approach encourages semantically meaningful changes to the reference images. For example, the \emph{Bag $\rightarrow$ T-Shirt} counterfactual using $l_{2}$ distance provides a noisy sleeve outline, however, the distance function entailed by our approach introduces a clear set of sleeves. 
As we allow explanations to stray further from the reference and closer to the desired class ($\alpha = 0.3$), rather than finding explanations that move out of the distribution and become adversarial, we instead introduced more nuanced changes that bring us closer to the prototypical form for the desired class. For example, consider counterfactual \emph{Shirt $\rightarrow$ Pullover}, pullovers generally have longer sleeves than torsos; decreasing $\alpha$ subtly shortens the waist.

In Appendix \ref{app:mnist} we show a similar comparison for the standard MNIST dataset. Appendix \ref{app:pets} further compares a more complex dataset for classifying RGB images of pets.
\subsection{Survey Evaluation}

\begin{figure*}[t]
    \centering
    \subfloat[German Credit
      \label{fig:sfig1}]{%
      \includegraphics[width=0.23\textwidth]{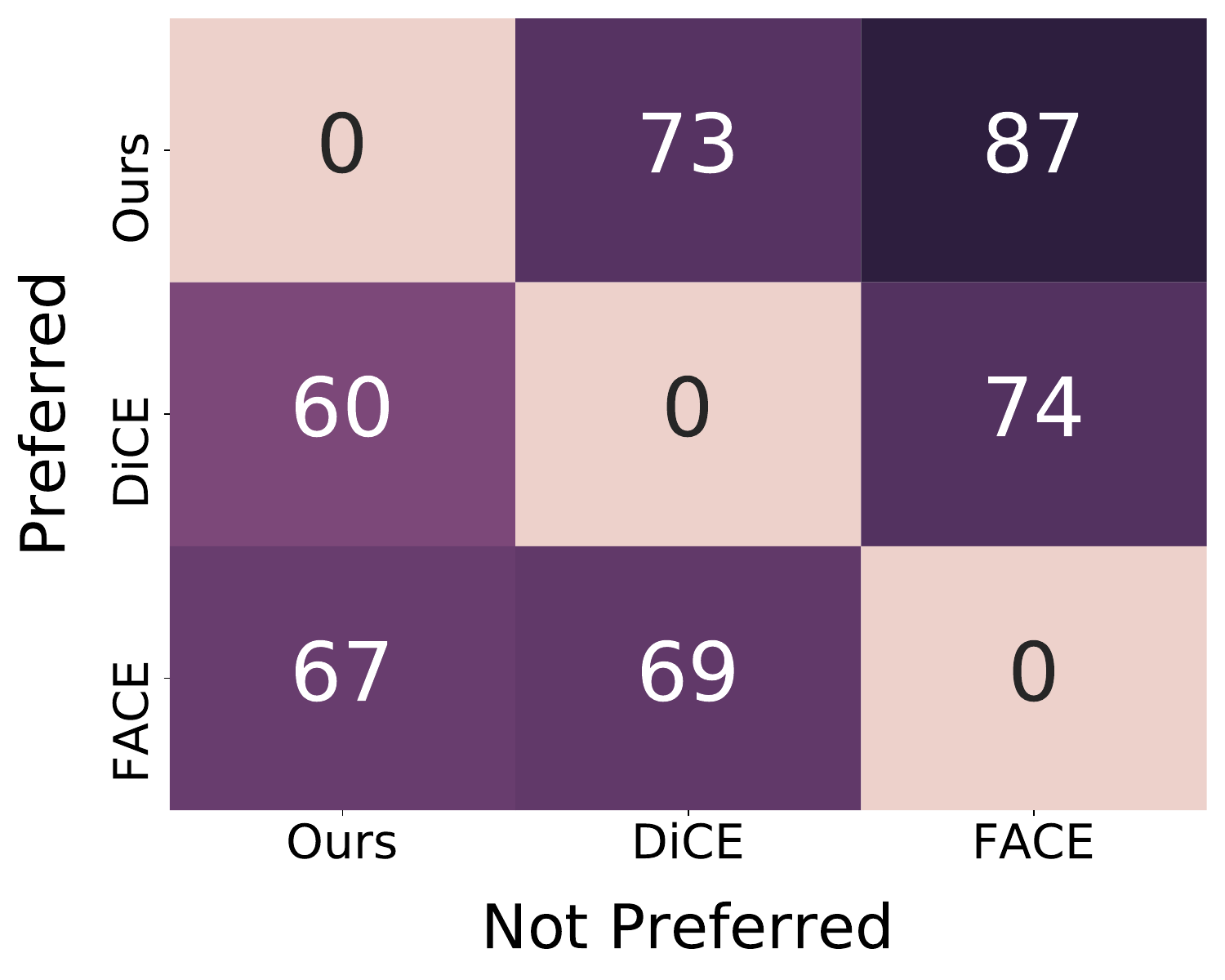}
    }
    \hspace{0.5in}
    \subfloat[LUCAS0
      \label{fig:sfig1}]{%
      \includegraphics[width=0.23\textwidth]{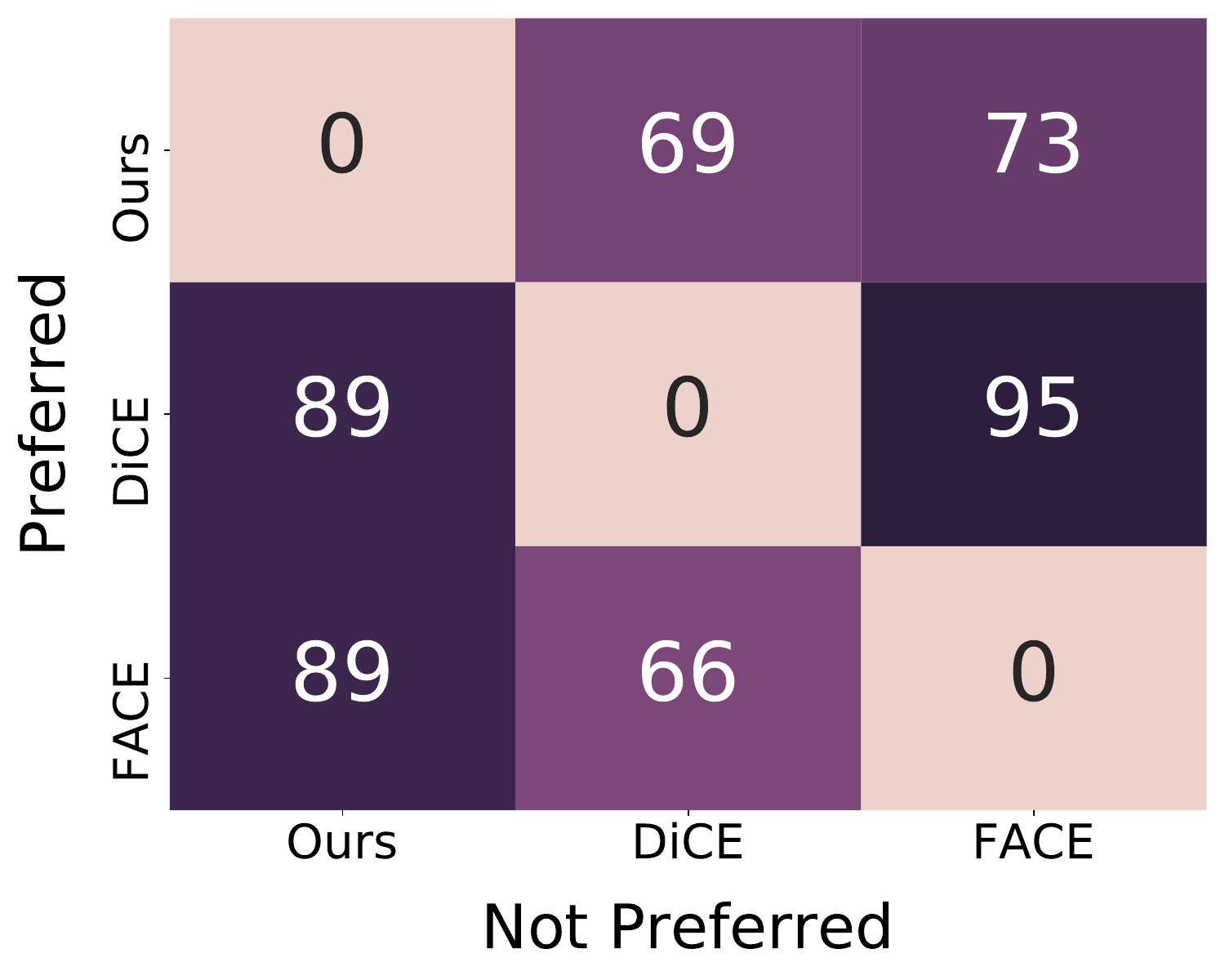}
    }
    \hspace{0.5in}
    \subfloat[Adult
      \label{fig:sfig2}]{%
      \includegraphics[width=0.23\textwidth]{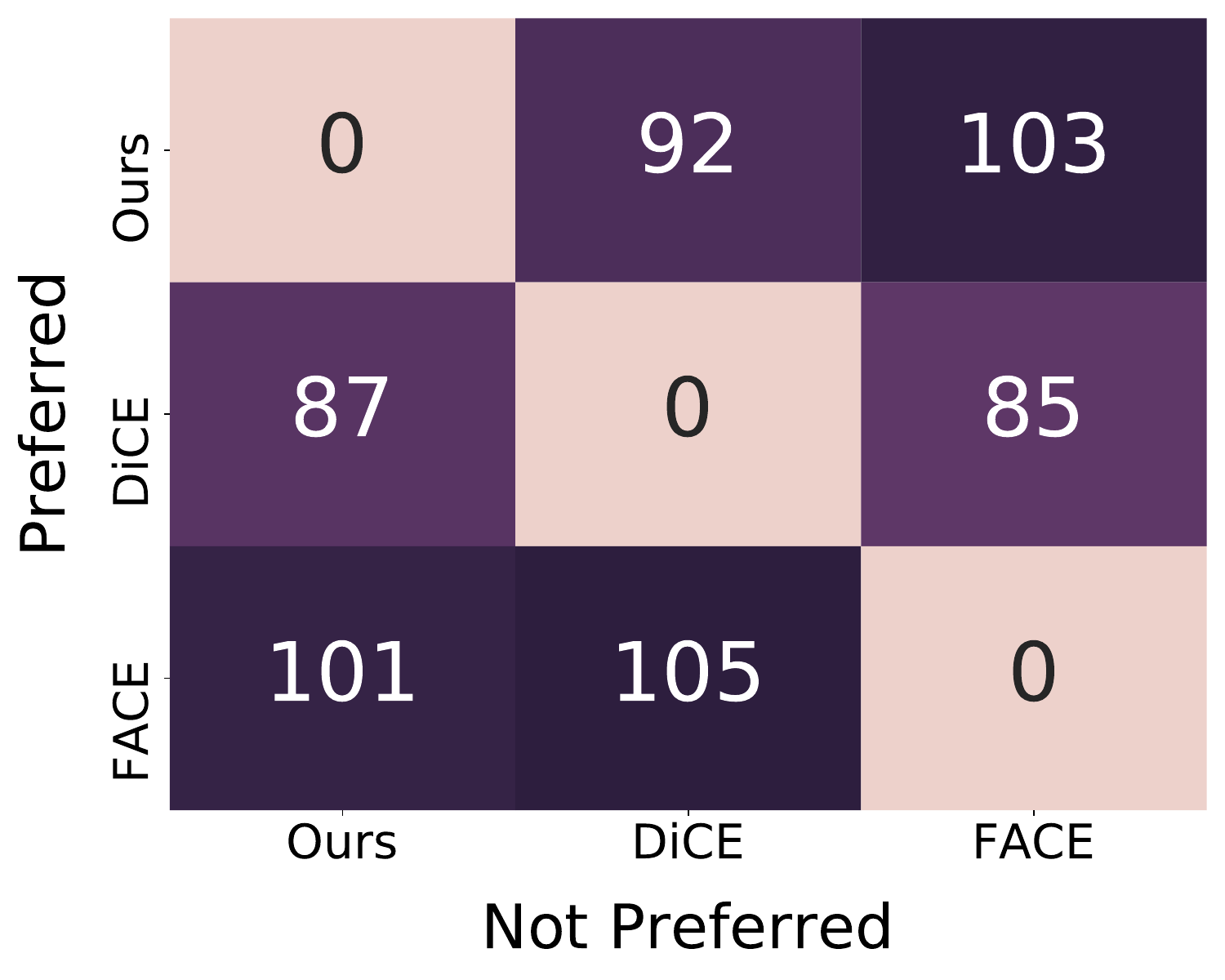}
    }
  
    \caption{Preference matrices for survey responses on each dataset. Each cell shows how often a respondent preferred the row method to the column method--darker colors imply a greater preference. Each method seems to excel on different types of data.}
    \label{fig:results}
\end{figure*}

We evaluated the subjective characteristics of our approach via a human-subjects survey on Amazon Mechanical Turk (AMT). Our evaluation proceeded as follows: Each participant was first introduced to the decision-making context; before being prompted to respond to a series of 12 pairwise comparison questions, in which they were provided randomly generated counterfactuals from two different methods at a time. Respondents chose which explanation was most `satisfying' to them, and wrote a short justification that describes the reasoning for their choice.  We aggregated the preferences and performed a binomial test to determine statistically significant preferences among methodologies with respect to each dataset. Appendix \ref{app:survey_screen} provides an example survey screen. 

\xhdr{Decision-making Contexts}
Each respondent was randomly assigned to one of three hypothetical decision-making contexts based on three tabular datasets: LUCAS \cite{guyon2008design}, Adult \cite{Dua:2019}, and German Credit \cite{Dua:2019}. As stated above, Appendix~\ref{app:datasets} provides details on these datasets, yet at a high-level: LUCAS allows us to investigate whether respondents' knowledge of causal dependencies influences their preferred explanation; Adult allows us to investigate to what extent a respondent's background knowledge of a social system influences their preferences; and German Credit allows us to investigate how respondents' preferences change when the model utilizes a large feature set, making it difficult for respondents to understand all dependencies. Due to space constraints, we leave the evaluation of LUCAS to Appendix \ref{}.

\xhdr{Counterfactual Generation Methods}
As described in our quantitative evaluation, one way of interpreting our approach is as a sliding scale between algorithms that sample counterfactuals from a region around a reference and algorithms that sample counterfactuals as instances from the underlying dataset. Our experiments investigate whether participants have a preference for one side or another in this dichotomy. Thus, we contrast a middle ground $\alpha$ in our approach with two existing counterfactual explanation methods that align with these extremes: \textbf{Diverse Counterfactual Explanations (DiCE)}~\citep{mothilal2020explaining} and \textbf{Feasible and Actionable Counterfactual Examples (FACE)}~\cite{poyiadzi2020face}. Note that While we benchmarked our approach by changing the distance metrics for existing implementations above, here, we generate counterfactuals from our approach by sampling from the conditional Gaussian distribution from Eq. \eqref{eq:xx'_joint}, as described in Appendix \ref{app:complex}. We initially hypothesize that participants prefer a set of actionable changes in line with ensuring plausibility above all else. This entails that preferences for would have the ordering from least to most preferred: `DiCE\ (\emph{Implausible})', `Ours\ (\emph{Relaxed\ Plausibility})', `FACE\ (\emph{Strictly\ Plausible} )'

\subsubsection{Findings}
\label{sec:findings}

For each dataset, we analyze the participants' preferences, and review the justifications for each preference. We conclude with a discussion of the commonalities and differences among justifications. 

\xhdr{German (N=$430$ comparisons)}
German credit shows no significant preference for one method over another, however, as shown in Fig.~\ref{fig:results} our approach is slightly preferred to both FACE ($p=0.062$) and DiCE ($p=0.149$). Respondents seemed to prefer explanations that were more different from the reference as they perceived these cases as more detailed. For example, one respondent justified their preference with:
\qt{
    `Method [FACE] seems more satisfactory to me because it is more descriptive in its credit requirements.'
}
Another with:
\qt{
    [Ours] includes more data that would matter more when making a decision.'
}
$28$ of the $430$ choices explicitly listed that having more detail was the primary reason for preferring a given explanation; only $2$ preferred having fewer changes.

In addition, participants gave a great deal more focus on those features for which their pre-existing beliefs align with credit worthiness:
\qt{
    `Id use [DiCE] because it mentions employment and his good credit score. It does not mention his other debts though. I had a hard time choosing because of that.'
}
Potentially due to the participants' existing intuition on the information relevant to credit worthiness, they may disregard explanations that do not fit their existing beliefs.

These factors may play into the reasons for why our proposed method was more preferred than the alternatives. While DiCE optimizes for minimal changes, explainees preferred a wider set of changes that allow for more flexibility in what sorts of changes could potentially be enacted. On the other hand, participants often listed continuous features such as the amount of credit requested or loan duration in months as a major reasons for choosing one explanation over another:
\qt{
    `Method [Ours] makes more sense because it provides valid reasons including credit amount and duration and employment duration...'
}
FACE finds explanations from within the dataset. Without a large number of samples from which to choose, the mix of features on very different scales may be giving more preference to methods such as ours or DiCE that allow for new points to be generated as explanations. Our method would thus be the preferred approach due to not being as susceptible to either case.

\xhdr{LUCAS (N=$481$ comparisons)}
As shown in Fig.~\ref{fig:results}, participants on the LUCAS dataset were found to have a statistically significant preference for DiCE to FACE  (p=$0.014$), a nearly significant preference for DiCE to Ours (p=$0.065$), and a slight preference between Ours and FACE ($p=0.119$).  

As LUCAS is a synthetic binary dataset with causal dependencies, respondents seemed to prefer explanations that fit more closely to their understanding of these causal relationships. For example, one respondent justified their preference as:
\qt{
   ``With lung cancer, smoking is such a strong indicator, or correlator.  Anxiety provides a reason why tey [sic] are a smoker, extra evidence.''
}. 
One participant had a particularly detailed understanding of the underlying dynamics:
\qt{
    ``The methodology of anxiety being the main factor in this prediction leads me to assume that the fact they have Yellow Fingers means they smoke, whereas Method [FACE] states they don't which is wrong...''
}
This would imply that participants prefer methodologies that better adhere to the true distribution of data. However, as DiCE, which does not use this information, has a statistically significant preference over the other methods, there may be another reason that supercedes faithfulness to the data distribution when determining preferences.

Some participants pointed to specific features as being less preferable to change:
\qt{
    ``It would make the person's life much harder cause he has the peer pressure mess with him.''
}
and
\qt{
    ``i take yellow fingers over anxiety any day.''
}
Rather than emphasizing plausibility, the underlying cost that a person places on each feature seems to play a greater role. DiCE may be the preferred method because making minimal changes with the greatest impact decreases the potential for changing ancillary features which people place a high cost on. By considering the conditional dependencies in our method or FACE, we are more likely to include the low-probability outcomes that correlate to these high-cost changes (e.g., facing peer pressure and anxiety without being a smoker).

\xhdr{Adult (N=$573$ comparisons)}
Respondents on this set of data gave no statistically significant preference for any particular method, however, as shown in Fig.~\ref{fig:results}, there was a small degree of preference for FACE over DiCE (p=$0.084$). 
Participant preference justifications also varied significantly. As in German Credit, a common theme that emerged was that participants seemed to prefer explanations that had a greater number of changes from the reference:
\qt{
    `Method [FACE] is much more detailed and gives more information to make a better informed decision of the person in question. Method [Ours] has less information makes it less satisfying and harder to fully judge the person.'
}
At least 40 of the 573 comparisons for this dataset justified their preference by a combination of `more details', `more information', and `less restrictions'. Some respondents even went so far as to choose the explanation with a greater number of changes because the alternative had too few changes:
\qt{
    `Method [FACE] has too few changes to get up to >50k a year.'
}
Respondents explicitly disagreed with the classifier because the changes were too subtle. In contrast, only 8 cases out of the 573 explicitly listed that they chose one explanation over another due to that explanation having fewer changes.

Outside of the number of changes, dependencies among covariates led to participants labeling potential explanations as implausible:
\qt{
    "Method [FACE] lists a doctorate but that degree probably isn't necessary for tech support."
}
Modeling dependencies between features is necessary in order to avoid such cases, however, no method excels here. 
Alternatively, many respondents chose a preferred explanation based on a single feature that made the most sense to them:
\qt{
    `contain [sic] technical level occupation'
} or 
\qt{
    `working hours is more than the other'
}. In cases where an option is unreasonable, participants default to the alternative, regardless of its plausibility. For example, one explanation suggested working 99 hours per week:
\qt{
    `99 hours is too many hours to compare to'
}

These reasons do not lend themselves to being solved by any of the considered benchmarks. While not a significant preference, the FACE algorithm does not return implausible points, however, when traversing the KNN graph, after a few steps, FACE no longer encourages making minimal changes. It seems that FACE is preferred due to its propensity of returning distant explanations, while guaranteeing plausibility.

\section{Discussion and Future Directions}
Here, we have introduced a new process for generating counterfactual explanations by revisiting their underlying generative model. In motivating this approach, we have shown that common optimization-based counterfactual explanation methods implicitly assume that counterfactuals do not come from the underlying data distribution, but are sampled from a ball centered at the reference point. This in turn leads to unrepresentative explanations for the underlying data distribution. We show an approach that is constructed to avoid this issue, while incorporating nuanced notions of plausibility.

In order to evaluate conditions of usability for our approach, we benchmarked our approach against several existing counterfactual generation methods and conducted an AMT survey in which respondents perform a binary forced-choice task expressing their preferences among explanation methods. We found no universal preference for one explanation approach regardless of the extent to which they encode plausibility or actionability. While participants understand the relationships among features, they seem to rely on a subjective notion of cost for certain modifications. As \citet{barocas2020hidden} and \citet{selbst2019fairness} highlight, explanations are often rational only in the context of ensuring a desired outcome from a model, but not with respect to the goals that individuals have for themselves. This is consistent with our observations.

Moreover, in contrast to conventional wisdom, we observed a preference for counterfactuals that are distant from a reference---based on the perception that they provide a  detailed plan of action, and subsequently, greater potential for actionable recourse. As raised by \citet{barocas2020hidden}, features may be relevant to multiple domains. Recommended changes may be beneficial in terms of a model's outcome but harmful in other cases (e.g., a counterfactual may recommend applying for a job with higher pay, but a lower paying position provides better health insurance). A large number of potential avenues for change may allow individuals to make many incremental lifestyle changes, as opposed to drastic changes in a small areas. 

We conclude by noting that in our evaluation, we allowed a great deal of freedom in how to define/interpret a `satisfying' explanation. Our findings indicate the need for more fine-grained hypotheses on usability conditions, including those that account for the explainee's subjective mental models. Finally we remark that our exploratory survey does not replace contextualized, application-specific evaluations needed to understand human perception of explanations. We leave this as critical avenue for future work to explore.

\bibliography{biblio.bib}
\bibliographystyle{tmlr}

\appendix
\onecolumn
\newpage
\section{Regularized Counterfactuals}
\label{app:regularization}

In Section \ref{sec:background}, we focused on analyzing the implications of the \citet{wachter2017counterfactual} objective for generating counterfactual explanations for a regression model with underlying Gaussian data. We show that expressing distance solely through dependence on the reference, $\vx$, can create a distribution of counterfactuals wholly independent from the underlying data distribution. 

As the underlying data is Gaussian, one may suspect that this lack of representation can be corrected by applying a Gaussian regularizer that encourages the distribution to be representative of the underlying data. In other words, for $\vx, \vx' \sim \mathcal{N}( \mu, \Lambda^{-1} )$, the issues mentioned in section \ref{sec:motivation} may be mitigated by updating the optimization problem presented in equation \eqref{eq:opt} to,
\begin{equation}
    \vx' = \arg \min_{\tilde{x}} || y' - f( \tilde{x} ) ||_{2}^{2} + \gamma_{1} || x - \tilde{x} ||_{2}^{2} + ( \tilde{x} - \mu )^{T} ( \gamma_{2} I \odot \Lambda ) ( \tilde{x} - \mu ).
    \label{eq:opt_reg}
\end{equation}
(Note that if the underlying data distribution is standard normal, then this regularizer is $\gamma_{2} || \vx' ||_{2}^{2}$). 

As done for Eq \eqref{eq:opt}, a quadratic formulation of Eq \eqref{eq:opt_reg}, tells us that this objective is also expressing a known Gaussian distribution, and the optimization problem is simply finding its mode (full distribution derivation and parameters provided in Appendix \ref{app:derivation_pgm2}),

\begin{figure}
    \centering
    \subfloat[Regularized PGM\label{fig:pgm3}]{%
      \includegraphics[width=0.18\textwidth]{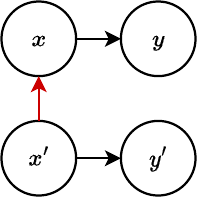}
    }
    \hfill
    \subfloat[$L = I, S = I$\label{fig:pgm3_distb}]{%
      \includegraphics[width=0.2\textwidth]{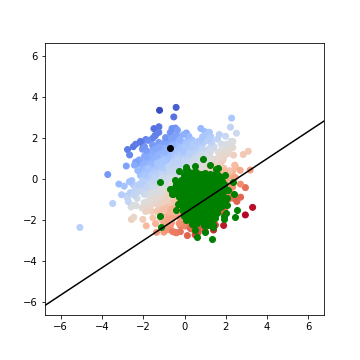}
    }
    \hfill    
    \subfloat[$L = 20 \cdot I, S = 0.1 \cdot I$\label{fig:pgm3_distc}]{%
      \includegraphics[width=0.2\textwidth]{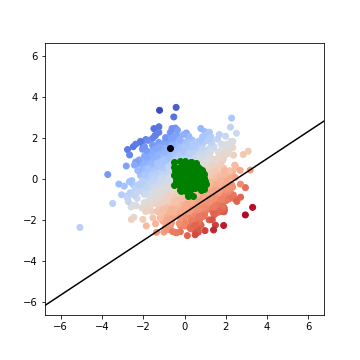}
    }
  
    \subfloat[$L = 0, S = I$\label{fig:pgm3_diste}]{%
      \includegraphics[width=0.2\textwidth]{figures/regularized_wachter_figs/y_10.0-L_1-S_0.1.png}
    } 
     \hfill  
    \subfloat[$L = I, S = I$\label{fig:pgm3_distd}]{%
      \includegraphics[width=0.2\textwidth]{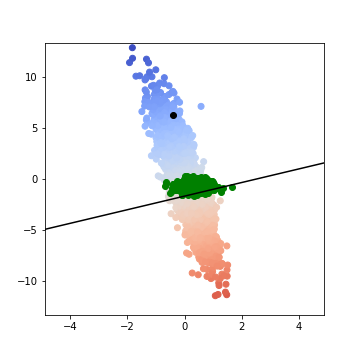}
    }
    \hfill
    \subfloat[$L = 1, S = 0.5 \cdot I$\label{fig:pgm3_distf}]{%
      \includegraphics[width=0.2\textwidth]{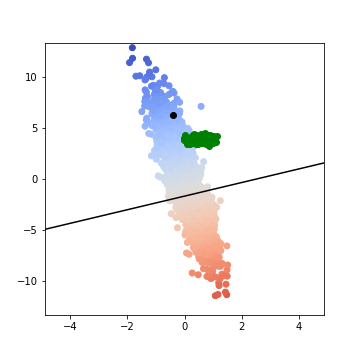}
    }

\caption{(a)  PGM underlying counterfactual generation with a regularizer to encourage in-distribution counterfactuals. (b,c,d,e,f - Black) Reference for the counterfactual. (b,c,d,f,g,h - Black Line) Desired predicted output, $y' = 10$ for the the regression problem, $y = 2 x_{1} - 3 x_{2} + 5$. (b,c,d,e,f - Green) Distribution entailed by Eq \eqref{eq:opt_reg}, where $L$ is the inverse variance of the residuals, and $S$ is the weighted euclidean distance between counterfactuals and reference. In all figures, $I$ is the identity matrix, the desired predicted output, $y' = 10$ (black line) and the underlying data distribution, (b,c) standard Gaussian, (d,e,f) $\vx \sim \mathcal{N}\bigg( \begin{pmatrix} 0 \\ 0 \end{pmatrix}, \begin{pmatrix} 4.04 & -7.80 \\ -7.80 & 17.00 \end{pmatrix} \bigg)$.}
\label{fig:pgm3_all}
\end{figure}

Moreover, the resultant counterfactual distribution is entailed by the PGM under Figure \ref{fig:pgm3}. One can see this by considering the factorization of the joint distribution for Figure \ref{fig:pgm3},
\begin{align*}
    p( \vx, \vx', y', y ) &= p( y' | \vx' ) p( y | \vx ) p( \vx | \vx' )  p( \vx' ) 
\end{align*}
As $\vx$ d-separates $\vx'$ from $y$, we drop $y$ in order to express the distribution over only the terms that are dependent on $\vx'$. The negative log probability of this distribution,
\begin{align*}
   - \log  p( \vx, \vx', y' )  &= - \log( \mathcal{N}( y' | A \vx' + b, L^{-1} ) ) - \log( \mathcal{N}( \vx | \vx', W^{-1} ) ) - \log( \mathcal{N}( \vx' | \mu, \Lambda^{-1} ) ),
\end{align*}
is equivalent to Eq.~\eqref{eq:opt_reg}, when $L$ and $W$ are the identity. In other words, by regularizing the counterfactual optimization problem, we are reversing the dependency on the counterfactual and reference, effectively going against our intuition about what counterfactuals are, by saying that the reference provided by a user is dependent on the set of counterfactuals.

We provide visualizations of this distribution in Figure \ref{fig:pgm3_all}. Empirically, it seems that while regularization creates a generative model that runs counter to our intuition, it does address some of the issues from before. Most notably in comparing figures \ref{fig:pgm1_distb} and \ref{fig:pgm3_distd}, this form of regularization does encourage the distribution to stay within the data distribution. However, as shown in Figure \ref{fig:pgm3_distf} this encouragement may not be enough. Moreover, as can be seen in Figure \ref{fig:pgm3_distb}, unlike in the previous case, this new distribution places lower emphasis on returning counterfactuals with the desired label. The proposed fix in Eq.~\eqref{eq:opt_reg} encourages a heavy trade-off between representativeness of the underlying data distribution and ensuring counterfactuals that tightly cluster around the desired label.

\newpage
\section{Extending the Proposed Framework to Complex Models}
\label{app:complex}

In Section \ref{sec:method_intro}, we focused on recommending a change to the graphical model that underlies Counterfactual Explanation generation methods. We introduced a prior that allows us to express the relationship between the reference $\vx$ and counterfactual $\vx'$ in terms of underlying data distribution. Section \ref{sec:method_intro} was restricted to the linear regression model; here, we show how to express this prior in more complex decision settings.

Consider a multi-class classification setting\footnote{There are many settings in which we would like to generate counterfactual explanations where one may not have access to the model itself (eg. a vision API) or if the decision-making model is non-differentiable (eg. decision-trees); WLOG, we can train a new model to mimic decisions and reduce the problem to the considered case} in which decisions are made by a differentiable model, $f : \mathcal{X} \rightarrow \mathcal{Y}$, where $\mathcal{Y} \in \{0,1\}^{m}$ is some categorical labeling. For a given outcome, we sample counterfactuals by first splitting the network into two sections; the first $N - 1$ layers being the feature representation, $r: \mathcal{X} \rightarrow \mathbb{R}^{m}$, and the second being a linear output layer. The full network takes the form, $f( x ) = \sigma( w^{T} r( x ) )$, where $\sigma( z_{i} ) = \frac{\exp( z_{i} )}{ \sum_{j} \exp( z_{j} ) }$ is the softmax function; the posterior distribution over the reference, $\vx$, explanation $\vx'$, and desired predicted outcome, $y \sim \mathrm{Categorical}( p )$ follows,
\begin{align}
    \begin{split}
    p( x' | y' x ) &= p( x | x' ) p( y' | x'; r ) p( x' ) \label{cf_deep} \\
    &= \mathcal{N}( x | \mu_{x|x'}, S_{x|x'}^{-1} ) \times ( y'^{T} \sigma( w^{T} r( x' ) ) \times \mathcal{N}( x' | \mu, \Lambda^{-1} ),
    \end{split}
\end{align}
Unlike the linear case, by considering the learned representation of the network, $r( x )$, we are introducing another degree of uncertainty over model weights. We can introduce a prior over the networks output weights, in order to capture this uncertainty, and return a fully Bayesian model. We marginalize over the set of all possible output weights under this prior in order to average out our uncertainty.
\begin{align}
\begin{split}
    p( x' |  x, y'; \phi, t ) &= \mathcal{N}( x | \mu_{x|x'}, S_{x|x'}^{-1} ) \times \\ \int_{w} &( y'^{T} \sigma( w^{T} r( x' ) ) + t^{T} \sigma( w^{T} r( \phi ) ) \times \mathcal{N}( w | 0, I ) \times \mathcal{N}( x' | \mu, \Lambda^{-1} ) d w.
    \label{cf_laplace},
\end{split}
\end{align}
where $\phi$ are all other points in the dataset, and $t$ are the corresponding outputs of the decision-maker for inputs $\phi$. Similarly to how one would perform a Bayesian Logistic Regression, we perform a Laplace Approximation on the integrand in order to simplify the process of marginalizing over the weights, and ensure that we have a Gaussian form for the counterfactual distribution.\footnote{This restriction to be Gaussian is not particularly necessary. As in the main text, we focus on the Gaussian case for this work in order to make our manipulation of the posterior more easily understandable and to allow for easier sampling, however, we can perform various off-the-shelf methods of sampling from a posterior distribution in order to sample arbitrary distributions.}  

As we consider the outcome, $y$ to be categorical, the integrand,
\begin{equation*}
   ( y'^{T} \sigma( w^{T} r( x' ) ) \times \mathcal{N}( w | 0, I ) \times \mathcal{N}( x' | \mu, \Lambda^{-1} ),
\end{equation*}
is effectively dependent only on $x'$. Thus, the Laplace Approximation, 
\begin{equation*}
    g( x' | y' ) \sim \mathcal{N}( \mu_{\tilde{x}}, \Lambda^{-1}_{\tilde{x}} ) \approx \int_{w} p( y' | x'; r ) p( w | \phi, t ) p( x' ) d w,
\end{equation*} 
can be considered as learning a new prior over the data distribution. Whereas $p( x' )$ may cover the entire data distribution, $g( x' | y )$ covers only the region of the data distribution that corresponds to label $y$.

Generating counterfactual explanations then amounts to sampling from the posterior,
\begin{equation}
    \nonumber
    g( x' | x, y' ) \appropto p( x | x' ) g( x' ),
\end{equation}
in which $p( x' | x, y' )$ is Gaussian, and $\appropto$ is defined as `approximately proportional to'. In Appendix \ref{sec:laplace}, we include a discussion on the practical considerations for incorporating the Laplace Approximation in this setting.

\subsection{Extending the Proposed Framework to Complex Data}
\label{app:cf_decode}
We often choose to use complex decision-making models, such as deep networks, due to the fact that the relationships in the data cannot be expressed through simple, linear relationships (eg. convolutional filters in images or recurrent architectures in time-series data). In such cases, we cannot directly sample from the counterfactual distribution in Eq \eqref{cf_deep}, due to the fact that we cannot express an effective prior over the data, ie. images cannot be reliably generated by randomly sampling pixel values. Without an effective prior on the space of counterfactuals, counterfactual explanations for complex data are functionally equivalent to adversarial perturbations, as has been pointed out in \cite{freiesleben2020counterfactual}. 

In such cases, engineers often opt to use generative models, which allow them to sample from an underlying latent space and pass this sample through a generator that maps into the input space. We follow a similar approach, by placing a prior not on the input space, but on a Gaussian latent space, and include the latent decoder, $\mathrm{d}: \mathbb{R}^{k} \rightarrow \mathcal{X}$, that maps from the latent space into the input space.
\begin{align}
    p( x, y, x', y' ) &= p( x | l; \mathrm{d} ) p( y | l; r, \mathrm{d} ) p( l );\ x' = \mathrm{d}( l ) \label{cf_decode} \\
    &= \mathcal{N}( x | \mathrm{d}( l ), S^{-1} ) \times ( y^{T} \sigma( A^{T} r( \mathrm{d}( l ) ) + b ) \times \mathcal{N}( l | \mu, \Lambda^{-1} ),
\end{align}
There are various ways that one may represent the Gaussian latent space (eg. Normalizing flows \cite{rezende2015variational} or Variational Auto Encoders \cite{kingma2013auto}). However, once this encoding/decoding is learned, the sampling process itself bears no further difference from Section \ref{app:complex}. 

Importantly, we can engineer the decoding layer to allow us to address the issues of normalizing features that have very different scales \cite{barocas2020hidden}. Commonly prior work on counterfactual explanations use the Median Absolute Deviation (MAD) under the L1 norm \cite{mothilal2020explaining} in order to allow for optimizing the counterfactual objective, however through this encoding/decoding approach, we can express any feature that we have to normalize through a Gaussian latent variable and decode into the desired scale. For example, one may encode income as the exponential of a Gaussian latent variable or one may encode categorical features as the softmax of a vector of independent Gaussians, and binary features as the sigmoid of a Gaussian. 

\subsection{Practical Consideration of the Laplace Approximation}
\label{sec:laplace}

While posterior sampling of the cases outlined in Sections \ref{app:complex} and \ref{app:cf_decode} can be accomplished via a myriad of methods, as stated above, we focus on the Gaussian case here in order to ensure that the counterfactual distribution from which we sample from remains tractable and well understood. In doing so, we have to approximate the likelihood and counterfactual prior Eq.~\eqref{cf_laplace} as Gaussian using the Laplace Approximation.
This method approximates an arbitrary distribution, $f_{x}$ as Gaussian through a two step procedure. First we set as the mean of the approximation the mode of $f_{x}$, ie. $\bar{x} \ni f_{x}( \bar{x} ) \geq f_{x}( x' )\ \forall\ x'$. We then set as the approximation's covariance, $\Sigma^{-1} = \nabla^{2} f_{x}( \bar{x} )$. One can see why this choice of covariance is used by performing a second order taylor expansion of $\log f_{x}$ around $\bar{x}$, and seeing that this is proportional to a Gaussian with mean $\bar{x}$ and covariance $\Sigma$.

For complex models, when performing the Laplace approximation over the classifier's learned representation, $r: \mathcal{X} \rightarrow \mathbb{R}^{m}$, and latent representation, $\mathrm{d}: \mathcal{R}^{k} \rightarrow \mathcal{X}$, finding the mode becomes intractable. Finding $\bar{x}\ s.t.\ f_{x}( \bar{x} ) \geq f_{x}( x' )\ \forall\ x'$, implies finding $\bar{x}\ s.t.\ f_{x}( \bar{x} ) \geq ( r \circ \mathrm{d} )( x' ) \forall x'$, in other words, we need to find the input that globally minimizes loss over the composition of two non-convex functions. 
Finding such a solution is infeasible, so the approximation will inevitably be based on local optima. Hence, the new conditional prior, $g( x | y )$ that we place on a counterfactual, while designed to cover the distribution of data that returns a desired, predicted label, instead covers only a portion of that space, and in some cases, may include the space of points from which we return different labels.

\newpage
\section{Derivation of Counterfactual Distribution Under Figure \ref{fig:pgm1}}
\label{app:derivation_pgm1}

Following the method outlined in \citet{bishop2006pattern}, in this section we show the derivation of the Gaussian counterfactual distribution entailed by the PGM in figure \ref{fig:pgm1}. 

The general approach allows us to express the joint distribution of multiple gaussian densities by considering the log probability of their joint distribution, and noting that,
'\begin{align*}
    - \log p( x ) &= \frac{1}{2} ( x - \mu )^{T} \Lambda ( x - \mu ) \\
        &= \frac{1}{2} \big( x^{T} \Lambda x - 2 \mu \Lambda x - \mu^{T} \Lambda \mu \big) \\
        &\implies x \sim \mathcal{N}( \Lambda^{-1} \Lambda \mu, \Lambda^{-1} ),
\end{align*}
thus the quadratic parameters that are dependent on the variable of interest make up the inverse covariance, and the product of this covariance matrix and the linear parameters make up the mean.

The counterfactual distribution entailed by figure \ref{fig:pgm1} is formed as the posterior of,
\begin{align*}
    p( \vx' | \vx, y' ) &\propto p( y' | \vx' ) p( \vx' | \vx ) p( \vx ) \\
    &= \mathcal{N}( y | A \vx' + b, L^{-1} ) \times \mathcal{N}( \vx' | \vx, W^{-1} ) \times \mathcal{N}( \vx | \mu, \Lambda^{-1} )
\end{align*}

The negative log probability is expressed as follows (Note that the prior $\mathcal{N}( \vx | \mu, \Lambda )$ is discarded as it is not dependent on the variable of interest $\vx'$:
\begin{align*}
    - \log( p( \vx' | \vx, y' ) ) &\propto ( y - A \vx' - b )^{T} L ( y - A \vx' - b ) + ( \vx' - \vx )^{T} W ( \vx' - \vx ) \\
    &= y^{T} L y - 2 y^{T} L A \vx' - 2 y^{T} L b + \vx'^{T} ( A^{T} L A ) \vx' + 2 b^{T} L A \vx' + b^{T} L b + \vx' W \vx' - 2 \vx W \vx' + \vx W \vx
\end{align*}

Grouping the quadratic terms,
\begin{align*}
    \vx'^{T} \Lambda_{cf} \vx' &= \vx'^{T} ( A^{T} L A + W ) \vx',
\end{align*}
which implies that our covariance, $\Lambda_{cf}^{-1} = ( A^{T} L A + W )^{-1}$,

Next we group the linear terms,
\begin{align*}
    -2 \mu^{T} \Lambda_{cf} \vx' = -2 ( y^{T} L A - b^{T} L A + \vx W )^{T} \vx',
\end{align*}
The mean of our distribution is then, $\mu_{cf} = \Lambda_{cf}^{-1} ( A^{T} L y' - A^{T} L b + W \vx )$

Thus the distribution of the counterfactual distribution for the Linear Regression case under PGM \ref{fig:pgm1} is,
\begin{align*}
        \vx' &\sim \mathcal{N}( \mu_{cf}, \Lambda_{cf}^{-1} ) \\
        \Lambda_{cf}^{-1} &= ( W + A^{T} L A )^{-1} \\
        \mu_{cf} &= \Lambda_{cf}^{-1} ( A^{T} L y' - A^{T} L b + W \vx ),
\end{align*}

\newpage
\section{Derivation of Counterfactual Distribution Under Figure \ref{fig:pgm2}}
\label{app:derivation_pgm2}

Following the approach from Appendix \ref{app:derivation_pgm1}, here, we derive the parameters of the counterfactual distribution under our proposed prior.

Before forming the posterior, recall that for the distribution,
\begin{equation*}
    \begin{pmatrix} \vx \\ \vx' \end{pmatrix} \sim \mathcal{N}\Bigg( \begin{bmatrix} \mu \\ \mu \end{bmatrix}, \begin{bmatrix} \Lambda^{-1} & W \\ W^{T} & \Lambda^{-1} \end{bmatrix} \Bigg),
\end{equation*}
the conditional distribution, $p( \vx | \vx' )$ is,
\begin{equation*}
    \mathcal{N}( \mu + W \Lambda ( \vx' - \mu ), \Lambda^{-1} - W \Lambda W^{T} )
\end{equation*}

The counterfactual distribution entailed by Figure \ref{fig:pgm2} is formed as the posterior of,
\begin{align*}
    p( \vx' | \vx, y' ) &\propto p( y' | \vx' ) p( \vx | \vx' ) p( \vx' ) \\
    &= \mathcal{N}( y' | A \vx' + b, L^{-1} ) \times \mathcal{N}( \mu + W \Lambda ( \vx' - \mu ), \Lambda^{-1} - W \Lambda W^{T} ) \times \mathcal{N}( \mu, \Lambda^{-1} )
\end{align*}

The negative log probability is expressed as follows:
\begin{align*}
    - \log( p( \vx' | \vx, y' ) ) &\propto ( y - A \vx' - b )^{T} L ( y - A \vx' - b ) \\
    &\hspace{1cm} + ( \vx - \mu - W \Lambda ( \vx' - \mu ) )^{T} ( \Lambda^{-1} - W \Lambda W^{T} )^{-1} ( \vx - \mu - W \Lambda ( \vx' - \mu ) ) \\
    &\hspace{1cm} + ( \vx' - \mu )^{T} \Lambda ( \vx' - \mu ) \\
\end{align*}

For brevity, let $K =  ( \Lambda^{-1} - W \Lambda W^{T} )^{-1}$. We can simplify the log probability to
\begin{align*}
    - \log p( \vx' | \vx, y' ) &\propto y^{T} L y - 2 y^{T} L A \vx' - 2 y^{T} L b + \vx' ( A^{T} L A ) \vx' + 2 b^{T} L A \vx' \\
    &\hspace{1cm}+ b^{T} L b + \vx^{T} K \vx - 2 \vx^{T} K ( \mu + W \Lambda ( \vx' - \mu ) ) + ( \mu + W \Lambda ( \vx' - \mu ) )^{T} K ( \mu + W \Lambda ( \vx' - \mu ) ) \\
    &\hspace{1cm}+ \vx'^{T} \Lambda \vx' - 2 \mu^{T} \Lambda \vx' + \mu^{T} \Lambda \mu
\end{align*}

Grouping the quadratic terms together,
\begin{align*}
    \vx' \Lambda_{cf} \vx' &= \vx' ( A^{T} L A ) \vx' + \vx'^{T} ( \Lambda W K W \Lambda ) \vx' + \vx'^{T} \Lambda \vx',
\end{align*}

Thus after substituting in $K =  ( \Lambda^{-1} - W \Lambda W^{T} )^{-1}$, the covariance of the distribution is, 
\begin{equation*}
    \Lambda_{cf}^{-1} = ( A^{T} L A + \Lambda W  ( \Lambda^{-1} - W \Lambda W^{T} )^{-1} W \Lambda + \Lambda )^{-1}
\end{equation*}

Grouping the linear terms together,
\begin{align*}
    -2 \mu^{T} \Lambda_{cf} \vx' &= -2 y^{T} L A \vx' + 2 b^{T} L A \vx' - 2 \vx^{T} K W \Lambda \vx' + 2 \mu^{T} K W \Lambda \vx' - 2 \mu^{T} K W \Lambda \vx' - 2 \mu^{T} \Lambda \vx' \\
    &= -2 y^{T} L A \vx' + 2 b^{T} L A \vx' - 2 \vx^{T} K W \Lambda \vx' - 2 \mu^{T} \Lambda \vx' \\
    &= - 2 ( y^{T} L A - b^{T} L A + \vx^{T} K W \Lambda + \mu^{T} \Lambda ) \vx'.
\end{align*}
This implies that the mean of the distribution of counterfactuals is,
\begin{align*}
    \mu_{cf} &= \Lambda_{cf}^{-1} ( A^{T} L y - A^{T} L b + \Lambda W K \vx + \Lambda \mu ) \\
    &= \Lambda_{cf}^{-1} ( A^{T} L y - A^{T} L b + \Lambda W ( \Lambda^{-1} - W \Lambda W^{T} )^{-1} \vx + \Lambda \mu )
\end{align*}

Thus the distribution of counterfactuals for the Linear Regression case under PGM \ref{fig:pgm2} is,
\begin{align*}
    \vx' &\sim \mathcal{N}( \mu_{cf}, \Lambda^{-1}_{cf} ) \\
    \Lambda^{-1}_{cf} &=  ( A^{T} L A + \Lambda W  ( \Lambda^{-1} - W \Lambda W^{T} )^{-1} W \Lambda + \Lambda )^{-1} \\
    \mu_{cf} &= \Lambda_{cf}^{-1} ( A^{T} L y - A^{T} L b + \Lambda W ( \Lambda^{-1} - W \Lambda W^{T} )^{-1} \vx + \Lambda \mu ).
\end{align*}

\subsection{Derivation of the Objective in Eq.~\eqref{eq:opt_ours}}
\label{app:derivation_our_objective}

In our proposed prior, the marginal distribution of counterfactuals is, $p( x' ) = \mathcal{N}( \mu, \Lambda^{-1})$, and the conditional distribution. $p( \vx | \vx' ) = \mathcal{N}( \mu + W \Lambda ( \vx' - \mu ), \Lambda^{-1} - W \Lambda W^{T} )$. 

Consider the negative log of the posterior distribution over our counterfactuals,
\begin{align*}
    - 2 \log( p( \vx' | \vx, y' ) ) &= - 2 \log( p( y' | \vx' ) ) - 2 \log( p( \vx | \vx' ) ) - 2 \log( p( \vx' ) ) \\
    &= - 2 \log( p( y' | \vx' ) ) \\
    &\hspace{1cm} - ( \vx - \mu - W \Lambda ( \vx' - \mu ) )^{T} ( \Lambda^{-1} - W \Lambda W^{T} )^{-1} ( \vx - \mu - W \Lambda ( \vx' - \mu ) ) \\
    &\hspace{1cm} - ( \vx' - \mu )^{T} \Lambda ( \vx' - \mu )
\end{align*}

Recall that $W = \alpha \Lambda^{-1}$, substitute this term into $- \log( p( \vx | \vx' )$,
\begin{align*}
    ( \vx - \mu - W \Lambda ( \vx' - \mu ) )^{T} &( \Lambda^{-1} - W \Lambda W^{T} )^{-1} ( \vx - \mu - W \Lambda ( \vx' - \mu ) ) \\
    &= ( \vx - \mu - \alpha ( \vx' - \mu ) )^{T} ( ( 1 - \alpha^{2} ) \Lambda^{-1} )^{-1} ( \vx - \mu - \alpha ( \vx' - \mu ) )
\end{align*}

We then discard all terms not dependent on $\vx'$,
\begin{align*}
( \vx - \mu - W \Lambda ( \vx' - \mu ) )^{T} &( \Lambda^{-1} - W \Lambda W^{T} )^{-1} ( \vx - \mu - W \Lambda ( \vx' - \mu ) ) \\
    &= \frac{2 \alpha}{ 1 - \alpha^{2}} ( \frac{\alpha}{ 2 } \vx' \Lambda \vx' + ( 1 - \alpha ) \mu \Lambda \vx' - \vx \Lambda \vx' )
\end{align*}

Substituting this back into our log posterior,
\begin{align*}
 - 2 \log( p( \vx' | \vx, y' ) ) &= - 2 \log( p( y' | \vx' ) ) \\
 & \hspace{1cm} - \frac{2 \alpha}{ 1 - \alpha^{2}} ( \frac{\alpha}{ 2 } \vx' \Lambda \vx' + ( 1 - \alpha ) \mu \Lambda \vx' - \vx \Lambda \vx' ) \\
 & \hspace{1cm} - \vx' \Lambda \vx' + 2 \mu \Lambda \vx' \\
 &= - 2 \log( p( y' | \vx' ) ) - \frac{1}{1-\alpha^{2}} ( \vx' )^{T} \Lambda \vx' + \frac{2 ( 1 -  \alpha ) }{ 1 - \alpha^{2} } \mu^{T} \Lambda \vx' + \frac{2 \alpha}{ 1 - \alpha^{2}} \vx^{T} \Lambda \vx' \\
 &= - 2 \log( p( y' | \vx' ) ) + \frac{1}{\alpha^{2} - 1} \bigg( ( \vx' )^{T} \Lambda \vx' - 2 \big( ( 1 -  \alpha ) \mu + \alpha \vx \big)^{T} \Lambda \vx' \bigg)
\end{align*}

Thus, we find the maximizer of our posterior by solving,
\begin{align*}
    \vx' &= \arg \min_{\tilde{x}} - 2 \log( p( y' | \tilde{x} ) ) + \frac{1}{\alpha^{2} - 1} \bigg( ( \tilde{x} )^{T} \Lambda \tilde{x} - 2 \big( ( 1 -  \alpha ) \mu + \alpha \vx \big)^{T} \Lambda \tilde{x} \bigg) \\
    &= \arg \min_{\tilde{x}}  - 2 \log( p( y' | \tilde{x} ) ) + \bigg( ( \tilde{x} )^{T} \Lambda \tilde{x} - 2 \big( ( 1 -  \alpha ) \mu + \alpha \vx \big)^{T} \Lambda \tilde{x} \bigg)
\end{align*}

Finally, substitute $-2 \log( p( y | \tilde{x} ) )$ with some convex loss function, and we get the form in Eq.~\eqref{eq:opt_ours},
\begin{align*}
    \vx' = \arg \min_{\tilde{x}} \hspace{3pt} \tilde{x}^{T} \Lambda \tilde{x} - 2 \tilde{x}^{T} \Lambda \big( ( 1 - \alpha ) \mu + \alpha \vx \big) + \gamma || y' - f_{\theta}( \tilde{x} ) ||.
\end{align*}

\newpage
\section{Derivation of Counterfactual Distribution Under Figure \ref{fig:pgm3}}
\label{app:derivation_pgm3}

Following the approach from Appendix \ref{app:derivation_pgm1}, here, we derive the parameters of the counterfactual distribution under the proposed prior.

The counterfactual distribution entailed by Figure \ref{fig:pgm3} is formed as the posterior of,
\begin{align*}
    p( \vx' | \vx, y' ) &\propto p( y' | \vx' ) p( \vx | \vx' ) p( \vx' ) \\
    &= \mathcal{N}( y' | A \vx' + b, L^{-1} ) \times \mathcal{N}( \vx' | \vx, W^{-1} ) \times \mathcal{N}( \mu, \Lambda^{-1} )
\end{align*}

The negative log probability is expressed as follows:
\begin{align*}
    - \log( p( \vx' | \vx, y' ) ) &\propto ( y - A \vx' - b )^{T} L ( y - A \vx' - b ) + ( \vx' - \vx )^{T} W ( \vx' - \vx ) + ( \vx' - \mu )^{T} \Lambda ( \vx' - \mu )\\
    &= y^{T} L y - 2 y^{T} L A \vx' - 2 y^{T} L b + \vx'^{T} ( A^{T} L A ) \vx' + 2 b^{T} L A \vx' + b^{T} L b +  \vx'^{T} W \vx' - 2 \vx^{T} W \vx' \\
    &\hspace{1cm}+ \vx^{T} W \vx + \vx'^{T} \Lambda \vx' - 2 \mu^{T} \Lambda \vx' + \mu^{T} \Lambda \mu
\end{align*}

Grouping the quadratic terms together,
\begin{align*}
    \vx' \Lambda_{cf} \vx' &= \vx' ( A^{T} L A + W + \Lambda).
\end{align*}

Thus the covariance of the distribution is, 
\begin{equation*}
    \Lambda_{cf}^{-1} = ( A^{T} L A + W + \Lambda)^{-1}
\end{equation*}

Grouping the linear terms together,
\begin{align*}
    -2 \mu^{T} \Lambda_{cf} \vx' = -2 ( y^{T} L A - b^{T} L A + \vx W + \mu \Lambda)^{T} \vx',
\end{align*}
This implies that the mean of the distribution of counterfactuals is,
$$\mu_{cf} = \Lambda_{cf}^{-1} ( A^{T} L y' - A^{T} L b + W \vx + \mu \Lambda )$$

Thus the distribution of counterfactuals for the Linear Regression case under PGM \ref{fig:pgm3} is,
\begin{align*}
    \vx' &\sim \mathcal{N}( \mu_{cf}, \Lambda^{-1}_{cf} ) \\
    \Lambda_{cf}^{-1} &= ( A^{T} L A + W + \Lambda)^{-1} \\
    \mu_{cf} &= \Lambda_{cf}^{-1} ( A^{T} L y' - A^{T} L b + W \vx + \mu \Lambda ),
\end{align*}

\newpage
\section{Accounting for Causal Relationships among Features in the Prior}
\label{app:causal_feasibility}

In Section \ref{sec:feasible_cf}, we introduced several notions of plausibility and showed how our approach can express them. Arguably, the most interesting form of plausibility focused on here are those features that do not change independently, but change as a result of other features changing (ie. mutable, non-actionable). In order to account for mutable, non-actionable features, we treat such features as being causal descendants of other features as a Linear Structural Causal Model (SCM), and re-evaluate the posterior with this mindset. In this section, we provide more detail on how causal relationships can be incorporated in not only the mutable, non-actionable features, but also in the initial prior over the data distribution.

In many real-word settings, we have some understanding that our covariates are not independent. One may assume that information such as education status, marital status, gender, etc are not simply correlated with outcomes such as income, but are direct \emph{causes} of their outcome. In recent years, there has been a great deal of work on the importance of a causal understanding of statistical outcomes \cite{peters2017elements}, and prior work has also challenged the community with finding new methods of incorporating causal dependencies on input features when explaining why individuals were subject to certain outcomes.

Counterfactual explanations that incorporate causal dependencies may give explanations that better provide actionable recourse for explainees by not only providing a better understanding of the downstream effects of changing one or more features, but also by decreasing the likelihood of making unrealistic explanations.

Consider the case in which there is some known directed acyclic graph (DAG) that describes the causal relationships between features, where the random variables, $C$, are the set of causes, and random variables, $E$ are the set of effects.  Underlying such graphs is some unknown structural causal model (SCM) that maps our causes to the effects with the non-deterministic function, $f_{E}$ and noise distributions $N_{E}$ and $N_{C}$. As stated in \cite{peters2017elements}, 
\begin{definition}[Structural Causal Model]
An SCM with graph $C \rightarrow E$, consists of two assignments, $C := N_{C}$ and $E := f_{E}( C, N_{E} )$,
\begin{align*}
    C &:= N_{C} \\
    E &:= f_{E}( C, N_{E} ),
\end{align*}
with $N_{E} \indep N_{C}$.
\end{definition}
This SCM entails a joint distribution, $P_{C,E}$ over $C$ and $E$. In order to ensure that generated counterfactual explanations incorporate causal dependencies, one simply has to assign this joint distribution over the SCM as the prior on the distribution of explanations. 

Yet, it is unlikely that one will have access to the true, underlying SCM in any real world case. In absence of other information, it is common to assume linear relations among covariates. In the larger context of the approach introduced in this work, using a linear model implies that we are  expressing a \emph{Linear Gaussian Additive Noise Model}. Importantly, this assumption does not necessarily imply a belief that the true data follows such an SCM, but that linear relationships will provide some greater amount of information on the true causal relationships than assuming independence. Should one assume that the data was generated from a Linear Gaussian SCM, specified by $C := \mathcal{N}( \mu, \Sigma )$ and $E :=  A c + b + \epsilon$, where $c \sim C$ and $\epsilon \sim \mathcal{N}( 0, \sigma I )$, then such an SCM will entail a joint Gaussian distribution with 
$$\hat{\mu} = \begin{bmatrix} \mu \\ A \mu + b \end{bmatrix}, \hspace{3mm} \hat{\Sigma} = \begin{bmatrix} \Lambda^{-1} & \Lambda^{-1} A^{T} \\ A \Lambda^{-1} & \sigma I + A \Lambda^{-1} A^{T} \end{bmatrix}$$ By traversing the DAG and calculating/updating $\hat{\mu}$ and $\hat{\Sigma}$ at each child node, we can calculate the full causal joint prior as Gaussian. Thus, we again express the process of generating counterfactual explanation as sampling from a known Gaussian distribution.

\newpage
\section{Reproducibility}
\label{app:reproducibility}

\subsection{Fashion MNIST and MNIST}

When generating the counterfactual images for both the Fashion MNIST dataset (Figure \ref{fig:fashion_mnist_all}) and the standard MNIST Dataset \ref{app:mnist}, we preprocess the dataset with Pytorch's grayscale preprocessor to constrain pixel intensity to be between $0$ and $1$, and use a train/test split of $60,000$ training images and $10,000$ test images, and stopped training after reaching 90.93\% accuracy and 97.64\% accuracy on their respective test sets. As mentioned in section \ref{sec:fashion_mnist}, we apply a logit transform, $f( x ) = \log( \frac{| x - 0.01 |}{ 1 - | x - 0.01 |} )$ to the grayscaled images and express the data distribution's mean and covariance as the mean and covariance of the dataset's logits. In order to ensure that the covariance matrix in non-singular, we apply a small degree of Gaussian noise to each of the pixel logits ($\epsilon \sim \mathcal{N}( 0, 0.05 )$ ). 

After preprocessing we optimized the \citet{wachter2017counterfactual} objective (Equation \eqref{eq:opt}) with the weight on the euclidean distance set to $\gamma = 10^{-3}$ and $10^{-5}$ respectively. 

Our approach sets the weight on the regularizer in equation \eqref{eq:opt_ours} to $\gamma = 10^{3}$ and $10^{2}$ respectively, and set $\alpha=0.99$ and $\alpha=0.3$. For both objectives, we use the negative log-likelihood loss between the desired predicted label, $y'$, and the predicted label, $f_{\theta}( x )$, and each used the Adam optimizer with a learning rate of $0.05$ over $N=1000$ steps. We initialize $\vx'$ to the logit of the original reference.

\subsection{MTurk Data Preparation}

Note that while section \ref{app:mnist} showed how we express our approach as an optimization task, we do not generate explanations for our method by optimizing Eq.~\eqref{eq:opt_ours}, we instead sample from the Gaussian posterior (Appendix \ref{app:complex}).

For each dataset, we define the underlying prior using the encoder/decoder scheme introduced in appendix \ref{app:cf_decode}. Categorical features are treated as a set of independent Gaussian random variables whose means are the logits of the proportion for each category of the feature, and whose variance is $1$. Continuous features are treated as Gaussian with mean and variance determined from the data. If features have different scales (eg. income and age), we first perform a log transform. All methods operate within this latent space, before being transformed back into their original scales. Categorical features use a softmax decoder with temperature parameter $0.01$

\subsection{MTurk Hyperparameters and Model Architectures}

For the Adult, LUCAS, and German Credit datasets, we train an MLP Classifier with 2 hidden layers (width 50 and 20 respectively). Counterfactuals generated through FACE use K=20, when building the K-Nearest Neighbor Graph. In optimizing DiCE, we determined hyperparameters through a grid search over the three quantitative metrics introduced in the original paper: Validity, Proximity, and Diversity.

\newpage
\section{MNIST Counterfactual Explanations}
\label{app:mnist}

\begin{figure}
    \centering
    \resizebox{\textwidth}{!}{ 
        \begin{tabular}{ c c c c c | c c c c c }
                 & \scalebox{1.5}{Reference} & \scalebox{1.5}{$L2$ Distance} & \scalebox{1.5}{Ours: $\alpha=0.99$} & \scalebox{1.5}{Ours: $\alpha=0.3$} & & \scalebox{1.5}{Reference} & \scalebox{1.5}{$L2$ Distance} & \scalebox{1.5}{Ours: $\alpha=0.99$} & \scalebox{1.5}{Ours: $\alpha=0.3$}~ \\
                \raisebox{8mm}{\rotatebox[origin=c]{0}{ \LARGE{$6 \rightarrow 8$}}} &
                 \includegraphics[width=0.22\linewidth]{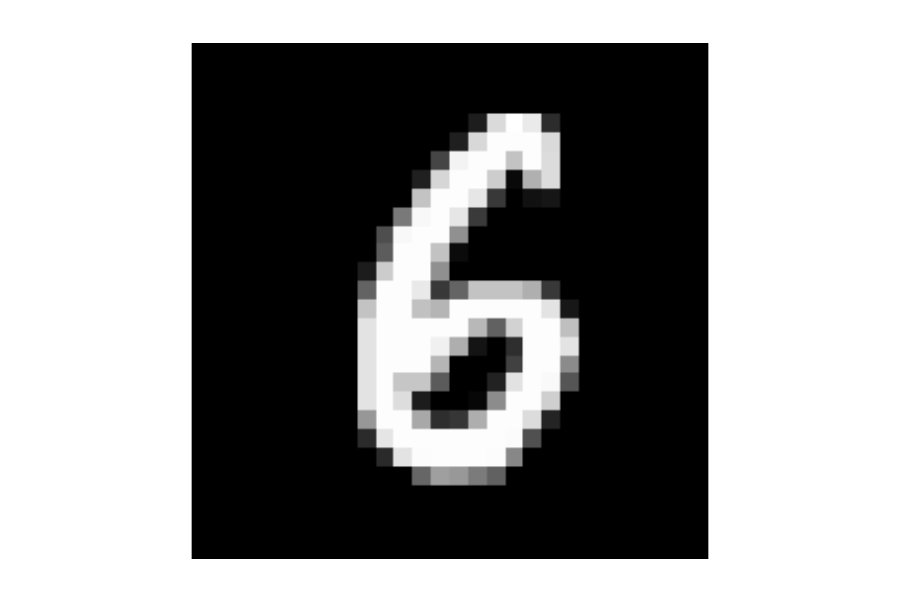} &
                 ~
                \includegraphics[width=0.22\linewidth]{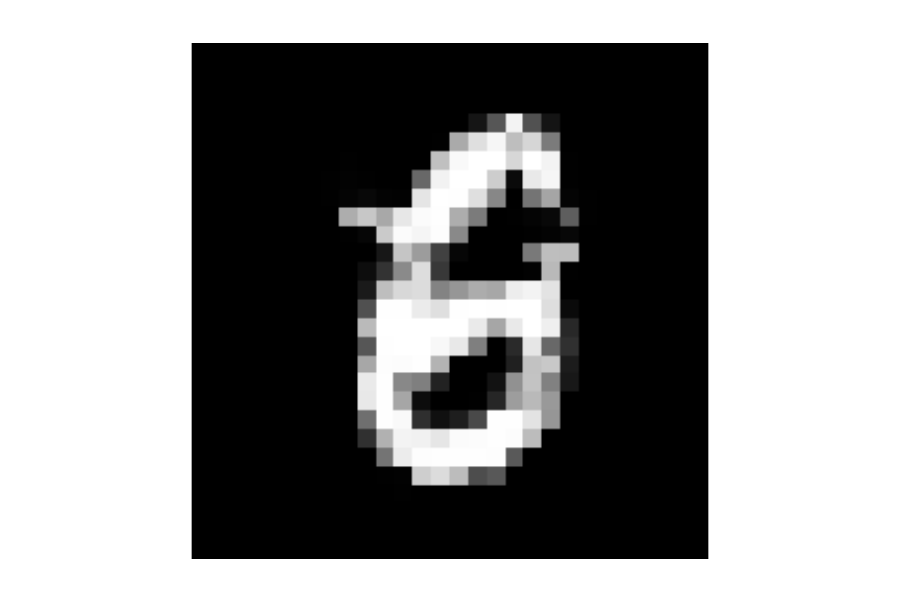} &
                ~
                \includegraphics[width=0.22\linewidth]{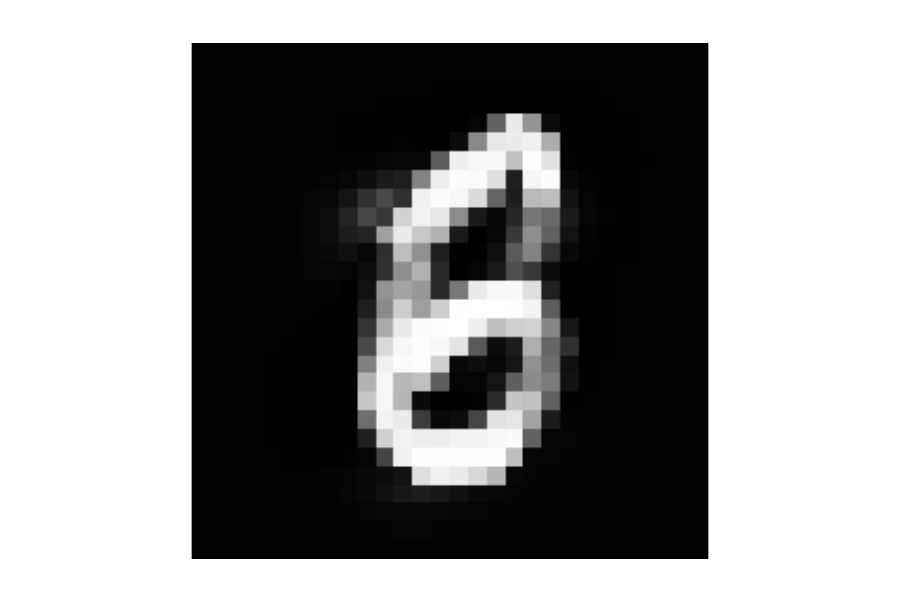} &
                ~
                \includegraphics[width=0.22\linewidth]{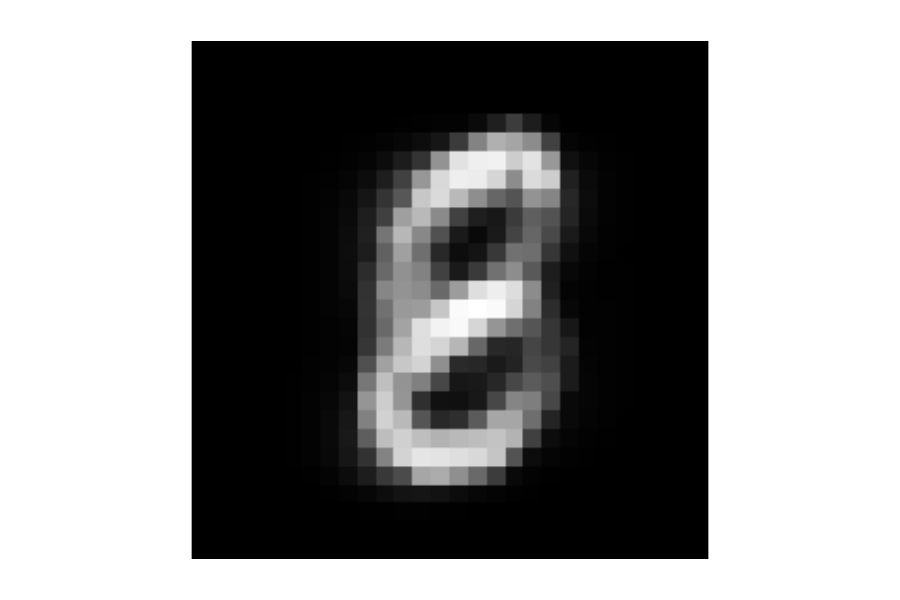} &
                ~
                \raisebox{8mm}{\rotatebox[origin=c]{0}{ \LARGE{$7 \rightarrow 9$}}} &
                ~
                 \includegraphics[width=0.22\linewidth]{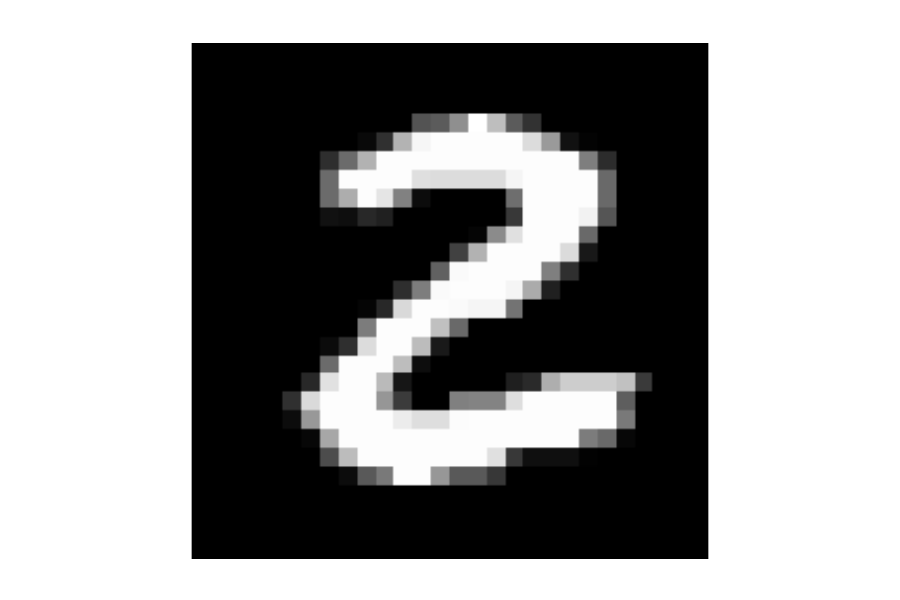} &
                 ~
                \includegraphics[width=0.22\linewidth]{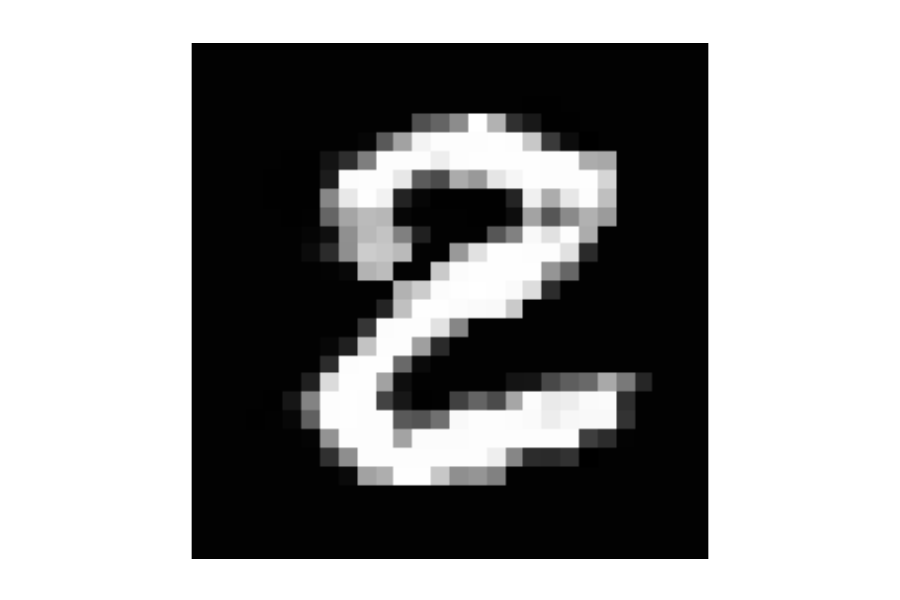} &
                ~
                \includegraphics[width=0.22\linewidth]{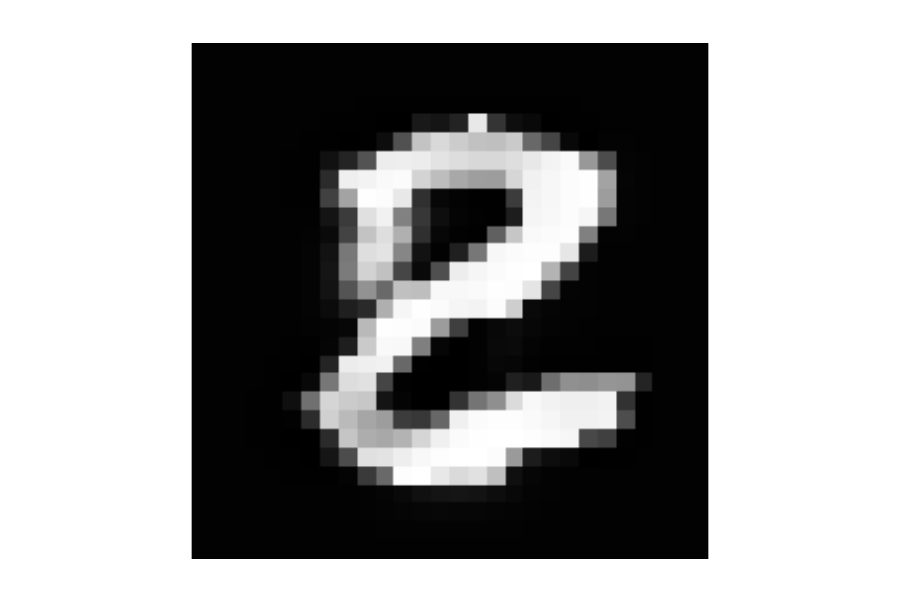} &
                ~
                \includegraphics[width=0.22\linewidth]{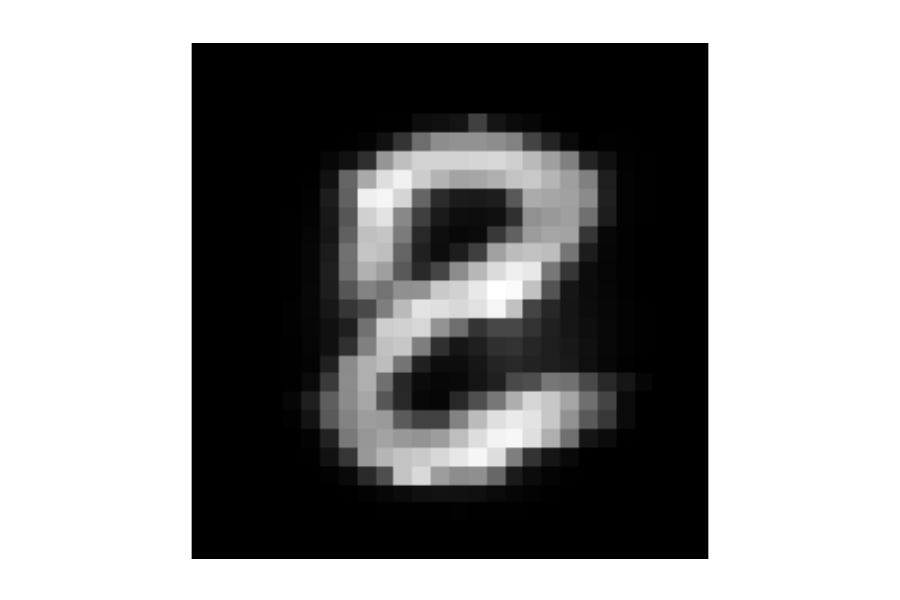} 
                \\
                \raisebox{8mm}{\rotatebox[origin=c]{0}{ \LARGE{$9 \rightarrow 4$}}} &
                 \includegraphics[width=0.22\linewidth]{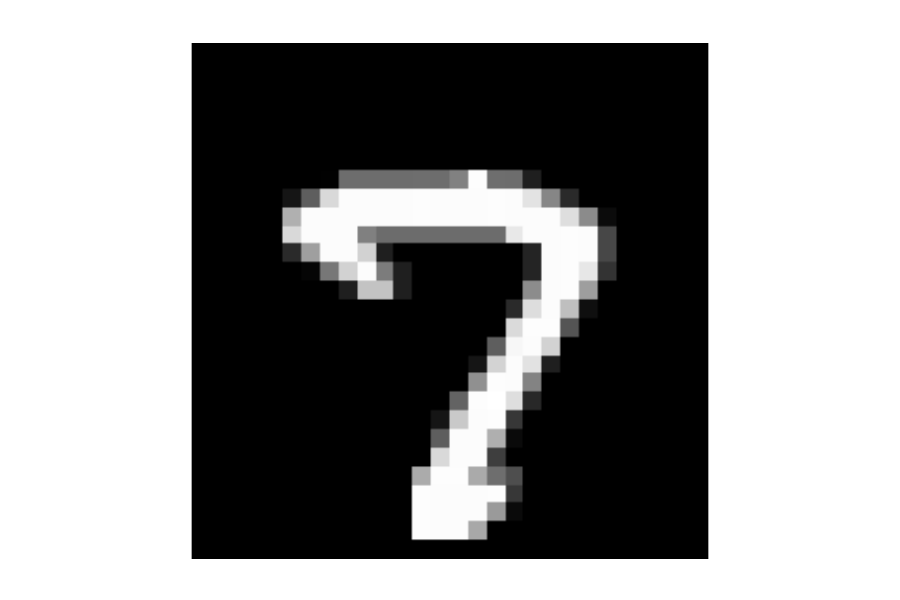} &
                 ~
                \includegraphics[width=0.22\linewidth]{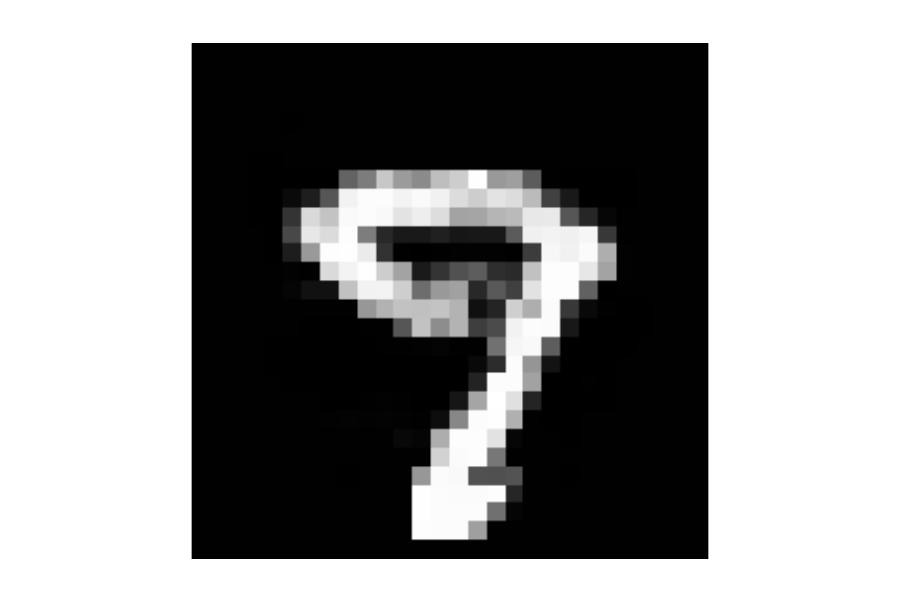} &
                ~
                \includegraphics[width=0.22\linewidth]{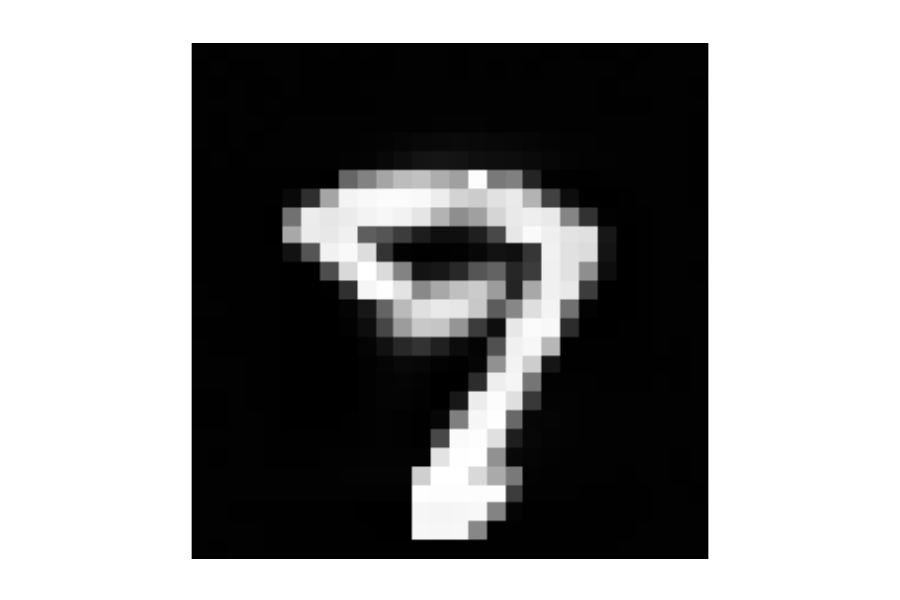} &
                ~
                \includegraphics[width=0.22\linewidth]{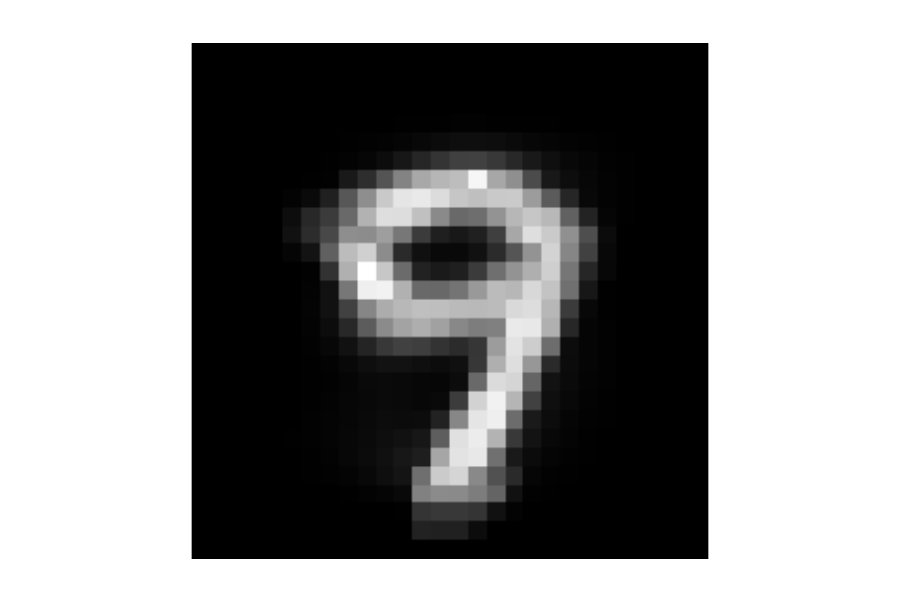} &
                ~ 
                \raisebox{8mm}{\rotatebox[origin=c]{0}{ \LARGE{$2 \rightarrow 8$}}} &
                ~
                 \includegraphics[width=0.22\linewidth]{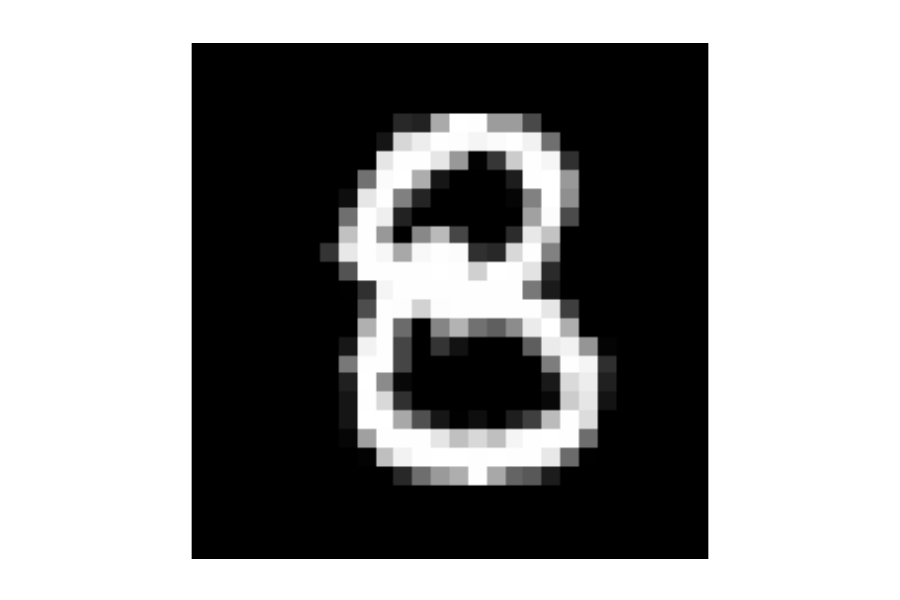} &
                 ~
                \includegraphics[width=0.22\linewidth]{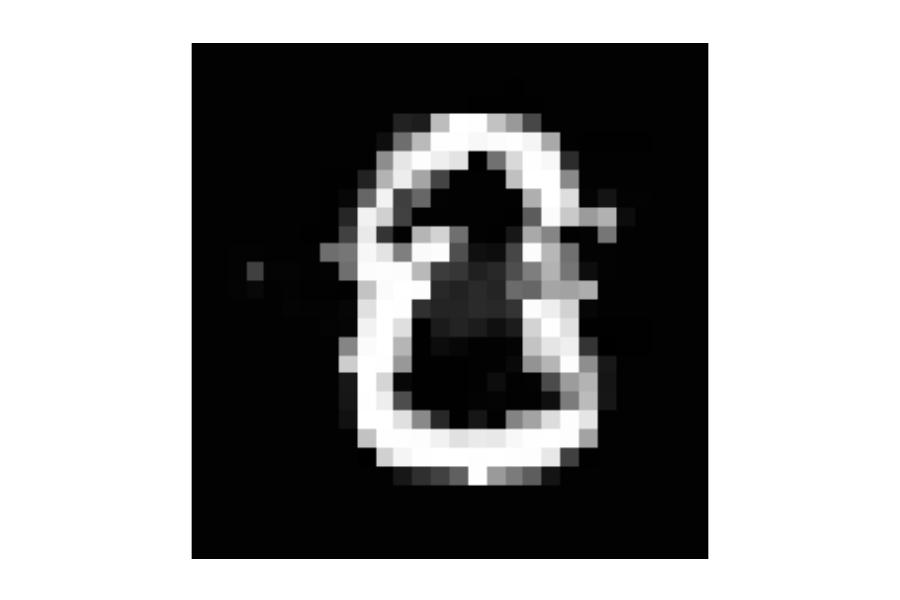} &
                ~
                \includegraphics[width=0.22\linewidth]{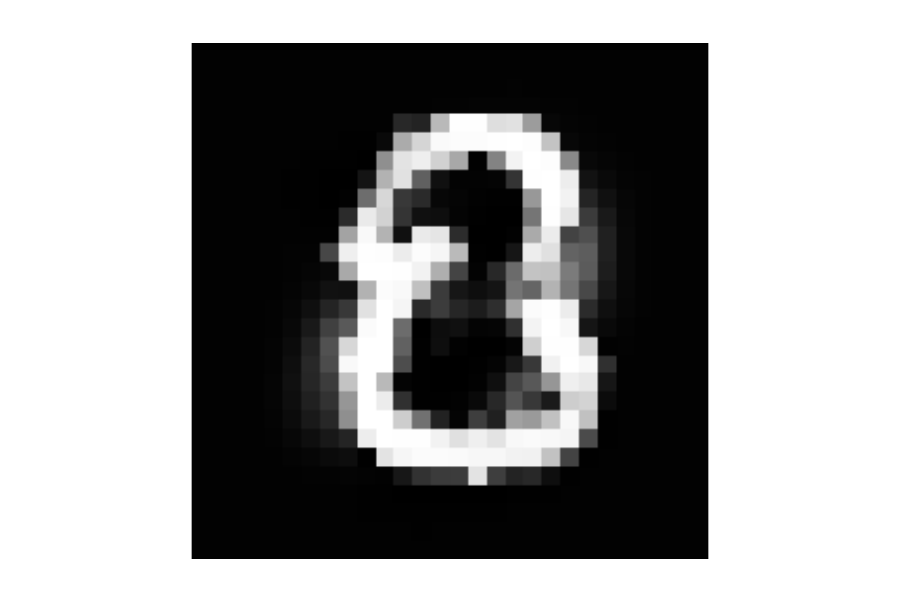}  &
                ~
                \includegraphics[width=0.22\linewidth]{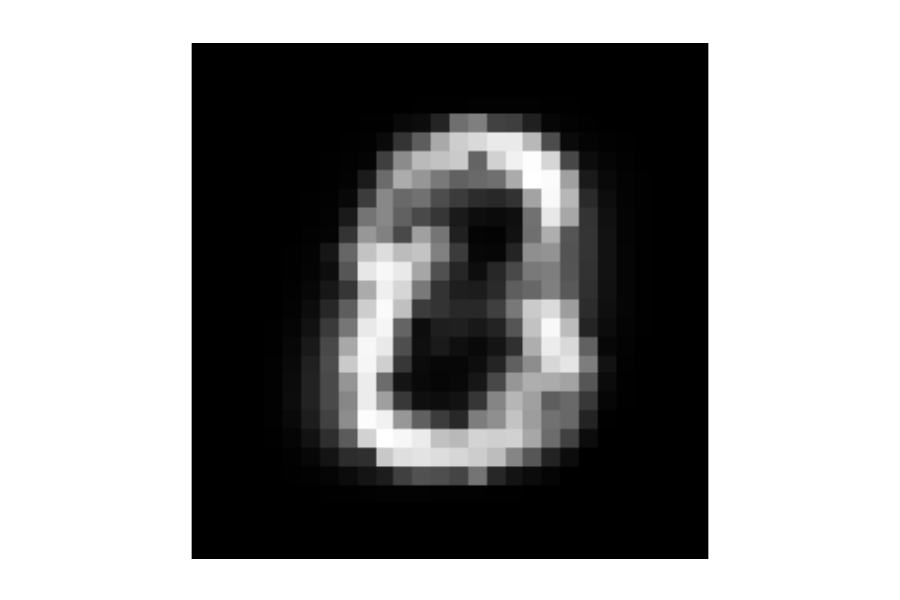}
            \\
               \raisebox{8mm}{\rotatebox[origin=c]{0}{ \LARGE{$3 \rightarrow 5$}}} &
                 \includegraphics[width=0.22\linewidth]{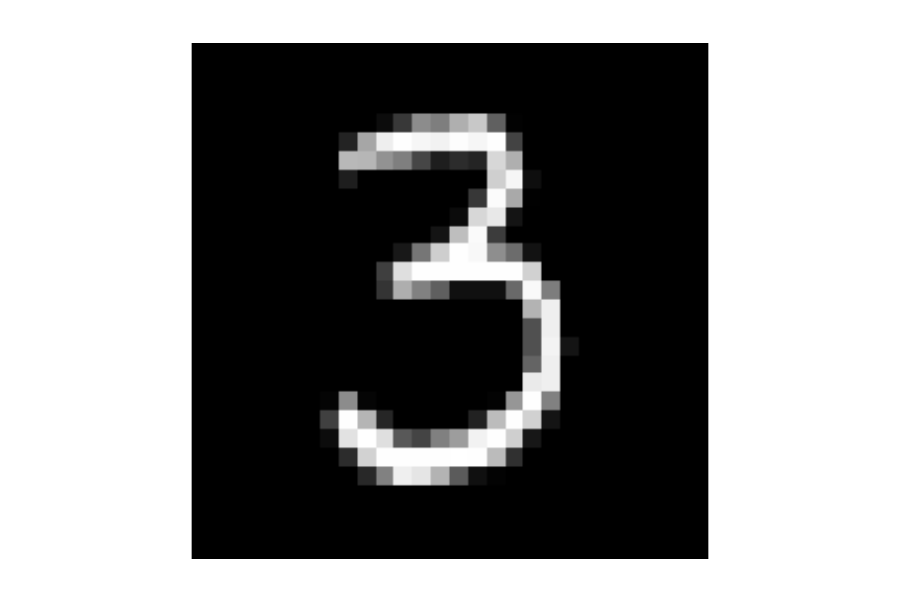} &
                 ~
                \includegraphics[width=0.22\linewidth]{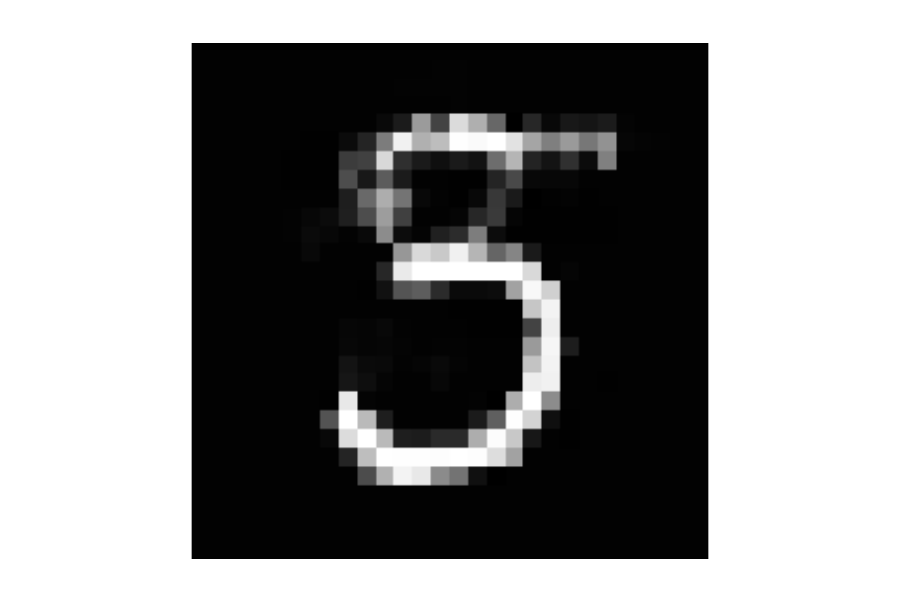} &
                ~
                \includegraphics[width=0.22\linewidth]{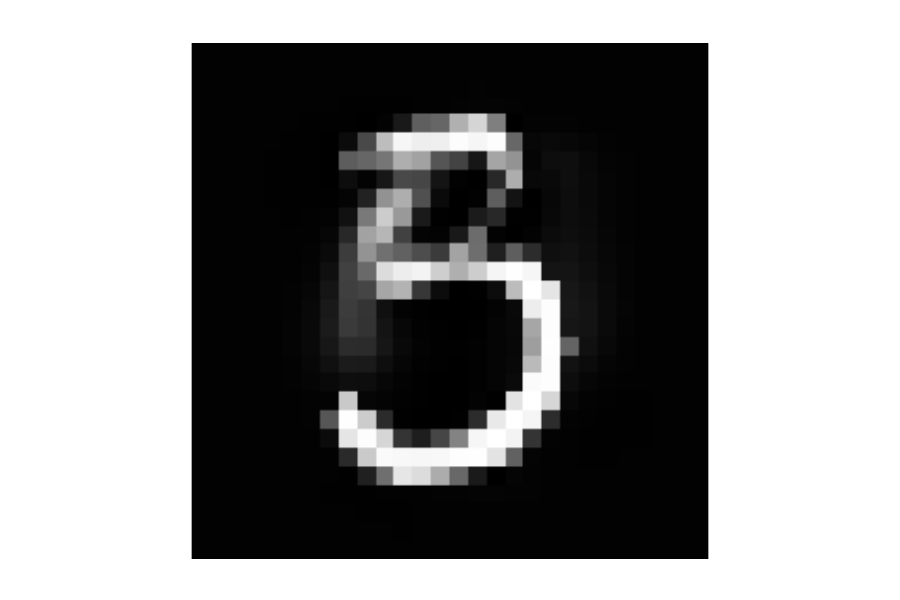} &
                ~
                \includegraphics[width=0.22\linewidth]{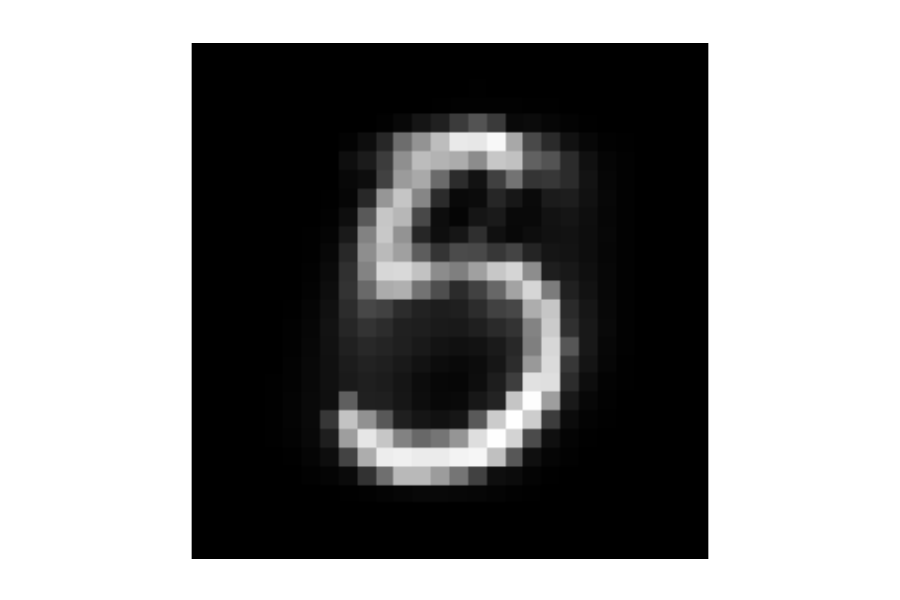} &
    
                \raisebox{8mm}{\rotatebox[origin=c]{0}{ \LARGE{$9 \rightarrow 4$}}} &
                 \includegraphics[width=0.22\linewidth]{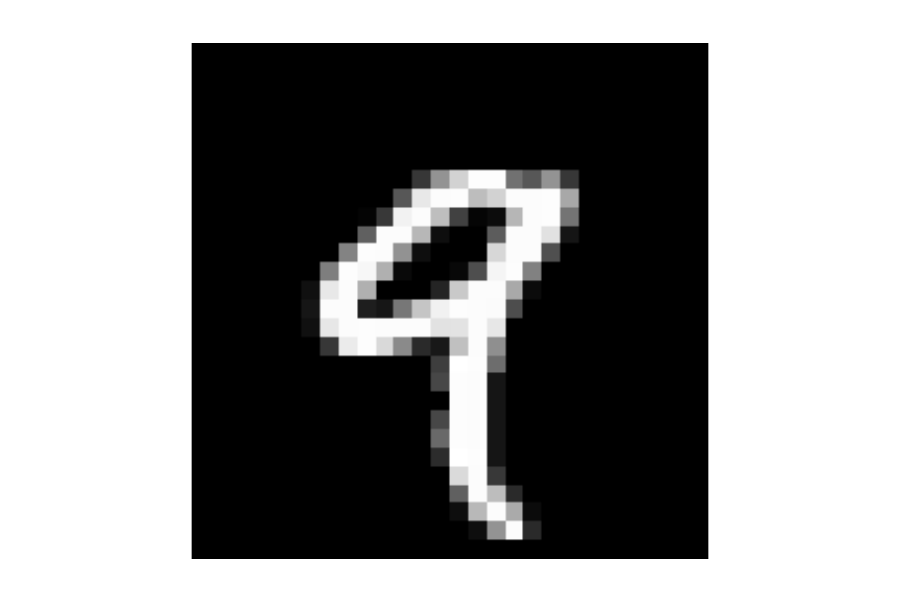} &
                 ~
                \includegraphics[width=0.22\linewidth]{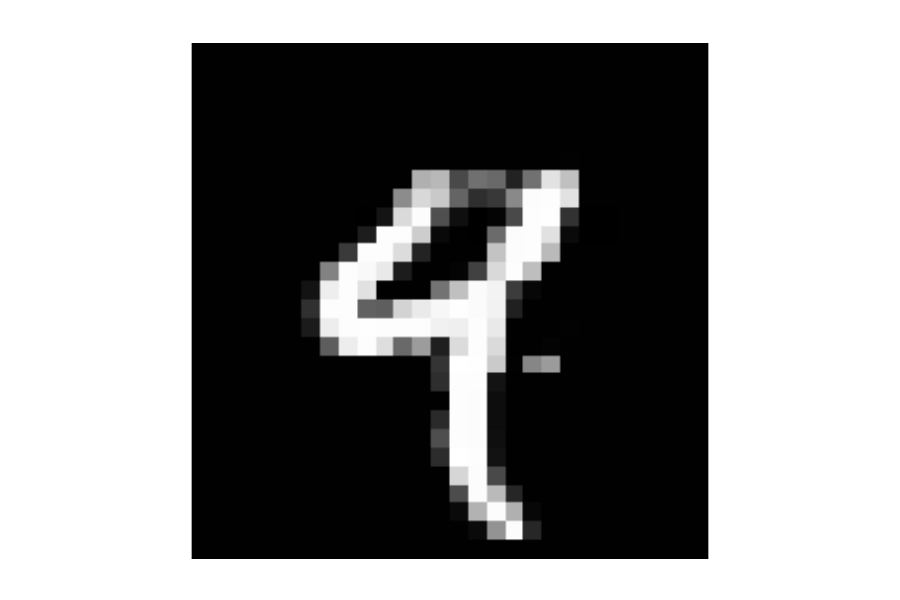} &
                ~
                \includegraphics[width=0.22\linewidth]{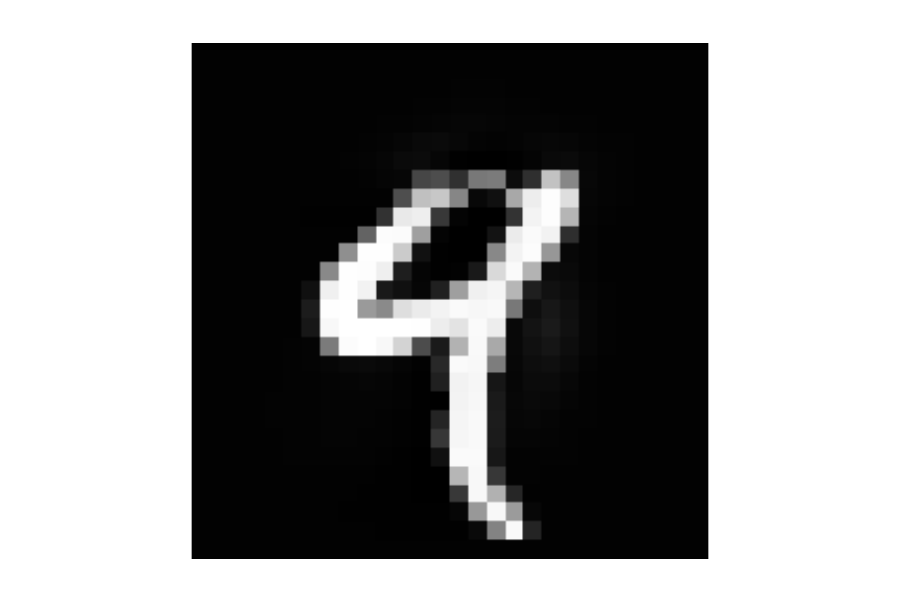} &
                ~
                \includegraphics[width=0.22\linewidth]{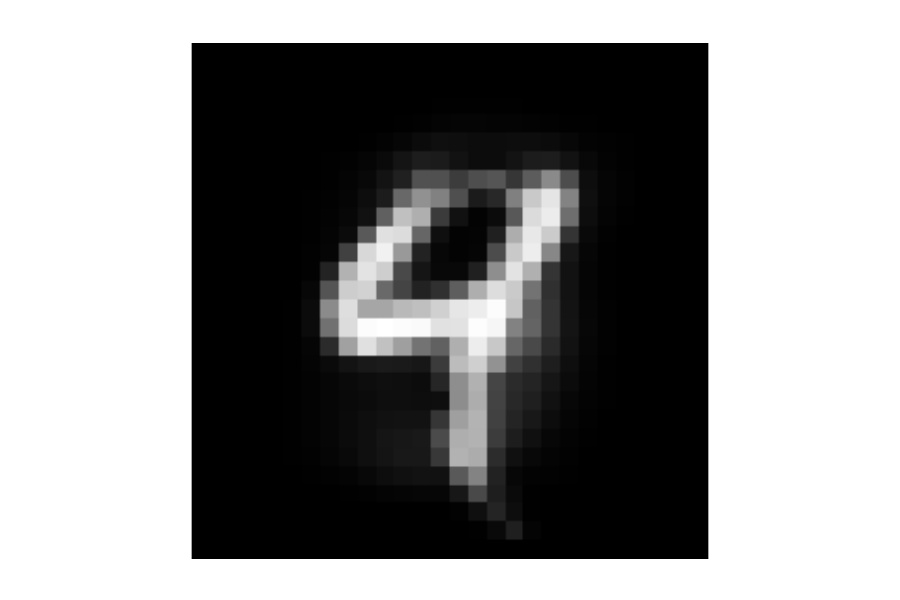}
    \end{tabular}
    }

    \caption{Example MNIST counterfactuals for different distance metrics; For $\alpha \rightarrow 1$, our counterfactuals are analogous to generating counterfactuals through euclidean distance, yet importantly, as we decrease $\alpha$ ie. decrease the reliance on similarity to the reference, rather than devolving into adversarial examples, we generate images closer to prototypical examples for the desired class.}
    \label{fig:mnist_all}
\end{figure}

In order to generate the images in Figure \ref{fig:mnist_all}, we train a simple convolutional neural network CNN, $f_{\theta}: \mathcal{X} \rightarrow \{0,1\}^{10}$ for the purpose of classifying handwritten digits from the the MNIST dataset \cite{lecun2010mnist}. Our training pipeline, including pre-processing is included in Appendix \ref{app:reproducibility}.

While not a dataset that one traditionally treats as Gaussian, we map MNIST into our setting by applying a logit transform, $\log \frac{| x - 0.01 |}{ 1 - | x - 0.01 |} $ to the grayscaled images and express the data distribution's mean and covariance as the mean and covariance of the dataset's logits. In order to ensure that the covariance matrix in non-singular, we apply a small degree of Gaussian noise to each of the pixel logits. As mentioned in Appendix \ref{app:cf_decode}, a more accurate Gaussian prior for MNIST involves a Gaussian latent space, however, in this case, we fit the prior directly for the purpose of providing a baseline comparison between the approaches.

Figures $9 \rightarrow 4$ and $7 \rightarrow 9$ show the most successful transitions from reference to counterfactual. Moreover, as can be inferred from our comparison of the linear models from Section \ref{sec:background}, setting $\alpha$ close to $1$ in our method returns nearly identical counterfactual explanations as those generated by restricting counterfactuals to be very close to the reference with respect to euclidean distance. 

As we allow explanations to stray further from the reference and closer to the desired class $\alpha = 0.3$, rather than finding explanations that move out of the distribution and become adversarial, we instead become closer to the prototypical form for the desired class. For example, in order achieve a greater degree of symmetry in the generated $8$ for Figure $2 \rightarrow 8$, $\alpha = 0.3$ cuts off the long tail from the reference $2$ unlike the comparison explanations. In a similar vein, Figure $9 \rightarrow 4$ cuts off the longer curved tail on the reference $9$ in order to get closer to an average $4$ from the data distribution. Likewise, Figure $7 \rightarrow 9$ rounds out the pointed edge of the reference $7$ in order to look like a more natural $9$.

\newpage
\section{Dog and Cat Classifier Explanations}
\label{app:pets}

Here, we show a comparison of the counterfactuals generated by the standard counterfactual optimization (Eq. \eqref{eq:opt}) and our approach (Eq. \eqref{eq:opt_ours}). In order to generate the images in Figure \ref{fig:pets}, we train a convolutional neural network (CNN), $f_{\theta}: [0,1]^{64 \times 64 \times 3} \rightarrow [0,1]$, as a binary classifier of cat and dog images, using a subset of the Kaggle Dogs vs Cats dataset \cite{kaggle/dogsvcat}. This subset consists of 9892 train images and 1000 test images that were able to be downsampled clearly to $64 \times 64$ pixel size via the pytorch Resize transform. Our classifier achieved $97.8\%$ accuracy on the test set.

In order to fit a Gaussian prior over this data, we applied a logit transform to the $[0,1]$ pixel intensities and calculated the mean and covariance across each RGB image channel. Similarly to \ref{app:reproducibility}, we apply a small amount of Gaussian noise to each pixel. 

In Figure \ref{fig:pets}, we show that our approach recommends more semantically meaningful changes to the images. Below each image is a map of recommended increases or decreases to pixel intensities in order to generate a counterfactual. Our approach follows the contours and facial structure of  the animal. For example, when generating counterfactual images of dogs to cats, our approach targets its attention to sharpening the eyes and flattening the nose, whereas $l_{2}$ distance engages with similar features, but generally introduces adversarial noise rather than semantically meaningful recommendations.

\begin{figure}[h!]
    \centering
    \includegraphics[scale=0.32]{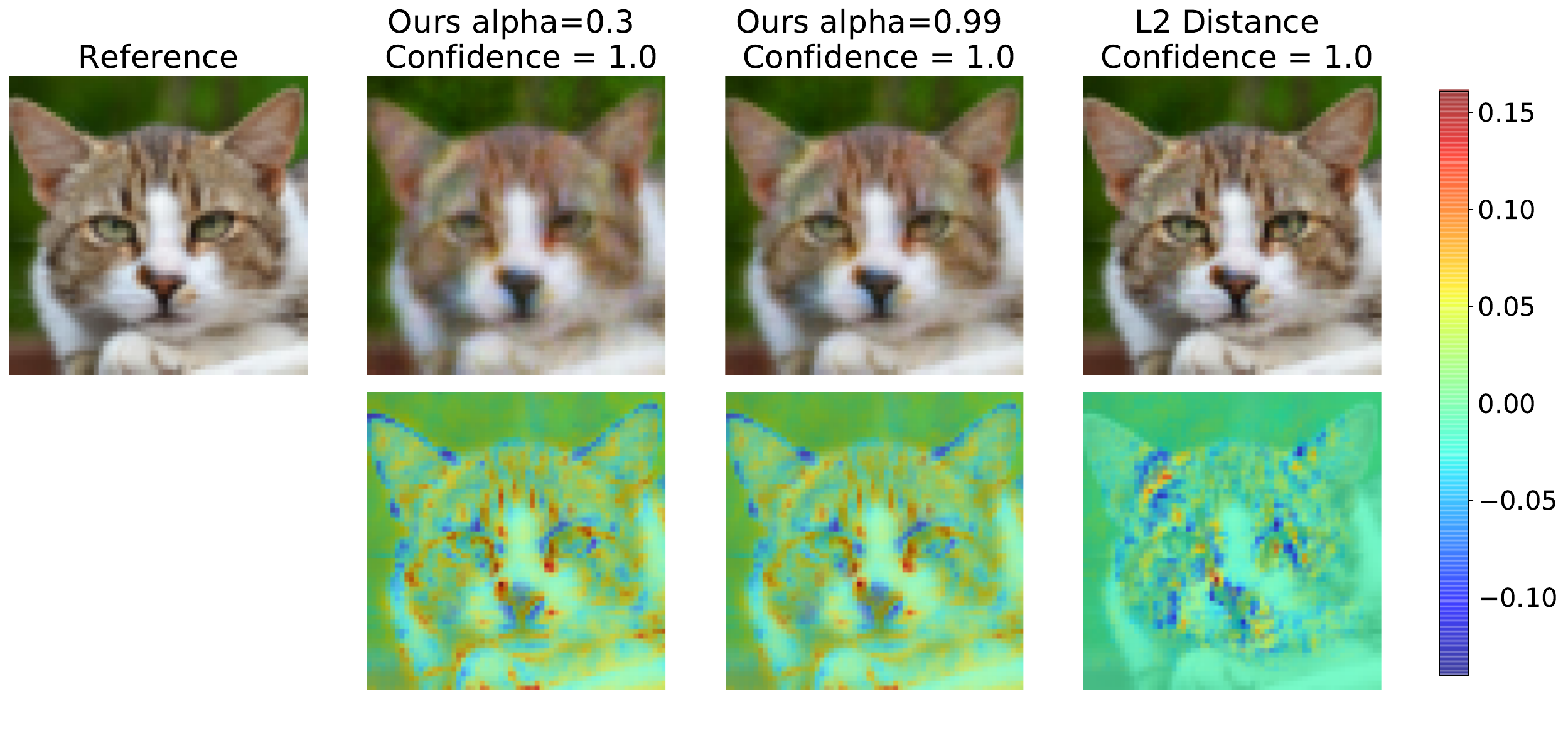}
    \includegraphics[scale=0.32]{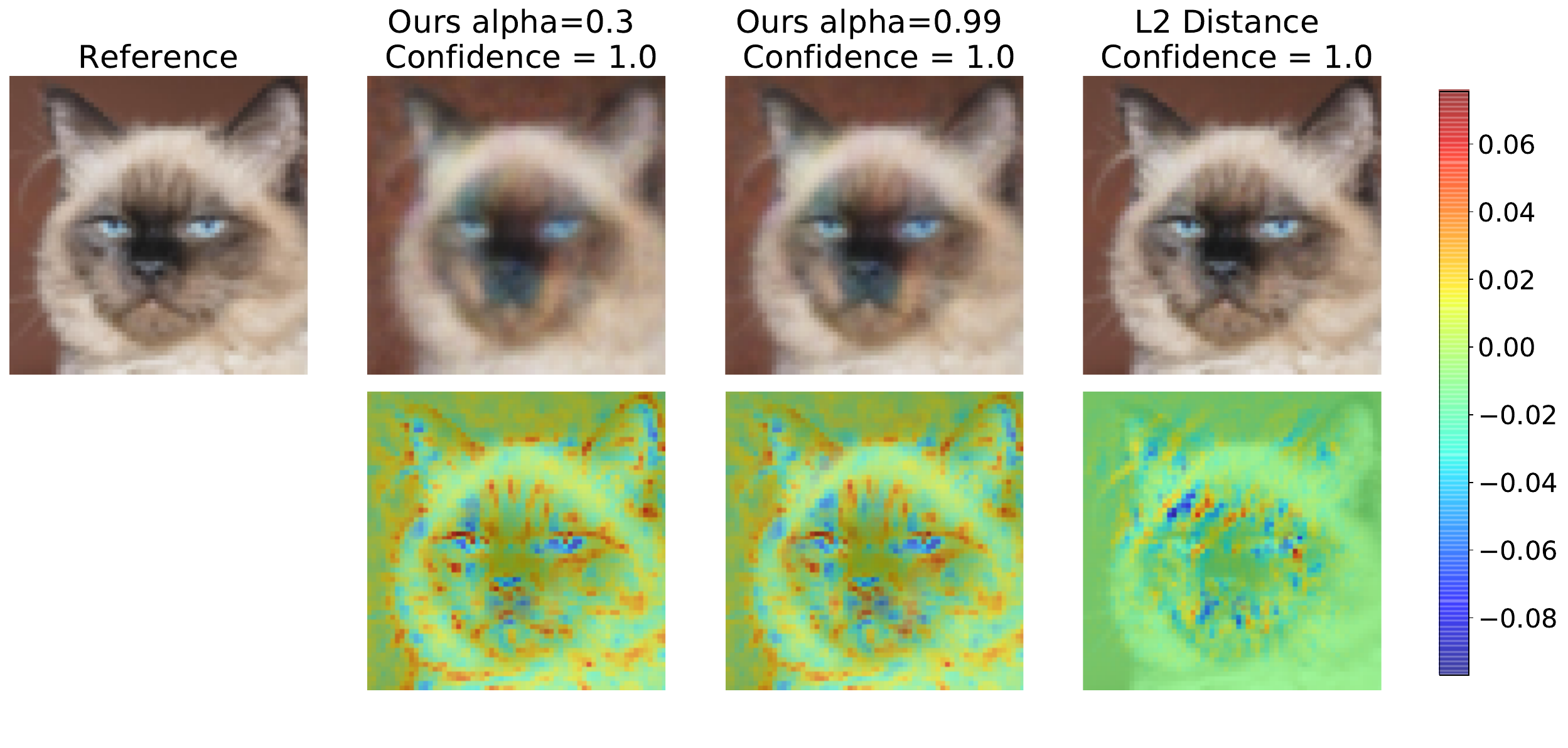}
    \caption{Example Dog to Cat (or vice-versa) counterfactuals for different distance metrics; Below each image, we include the map of pixel changes recommended by each distance metric in order to classify as the counterfactual class. Our approach encourages manipulating semantically meaningful features, and more strongly follow the contours of the animals' face.}
    \label{fig:pets}
\end{figure}
\begin{figure}\ContinuedFloat
    \centering
    \includegraphics[scale=0.32]{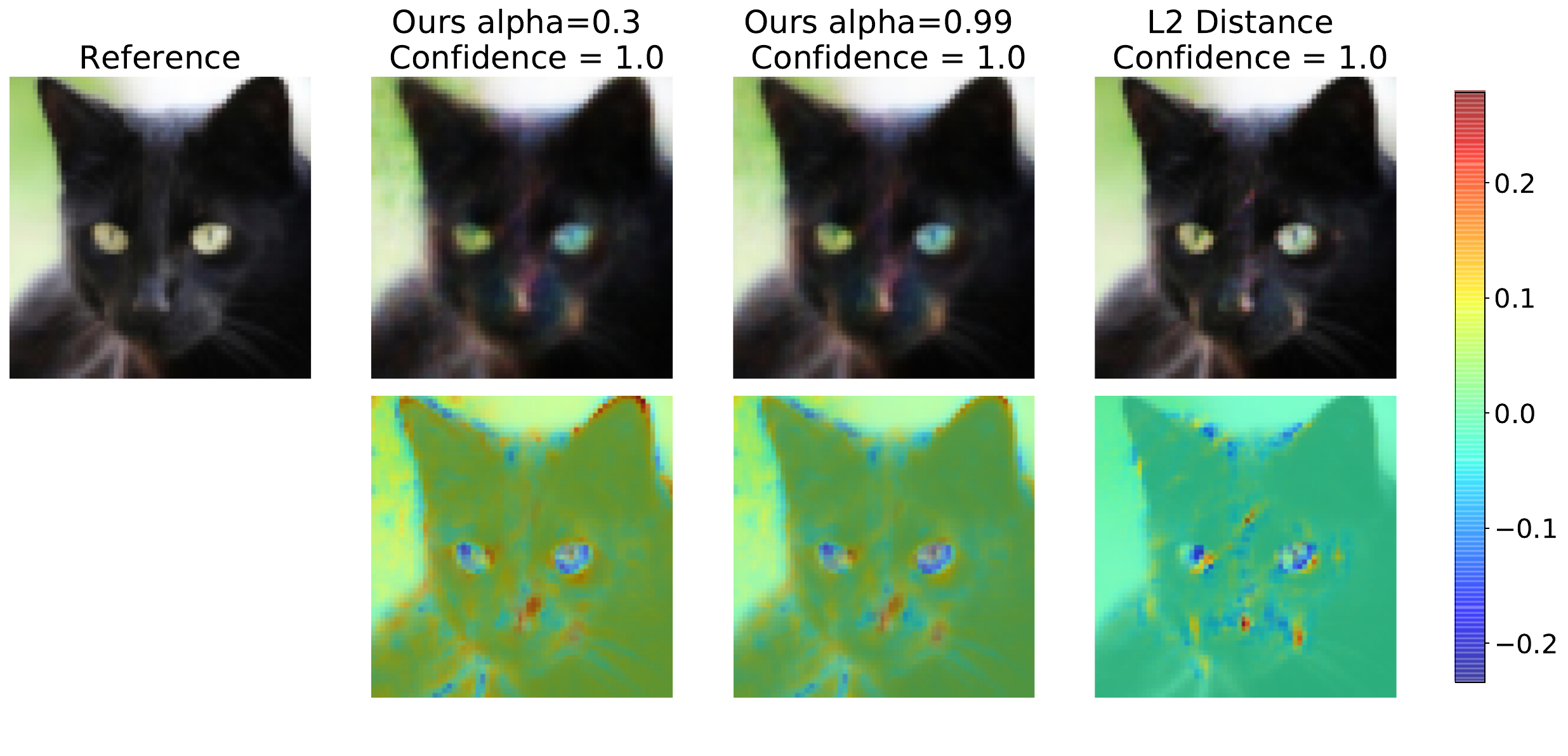}
    \includegraphics[scale=0.32]{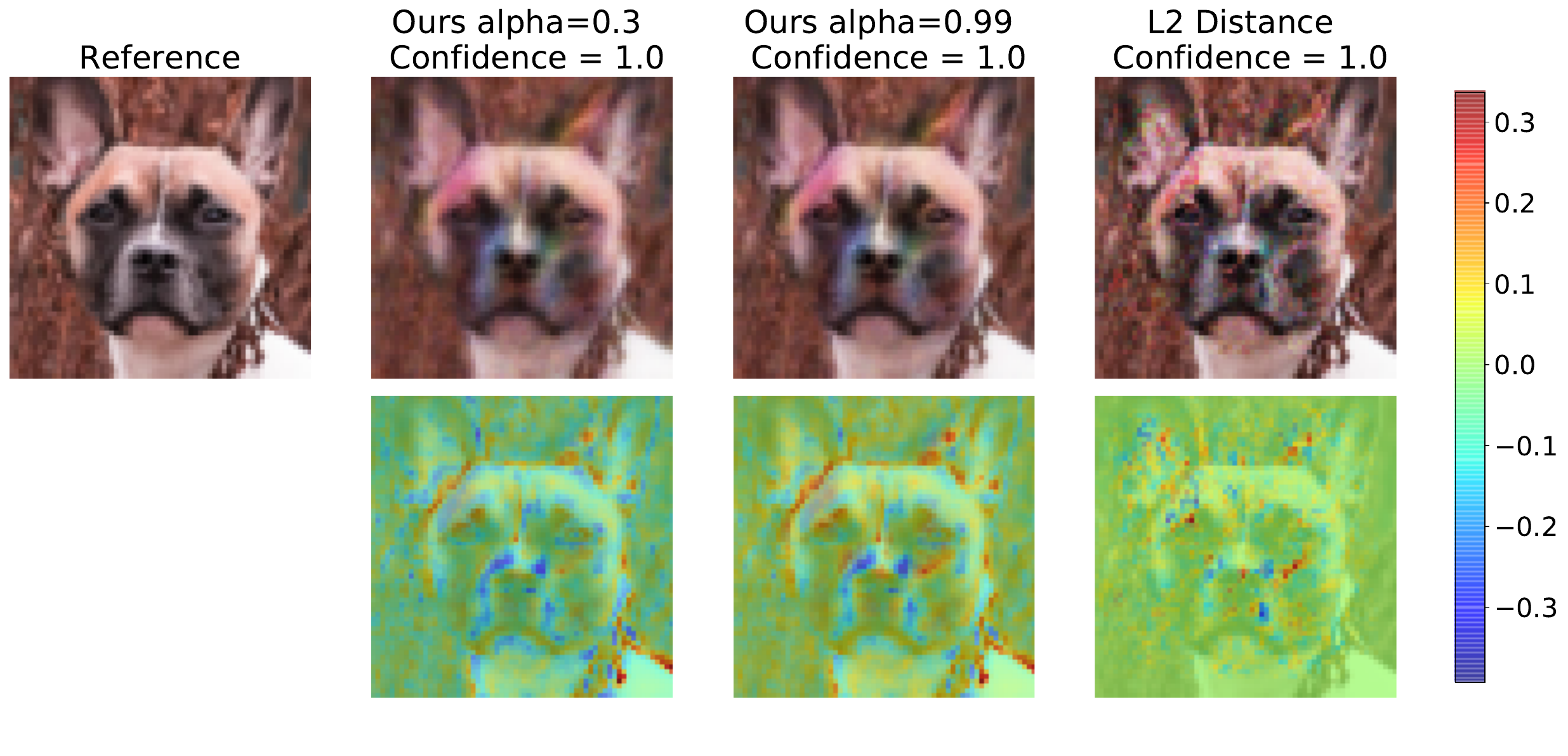}
    \includegraphics[scale=0.32]{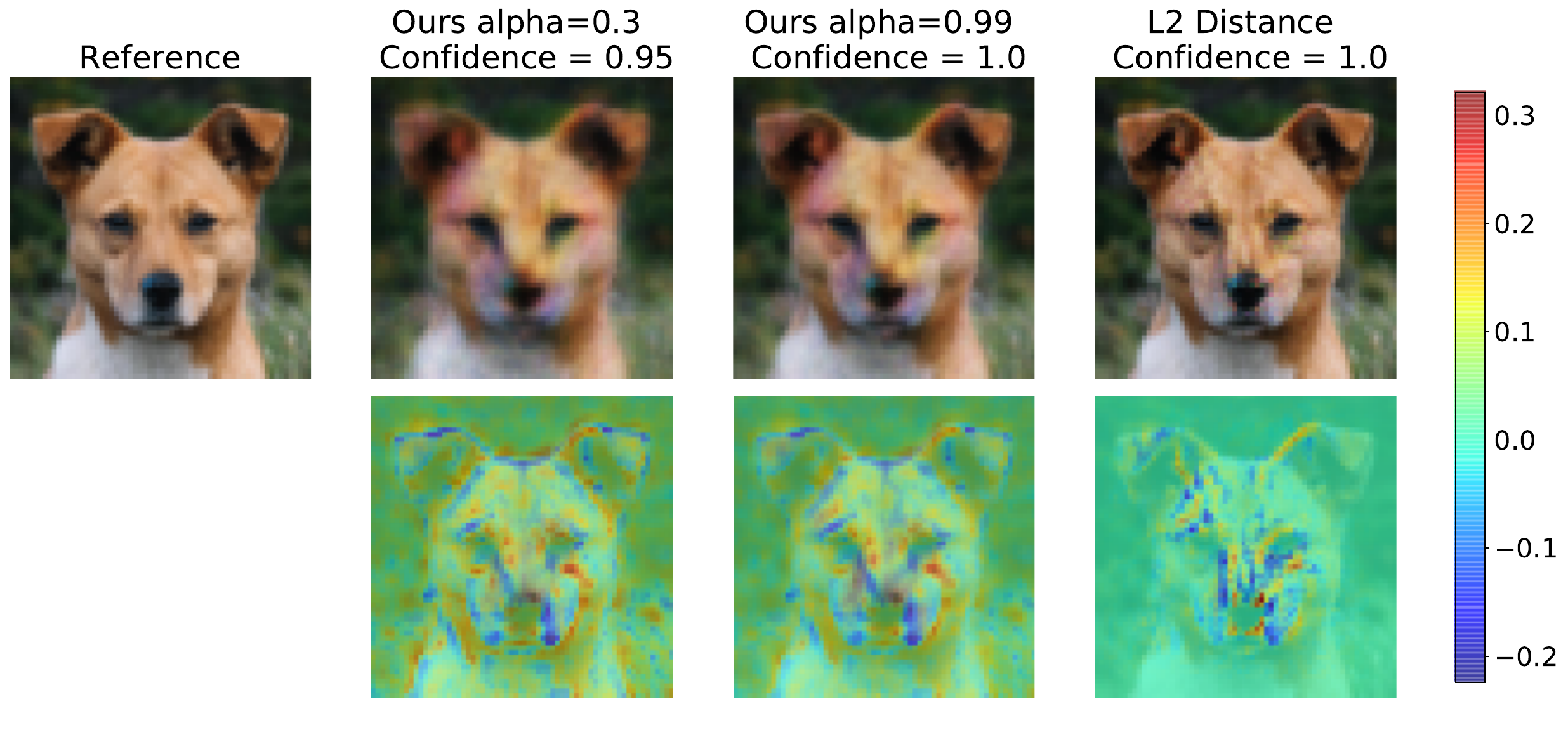}
    \caption{Example Dog to Cat (or vice-versa) counterfactuals for different distance metrics; Below each image, we include the map of pixel changes recommended by each distance metric in order to classify as the counterfactual class. Our approach encourages manipulating semantically meaningful features, and follow the contours of the animals' face.}
    \label{fig:pets}
\end{figure}
\begin{figure}\ContinuedFloat
\includegraphics[scale=0.32]{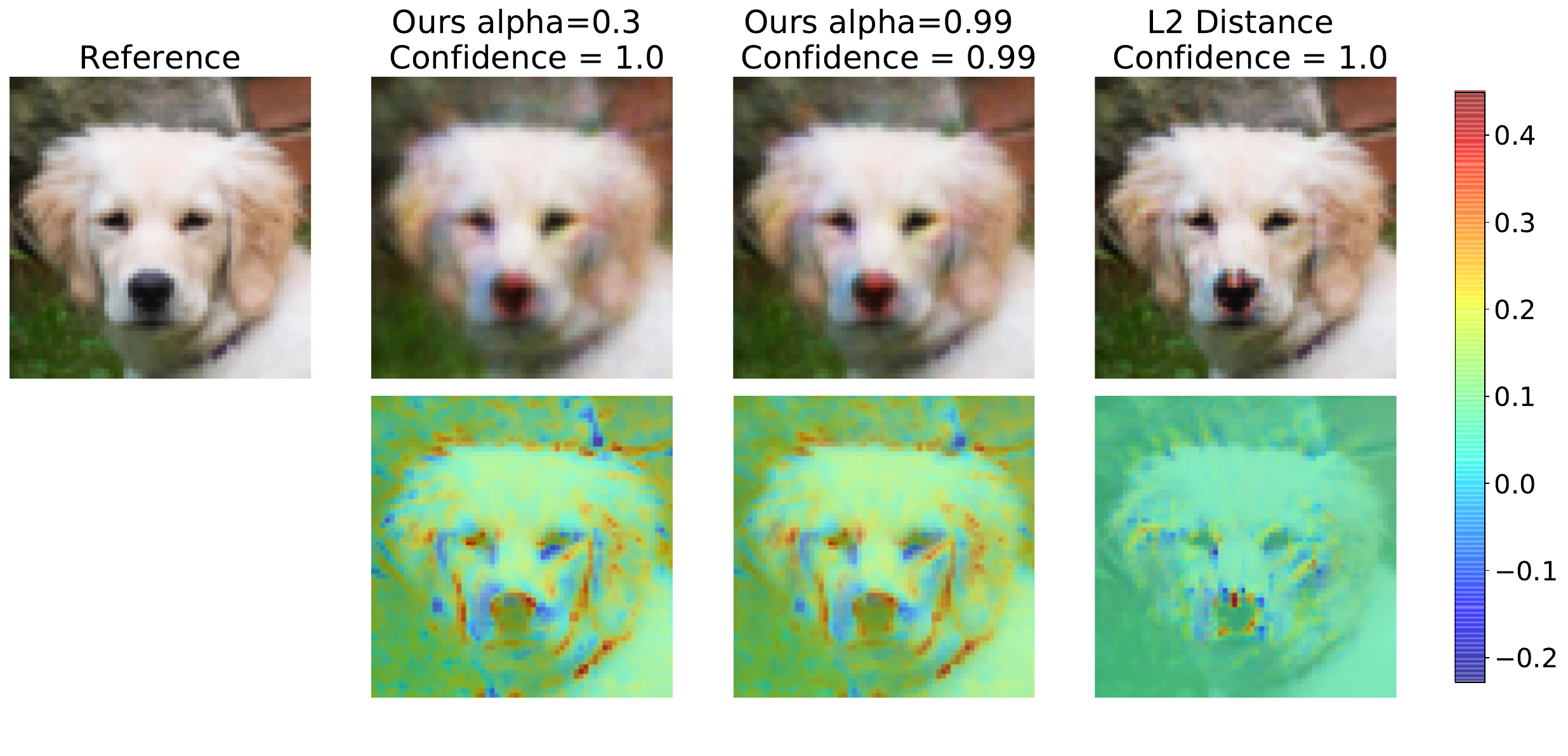}
    \caption{Example Dog to Cat (or vice-versa) counterfactuals for different distance metrics; Below each image, we include the map of pixel changes recommended by each distance metric in order to classify as the counterfactual class. Our approach encourages manipulating semantically meaningful features, and follow the contours of the animals' face.}
    \label{fig:pets}
\end{figure}




\end{document}